\documentclass{article} %
\usepackage[T1]{fontenc}
\usepackage{iclr2027_conference,times}

\usepackage{amsmath,amsfonts,bm}

\def\eqref#1{equation~\ref{#1}}

\def\1{\bm{1}}

\def\vzero{{\bm{0}}}

\def\vc{{\bm{c}}}

\def\ve{{\bm{e}}}

\def\vs{{\bm{s}}}

\def\vz{{\bm{z}}}

\def\mI{{\bm{I}}}

\def\mM{{\bm{M}}}

\DeclareMathAlphabet{\mathsfit}{\encodingdefault}{\sfdefault}{m}{sl}
\SetMathAlphabet{\mathsfit}{bold}{\encodingdefault}{\sfdefault}{bx}{n}

\def\gE{{\mathcal{E}}}

\def\gG{{\mathcal{G}}}

\def\gV{{\mathcal{V}}}

\newcommand{\E}{\mathbb{E}}
\newcommand{\Ls}{\mathcal{L}}
\newcommand{\R}{\mathbb{R}}

\newif\ifrebuttal
\rebuttaltrue
\providecommand{\rebuttalcolorname}{black}
\providecommand{\rebuttal}[1]{\ifrebuttal{\color{\rebuttalcolorname}#1}\else#1\fi}
\providecommand{\rebuttalcolor}{\ifrebuttal\color{\rebuttalcolorname}\fi}
\providecommand{\rebuttaltwocolorname}{black}
\providecommand{\rebuttaltwo}[1]{\ifrebuttal{\color{\rebuttaltwocolorname}#1}\else#1\fi}
\providecommand{\rebuttaltwocolor}{\ifrebuttal\color{\rebuttaltwocolorname}\fi}

\newcommand{\src}{u}
\newcommand{\dst}{v}
\newcommand{\efeat}{\ve}
\newcommand{\getemb}{E}

\newcommand{\concat}{\mathrm{concat}}
\newcommand{\temp}{\gamma}
\newcommand{\tdiff}{\tau}
\newcommand{\reconscore}{score}
\newcommand{\tstrat}{Strategy}

\newcommand{\reddits}{Reddit 12F\xspace}
\newcommand{\redditf}{Reddit 172F\xspace}
\newcommand{\nbsteps}{NBS\xspace}

\newcommand{\modelname}{DyGDi\xspace}
\newcommand{\modelfullname}{Dynamic Graph Diffusion\xspace}
\newcommand{\dggen}{DG-Gen\xspace}
\newcommand{\tigger}{Tigger\xspace}
\newcommand{\tiggerind}{Tigger-I\xspace}

\newcommand{\memcollapse}{memory matching collapse\xspace}
\newcommand{\Memcollapse}{Memory matching collapse\xspace}

\newcommand{\wincollapse}{window matching collapse\xspace}
\newcommand{\Wincollapse}{Window matching collapse\xspace}

\newcommand{\eventclass}{event class information}

\newcommand{\seen}{\gV_{\text{seen}}}
\newcommand{\unseen}{\gV \setminus \gV_{\text{seen}}}

\newcommand{\zpast}{\vz_{\text{past}}}
\newcommand{\zsum}{\vz_{\text{sum}}}
\newcommand{\zcur}{\vz}
\newcommand{\znoisy}{\vz_{\tdiff}} %
\newcommand{\zcurhat}{\hat{\vz}} %

\newcommand{\nodedegree}{ND\xspace}
\newcommand{\clocen}{CLO\xspace}
\newcommand{\btwc}{BC\xspace}

\newcommand{\wedgecount}{WC\xspace}
\newcommand{\powerlaw}{PLE\xspace}
\newcommand{\nbcc}{NCC\xspace}
\newcommand{\lcc}{LCC\xspace}
\newcommand{\rel}{REL\xspace}

\newcommand{\featprec}{round-mode=places,round-precision=2,table-format=1.2}
\newcommand{\topoprec}{round-mode=figures,round-precision=3,exponent-mode=scientific,output-exponent-marker=\text{e},mode=text,table-format=1.2e-1}
\newcommand{\jlprec}{round-mode=places,round-precision=2,table-format=1.2}
\newcommand{\pctprec}{round-mode=places,round-precision=0}

\newcommand{\pctofcolor}{teal}
\edef\pctof#1{{\noexpand\scriptsize\noexpand\color{\noexpand\pctofcolor}(\noexpand\num[\pctprec]{#1}\noexpand\%)}}

\newcommand{\wrapmean}[1]{#1\xspace}
\newcommand{\wrapmedian}[1]{#1\xspace}

\newcommand{\colmodel}{Model}
\newcommand{\coldataset}{Dataset}
\newcommand{\colwindow}{Window}

\newcommand{\colbetweennessMedaeMean}{\wrapmedian{\wrapmean{BC}}}

\newcommand{\colclosenessMedaeMean}{\wrapmedian{\wrapmean{CLO}}}

\newcommand{\coldegreeMedaeMean}{\wrapmedian{\wrapmean{ND}}}

\newcommand{\coleventsPerSecond}{Events/s}

\newcommand{\colfeatureMmdMean}{MMD}

\newcommand{\colfeatureWassersteinMean}{Wasserstein}

\newcommand{\colpairjsmean}{Pairwise JS}

\newcommand{\colperfeatjsmean}{Per-feature JS}

\newcommand{\coljlSimilarity}{JL}

\newcommand{\collccMedae}{\wrapmedian{LCC}}

\newcommand{\colnComponentsMedae}{\wrapmedian{NCC}}

\newcommand{\colplmedae}{\wrapmedian{PLE}}

\newcommand{\colrelEdgeDistrEntropyMedae}{\wrapmedian{REL}}

\newcommand{\colwedgeCountMedae}{\wrapmedian{WC}}

\newcommand{\colskiprate}{Skip rate}

\usepackage{amsthm} %
\usepackage{hyperref}
\usepackage[noabbrev,capitalise]{cleveref}

\usepackage{url}
\usepackage{xspace}
\usepackage{amssymb} %
\usepackage{float}
\usepackage{algorithm}
\usepackage{algpseudocode}
\usepackage{colortbl}
\usepackage{booktabs}
\usepackage{array}
\usepackage{siunitx}
\usepackage{arydshln} %
\usepackage{multirow}
\usepackage{graphicx} %
\newlength{\speedfigheight} %
\newcolumntype{Z}{>{\setbox0=\hbox\bgroup}c<{\egroup}}
\edef\tmpcoltype{\noexpand\newcolumntype{F}{S[\featprec]}}\tmpcoltype
\edef\tmpcoltype{\noexpand\newcolumntype{T}{S[\topoprec]}}\tmpcoltype
\edef\tmpcoltype{\noexpand\newcolumntype{J}{S[\jlprec]}}\tmpcoltype
\usepackage{xfp}
\usepackage{collcell}
\usepackage{etoolbox}
\newcommand{\skiprate}[1]{\ifstrequal{#1}{{-}}{-}{\num[round-mode=places,round-precision=1]{\fpeval{100*(#1)}}\%}}
\newcolumntype{P}{>{\collectcell\skiprate}l<{\endcollectcell}}
\usepackage{standalone}
\usepackage{tikz}
\usepackage{pgf}
\usepackage{xcolor}
\usetikzlibrary{arrows.meta,fit,calc,patterns}

\definecolor{srcgreen}{HTML}{B2F2BB} %
\definecolor{dstred}{HTML}{FFC9C9}   %
\definecolor{ink}{HTML}{1E1E1E}
\definecolor{dioBlue}{HTML}{DAE8FC}
\definecolor{dioOrange}{HTML}{FFE6CC}
\definecolor{dioPurple}{HTML}{E1D5E7}
\definecolor{beige}{HTML}{F3EAD3}

\ifdefined\zcur\else

\fi

\tikzset{dyl/.style={inner sep=0pt, outer sep=0pt}}
\ifdefined\zcur\else

\fi

\definecolor{oursValidDot}{HTML}{E34948}
\definecolor{dggenValidDot}{HTML}{4A3AA7}
\definecolor{tiggerValidDot}{HTML}{008300}
\definecolor{oursWithinDot}{HTML}{F2ADAD}
\definecolor{dggenWithinDot}{HTML}{AEA6D7}
\definecolor{tiggerWithinDot}{HTML}{8CC78C}
\newcommand{\capdot}[1]{\begin{tikzpicture}[baseline=-0.6ex]\filldraw[fill=#1, draw=black, line width=0.4pt] (0,0) circle (2.2pt);\end{tikzpicture}}
\usepackage{datetime}
\newdateformat{compiledate}{\twodigit{\THEDAY}/\twodigit{\THEMONTH}/\THEYEAR}
\usepackage{eso-pic}

\makeatletter
\DeclareRobustCommand\onedot{\futurelet\@let@token\@onedot}
\def\@onedot{\ifx\@let@token.\else.\null\fi\xspace}
\makeatother

\def\eg{\emph{e.g}\onedot} 
\def\ie{\emph{i.e}\onedot}

\title{DyGDi: Diffusion Model for Continuous-Time Dynamic Graph Generation}

\author{\footnotesize Tsiry Mayet$^{1,*}$, No\'{e} Gille$^{2}$, Simon Bernard$^{1}$, Maxime Berar$^{1}$, Cl\'{e}ment Chatelain$^{2}$, S\'{e}bastien Adam$^{1}$ \\[4pt]
\footnotesize $^{1}$Univ Rouen Normandie, LITIS UR 4108, F-76000 Rouen, France \\
\footnotesize $^{2}$INSA Rouen Normandie, LITIS UR 4108, F-76000 Rouen, France
}

\iclrfinalcopy %
\begin{document}

\maketitle
\lhead{Preprint} %
{\renewcommand{\thefootnote}{*}\footnotetext{Corresponding author: \texttt{tsiry.mayet@gmail.com}}}

\begin{abstract}
Dynamic graph generation has emerged as a powerful paradigm for modeling evolving structural phenomena.
By synthesizing a stream of events, it can capture real-world network evolution over time.
However, generation remains limited and slow as dynamic graphs grow in both size and feature dimensionality, limiting practical use.
We introduce \modelfullname (\modelname), a new dynamic graph generation model that leverages diffusion to jointly model a window of events and their features.
\modelname generates the entire window in a single pass.
This yields faster generation and lets events and features attend to one another, allowing the model to learn more coherent dynamics.
To address a node-identity issue arising from this parallel generation, we introduce a new event tokenization that injects more information about event appearance,
anchoring newly generated nodes to their identities.
Altogether, we show that \modelname trains and generates more efficiently than previous models and scales better.
It produces graphs with high-quality node-feature and topological properties.
Code \textit{will be provided} on GitHub.
\end{abstract}

\section{Introduction}

Many real-world systems are naturally described as dynamic graphs, such as social networks, citation graphs, financial transactions, and traffic-control systems \citep{ji_systematic_2025,yang2024dynamic,8697151}.
Synthetically generating such data streams is challenging, as the generated content must respect the temporal structure and topology of the original data without simply replicating it \citep{guptaTIGGERScalableGenerative2022}.
Prior models face scaling challenges rooted in their adjacency-matrix representation, which is unsuitable for large graphs \citep{guptaTIGGERScalableGenerative2022}.
Beyond the topological structure, a model must also generate realistic, evolving features for nodes and edges.
Finally, it must generalize inductively to nodes and timestamps unseen during training.
Generating such data has applications in predicting graph evolution over time, anomaly detection, social-network simulation, and recommendation systems \citep{yang2024dynamic}.
Given a dynamic graph, we therefore consider the task of inductively generating sequences of unseen events and nodes consistent with the original data dynamics.

Existing works represent a dynamic graph in one of two main ways \citep{yang2024dynamic}, either as a \textit{discrete-time} dynamic graph (DTDG) or as a \textit{continuous-time} dynamic graph (CTDG).
DTDG generative models view a graph as a sequence of discrete static graphs, either by aggregating events into a list of graph snapshots or by building a single static graph in which recurring nodes are linked by temporal edges.
Both representations enable the reuse of static-graph generation methods, but discrete snapshots suffer a loss of temporal resolution from the discretization process, and the temporal-edge construction scales poorly to large numbers of timestamps due to the size of the resulting graph.
CTDG generative models instead view a graph as a sequence of events, each occurring at a single timestamp, where an event may be an edge addition or deletion, a node addition or deletion, or a transformation of node or edge features.
Such methods have been shown to scale to large graphs better than discrete-time models \citep{hosseiniDeepProbabilisticFramework2025}.
However, existing inductive CTDG generators are autoregressive, sequential event sampling makes generation slow, and accelerating it by sampling batches of events in parallel generates the events of a batch independently, sacrificing their coherence.

Diffusion models, which originated in image generation \citep{sohl2015deep,ho2020denoising}, mark a major advance in generative modelling and have since been extended to other modalities such as video \citep{ho2022video} and text \citep{arriolaBlockDiffusionInterpolating2024}.
They are particularly attractive for their conditioning flexibility \citep{mayet2025multiple,wewer2025spatial}.
They operate by corrupting data through a forward process and learning to reverse it back to the original distribution.
Recent work has adapted them to static-graph data, for example in molecule design \citep{vignacDiGressDiscreteDenoising2023,jo2022score,hoogeboom2022equivariant}, community-network and large-scale graph generation \citep{niu2020permutation,jo2022score,limnios2024sagess,bernecker2025random,li2024graphmaker}, and protein design \citep{trippediffusion}.
Despite these promising results, such methods face two limitations: they remain confined to static-graph generation, and those that model the full adjacency matrix cannot scale to graphs with large numbers of nodes or timestamps.

We therefore propose \modelfullname (\modelname), the first diffusion-based generative model for CTDGs.
It generates an entire window of events simultaneously, with node embeddings derived from a memory bank encoder \citep{rossiTemporalGraphNetworks2020}.
Because this formulation does not rely on fixed node identities, it supports inductive generation of unseen nodes and timestamps.
To support inductive generation of new nodes, we introduce event-class information and a random embedding for new nodes.
The resulting model trains stably across window sizes, alleviating the scalability and memory issues of prior adjacency-matrix models, and can generate sequences of arbitrary length.
\noindent\textbf{Our main contributions are summarized as follows:}\\
   \noindent\textbullet \ 
   We propose the first diffusion-based generative model for CTDGs, generating windows of events simultaneously conditioned on past events.
   \\ 
   \noindent\textbullet \ 
   We identify and formalize \memcollapse and \wincollapse in inductive generation and resolve them, enabling inductive generation of new nodes.
   \\
   \noindent\textbullet \
   We conduct extensive experiments showing that \modelname achieves state-of-the-art dynamic graph generation, in both topology and feature distribution, and substantially faster training and generation than previous models.

\section{Preliminaries}
\textbf{Continuous-Time Dynamic Graphs.}
A static graph $\gG = (\gV, \gE)$ consists of $n$ vertices $\gV = \{1, \dots, n\}$ and edges $\gE \subseteq \gV \times \gV$.
Continuous-time dynamic graphs (CTDGs) extend this formalism to graphs that evolve over time. Rather than a sequence of discrete snapshots, a CTDG is an ordered sequence of time-stamped events:
\begin{equation}
\label{eq:def:graph_c}
   \gG = \{ x(t_1), x(t_2), \dots \}
\end{equation}
where each event $x(t_i)$, occurring at time $t_i$, may be an edge or node addition or deletion, or a transformation of node or edge features.
Each event is characterised by
\begin{equation}
\label{eq:def:event}
   x(t_i) = \left( \src, \dst, t_i, \efeat_{\src, \dst}(t_i) \right)
\end{equation}
with $\src, \dst \in \gV$ and $t_i \in \R$, where the edge feature $\efeat_{\src, \dst}(t_i)$ encodes attributes of the directed edge from $\src$ to $\dst$, such as edge weights, types, or other relational metadata.

\textbf{Diffusion Models and Flow Matching.} Diffusion models learn a target data distribution $p_1$ by reversing a noising process. A forward process progressively corrupts a data point $x_1 \sim p_1$ towards a source distribution $p_0$ (typically Gaussian noise, $x_0 \sim \mathcal{N}(0, \mI)$), yielding intermediate latent variables $x_\tdiff \sim q(x_\tdiff \mid x_1)$. Since $t$ is reserved for graph time, we denote the diffusion timestep by $\tdiff$.
Rather than parameterizing the reverse process through noise or score prediction \citep{ho2020denoising,song2021scorebased}, we adopt flow matching \citep{lipman2023flow}, which learns a continuous velocity field $v_\theta$ transporting samples from the source distribution $p_0$ to the data distribution $p_1$ via an Ordinary Differential Equation (ODE):
\begin{equation}
\label{eq:ode}
   \dot{x} = v_\theta(x, \tdiff) \quad\text{with target}\quad v^*(x_\tdiff, \tdiff) = x_1 - x_0.
\end{equation}
The forward process interpolates linearly between noise and data:
\begin{equation}
\label{eq:forward}
   x_\tdiff = (1-\tdiff)\, x_0 + \tdiff\, x_1, \quad \tdiff \sim \mathcal{U}(0, 1).
\end{equation}
The network $v_\theta$ is trained to match the target velocity field with a mean-squared-error objective:
\begin{equation}
\label{eq:loss:fm}
   \Ls_\text{FM}(\theta) = \E_{
      x_0 \sim p_0,\,
      x_1 \sim p_1,\,
      \tdiff \sim \mathcal{U}(0, 1)
   }
   \left\|
   v_\theta \left(x_\tdiff, \tdiff \right)
   - v^*\left(x_\tdiff, \tdiff\right)
   \right\|_2^2.
\end{equation}
At generation time, new samples are obtained by integrating the learned ODE in \cref{eq:ode} from $x_0 \sim p_0$ to $x_1 \sim p_1$.

\textbf{Conditional Generation.} Generation can be conditioned on auxiliary information $c$ (\eg context, class, or history), by extending the velocity field and its target to $v_\theta(x_\tdiff, \tdiff \mid c)$ and $v^*(x_\tdiff, \tdiff \mid c) = x_1 - x_0$. The training objective in \cref{eq:loss:fm} then becomes
\begin{equation}
\label{eq:loss:fm:cond}
   \Ls_\text{FM}(\theta) = \E_{
      x_0 \sim p_0,\,
      x_1 \sim p_1,\,
      \tdiff \sim \mathcal{U}(0, 1)
   }
   \left\|
   v_\theta \left(x_\tdiff, \tdiff \mid c \right)
   - v^*\left(x_\tdiff, \tdiff\right)
   \right\|_2^2,
\end{equation}
and samples are generated for a fixed condition $c$ by integrating $\dot{x} = v_\theta(x, \tdiff \mid c)$ from $x_0 \sim p_0$ to $x_1 \sim p_1$.

\section{The Proposed Event Diffusion Model}
\begin{figure}[t]
   \centering
   \begin{tikzpicture}[
    font=\sffamily,
    >={Stealth[length=9,width=7]},
    box/.style={draw=ink, line width=1pt, rounded corners=3, inner sep=0pt},
    eventbox/.style={box, dash pattern=on 4pt off 4pt},
    title/.style={font=\sffamily\footnotesize\bfseries, scale=1.5},
    event/.style={circle, draw=ink, line width=1pt, minimum size=6mm,
                  inner sep=0pt, font=\normalsize},
    tokcell/.style={draw=ink, line width=0.6pt, rounded corners=0.5pt,
                    inner sep=0pt},
    edge/.style={draw=ink, line width=1pt},
    flow/.style={->, draw=ink, line width=0.8},
    timeline/.style={->, draw=ink, line width=0.8},
    lbl/.style={font=\sffamily\footnotesize, inner sep=1pt, scale=1.5},
    evlbl/.style={font=\sffamily\normalsize, inner sep=2pt, scale=1.5}, %
    rowlbl/.style={font=\sffamily\footnotesize, scale=1.8, anchor=south west,
                   inner sep=2pt},
]

\def\evPairDx{1.015} %
\def\evRise{0.797}   %
\def\axisY{6.119}    %
\def\evTick{0.131}   %
\def\fdBulge{1.233}  %

\def\tokWh{0.375}    %
\def\tokH{1.5}       %
\def\tokHh{0.75}     %
\def\tokGap{0.07}    %
\def\tokPad{0.08}    %

\newcommand{\eventinbox}[9]{%
  \pgfmathsetmacro\f{(#2-0.5)/#3}%
  \pgfmathsetmacro\ox{\evPairDx*cos(#4)-\evRise*sin(#4)}%
  \pgfmathsetmacro\oy{\evPairDx*sin(#4)+\evRise*cos(#4)}%
  \pgfmathsetmacro\hox{-\ox/2}%
  \pgfmathsetmacro\hoy{-\oy/2}%
  \coordinate (src) at ($(#1.west)!\f!(#1.east)+(\hox,\hoy)$);%
  \coordinate (dst) at ($(src)+(\ox,\oy)$);%
  \draw[edge] (src) -- (dst);%
  \node[evlbl, anchor=west, yshift=-3pt, scale=0.7] at ($(src)!0.5!(dst)$) {#9};%
  \node[event, fill=#5] at (src) {#7};%
  \node[event, fill=#6] at (dst) {#8};%
  \draw[edge] ($(src |- axisL)+(0,-\evTick)$) -- ($(src |- axisL)+(0,\evTick)$);%
}
\def\noisemode{noise}
\def\emptymode{cleanempty}
\newcommand{\stacktok}[2]{%
  \pgfmathsetmacro\inW{\tokWh-\tokPad}%
  \pgfmathsetmacro\stackH{\tokH-2*\tokPad}%
  \pgfmathsetmacro\cellH{(\stackH-4*\tokGap)/5}%
  \def\tmode{#2}%
  \draw[draw=ink, line width=1pt, rounded corners=1.5]
        ($(#1)+(-\tokWh,\tokHh)$) rectangle ($(#1)+(\tokWh,-\tokHh)$);%
  \foreach \i in {0,...,4} {%
    \pgfmathsetmacro\ytop{\stackH/2-\i*(\cellH+\tokGap)}%
    \pgfmathsetmacro\ybot{\ytop-\cellH}%
    \ifcase\i\relax\def\base{dioBlue}\or\def\base{srcgreen}\or\def\base{dstred}%
      \or\def\base{dioOrange}\else\def\base{dioPurple}\fi%
    \def\drawcell{1}%
    \ifx\tmode\emptymode\ifnum\i=0\relax\def\drawcell{0}\fi\fi%
    \ifnum\drawcell=1\relax
      \ifx\tmode\noisemode
        \draw[tokcell, fill=\base!45!gray]
              ($(#1)+(-\inW,\ytop)$) rectangle ($(#1)+(\inW,\ybot)$);%
        \draw[tokcell, pattern=north east lines, pattern color=ink!70]
              ($(#1)+(-\inW,\ytop)$) rectangle ($(#1)+(\inW,\ybot)$);%
      \else
        \draw[tokcell, fill=\base]
              ($(#1)+(-\inW,\ytop)$) rectangle ($(#1)+(\inW,\ybot)$);%
      \fi
    \fi
  }%
}
\newcommand{\boxrow}[3]{%
  \foreach \k in {1,...,#2} {%
    \pgfmathsetmacro\f{(\k-0.5)/#2}%
    \coordinate (tc) at ($(#1.west)!\f!(#1.east)$);%
    \stacktok{tc}{#3}%
  }%
}
\newcommand{\nodetok}[2]{%
  \pgfmathsetmacro\inW{\tokWh-\tokPad}%
  \draw[tokcell, fill=#2]
        ($(#1)+(-\inW,0.25)$) rectangle ($(#1)+(\inW,-0.25)$);%
}

\node[eventbox, fit={(-24.143,8.482) (-14.892,6.395)}] (clusterPast) {};
\node[eventbox, fit={(-14.762,8.482) (-7.977,6.395)}]  (clusterCur)  {};
\node[lbl, scale=1.2, anchor=south] at (clusterPast.north) {past events};
\node[lbl, scale=1.2, anchor=south] at (clusterCur.north)  {current events};
\node[title, align=center, rotate=90, scale=0.8] at (-24.3,3.75)
     {EVENT TOKENIZATION\\[1pt]{\scriptsize(\S\ref{sec:tokenization})}};
\node[box, fit={(-23.56,5.25) (-17.06,3.1)}]           (rowPast)  {};
\node[box, fit={(-23.56,2.55) (-17.06,1.55)}]          (rowSum)   {};
\node[box, fit={(-15.285,5.25) (-10.41,3.1)}]          (rowCur)   {};
\node[box, fit={(-15.285,2.575) (-10.41,0.4)}]         (rowNoisy) {};
\node[box, fill=beige, fit={(-15.812,-0.239) (-9.882,-1.529)}] (model) {};
\node[title] at (model) {DIFFUSION MODEL\ {\large$v_\theta$}};
\node[box, fit={(-15.285,-2.167) (-10.41,-4.342)}]     (rowHat)   {};
\node[box, fit={(-7.482,0.878) (-4.538,-0.63)}]        (mse) {};
\node[title, align=center] at (mse) {Loss\\[1pt]{\scriptsize(Eq.~\ref{eq:loss:window})}};

\coordinate (axisL)   at (-25.578,\axisY);  %
\coordinate (axisR)   at (-6.54,\axisY);    %

\eventinbox{clusterPast}{1}{4}{0}{srcgreen}{dstred}{}{}{}
\eventinbox{clusterPast}{2}{4}{0}{srcgreen}{dstred}{}{}{}
\eventinbox{clusterPast}{3}{4}{0}{srcgreen}{dstred}{}{}{}
\eventinbox{clusterPast}{4}{4}{0}{srcgreen}{dstred}{}{}{}
\eventinbox{clusterCur}{1}{3}{0}{srcgreen}{dstred}{}{}{}
\eventinbox{clusterCur}{2}{3}{0}{srcgreen}{dstred}{$\src$}{$\dst$}{$e_{\src,\dst}(t_i)$}
\eventinbox{clusterCur}{3}{3}{0}{srcgreen}{dstred}{}{}{}
\coordinate (tiCell) at ($(clusterCur.west)!0.5!(clusterCur.east)$);
\node[evlbl, anchor=north] at ($(tiCell |- axisL)+(-0.508,-0.203)$) {$t_i$};

\boxrow{rowPast}{4}{cleanempty}                                 %
\foreach \k/\col in {1/srcgreen,2/dstred,3/srcgreen,4/dstred} {%
  \pgfmathsetmacro\f{(\k-0.5)/4}%
  \coordinate (nc) at ($(rowSum.west)!\f!(rowSum.east)$);%
  \nodetok{nc}{\col}%
}
\boxrow{rowCur}{3}{clean}                                       %
\boxrow{rowNoisy}{3}{noise}                                     %
\boxrow{rowHat}{3}{clean}                                       %
\coordinate (zt) at ($(rowCur.west)!0.5!(rowCur.east)$);
\node[lbl, rotate=90] at ($(zt)+(-\tokWh-0.25,0)$) {$z(t_i)$};
\node[rowlbl] at (rowPast.north west)  {$\zpast$};
\node[rowlbl] at (rowSum.north west)   {$\zsum$};
\node[rowlbl] at (rowCur.north west)   {$\zcur$};
\node[rowlbl] at (rowNoisy.north west) {$\znoisy$};
\node[rowlbl] at (rowHat.north west)   {$\zcurhat$};
\node[lbl, anchor=west, align=left] at
  ($(rowCur.east)!0.5!(rowNoisy.east)+(\fdBulge,0)$)
  {forward\\diffusion\\{\scriptsize(Eq.~\ref{eq:forward})}};

\draw[timeline] (axisL) -- (axisR);                                   %
\node[lbl, anchor=north west] at (-7.439,5.612) {time};
\draw[flow] (clusterPast.south -| rowPast.north) -- (rowPast.north);  %
\draw[flow] (clusterCur.south  -| rowCur.north)  -- (rowCur.north);   %
\draw[flow] (rowCur.east) .. controls
      ($(rowCur.east)!0.5!(rowNoisy.east)+(\fdBulge,0)$) .. (rowNoisy.east);
\draw[edge] (rowPast.east) -- (-16.4,4.175);
\draw[edge] (rowSum.east)  -- (-16.4,2.05);
\fill[ink] (-16.4,2.05) circle (1.5pt);
\draw[flow] (-16.4,4.175)  |- (model.west);
\node[lbl, fill=white, inner sep=2pt] at (-16.9,0) {context};
\draw[flow] (model.north |- rowNoisy.south) -- (model.north);        %
\draw[flow] (model.south) -- (model.south |- rowHat.north);          %
\draw[flow, densely dotted] (rowCur.east) -| (mse.north);
\draw[flow, densely dotted] (rowHat.east) -| (mse.south);

\def\lcx{-24.5}      %
\def\lcy{-2.8}       %
\def\lWh{0.55}       %
\def\lHh{1.5}        %
\def\lPad{0.1}
\def\lGap{0.12}
\pgfmathsetmacro\lInW{\lWh-\lPad}
\pgfmathsetmacro\lStackH{2*\lHh-2*\lPad}
\pgfmathsetmacro\lCellH{(\lStackH-4*\lGap)/5}
\draw[draw=ink, line width=1pt, rounded corners=2]
     (\lcx-\lWh,\lcy+\lHh) rectangle (\lcx+\lWh,\lcy-\lHh);
\foreach \i/\col/\txt in {%
   0/dioBlue/{event information},
   1/srcgreen/{$E(\src)$},
   2/dstred/{$E(\dst)$},
   3/dioOrange/{$\Delta t_i$},
   4/dioPurple/{$e_{\src,\dst}(t_i)$}} {%
  \pgfmathsetmacro\cyc{\lcy+\lStackH/2-\i*(\lCellH+\lGap)-\lCellH/2}%
  \draw[tokcell, fill=\col]
       (\lcx-\lInW,\cyc+\lCellH/2) rectangle (\lcx+\lInW,\cyc-\lCellH/2);%
  \node[lbl, anchor=west] at (\lcx+\lWh+0.15,\cyc) {\txt};%
}
\node[lbl, anchor=south west] at (\lcx-\lWh,\lcy+\lHh+0.08)
     {$z(t_i)$~(Eq.~\ref{eq:tokenization})};

\end{tikzpicture}
   \caption{
      \textbf{\modelfullname (\modelname) training process.}
      The graph is represented as a sequence of events and tokenized by adding event information, with each node identity encoded by a continuous graph encoder.
      A window of past events serves as context, while the current window of events is noised and predicted.
      The diffusion model generates multiple events and their features simultaneously.
   }
   \label{fig:model:overview}
\end{figure}
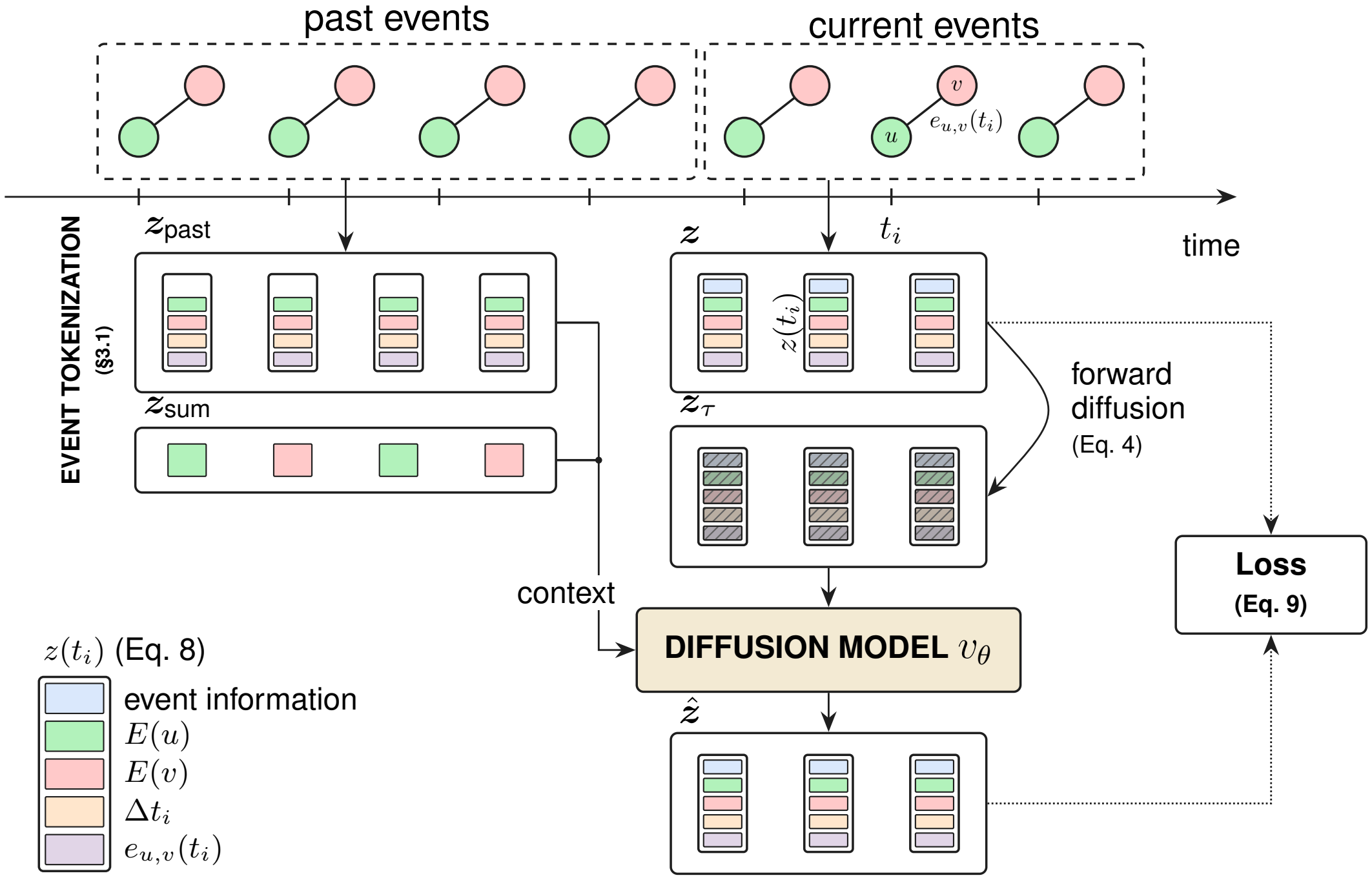

In \modelfullname (\modelname), the underlying problem of dynamic graph generation is cast as a sequence-to-sequence problem.
Consequently, events are treated as a sequence of tokens, and the diffusion process operates directly on this token sequence.
The diffusion model is conditioned on past event embeddings and denoises a window of current event embeddings.
The embeddings are produced by a memory module that learns a mapping from node identifiers to latent vectors, and the inverse mapping from an embedding back to a node identifier is performed at generation time via a nearest-neighbour search.
An overview of \modelname is provided in \cref{fig:model:overview}.

\subsection{Dynamic Graph Generation as a Sequence-to-Sequence Problem}

\textbf{Event Tokenization.}
\label{sec:tokenization}
To enable the diffusion process to operate on a CTDG, each event is mapped to a fixed-dimensional token.
For every node $i \in \gV$, we obtain a latent representation through an embedding function $\getemb : \gV \rightarrow \R^{d_n}$ implemented by a memory module (a Temporal Graph Network, or TGN~\citep{rossiTemporalGraphNetworks2020}, in our instantiation), where $d_n$ denotes the node embedding dimension.
The memory module maintains a fixed-size bank of $N$ slots $\mM = \{ \vs_1, \dots, \vs_N \} \subset \R^{d_n}$, where $N$ is a hyperparameter that upper-bounds the number of nodes the model can track.
Each slot is associated with at most one node identifier, and slots that have not yet been bound to an observed identifier remain at their zero initialisation.
Denoting by $\seen \subseteq \gV$ the set of node identifiers observed in at least one event prior to the current tokenization step, we define
\begin{equation}
\label{eq:getemb}
   \getemb(i) = \vs_i \text{ if } i \in \seen\text{, and } \vzero \text{ otherwise.}
\end{equation}

Given an event $x(t_k) = (\src, \dst, t_k, \efeat_{\src,\dst}(t_k))$ with source and destination identifiers $\src, \dst \in \gV$, its token is obtained by concatenating the source and destination node embeddings, the inter-event interval $\Delta t_k = t_k - t_{k-1}$, and the edge feature:
\begin{equation}
\label{eq:tokenization}
   z(t_k) = \concat\left[ \getemb(\src), \getemb(\dst), \Delta t_k, \efeat_{\src,\dst}(t_k) \right] \in \R^{d}.
\end{equation}
A CTDG is thereby mapped to an ordered sequence of tokens $(z(t_1), z(t_2), \dots)$.

\textbf{Predicting a Window of Events.}
\label{sec:predicting_window}
At both training and inference time, \modelname generates a sliding window of $L$ current event tokens $\zcur = (z(t_{k+1}), \dots, z(t_{k+L}))$, conditioned on a context of $M$ past tokens $\zpast = (z(t_{k-M+1}), \dots, z(t_{k}))$ that summarises the recent history of the graph.
In addition, \modelname is provided with a graph-level summary of the CTDG, $\zsum = \{\getemb(i) : i \in \seen\}$, the embedding of every seen node (i.e.\ the raw memory state maintained by the TGN memory module), summarising that node's history up to $t_k$, although a recurrent network maintaining a hidden state may be used.
The model thus learns the joint conditional distribution $p(\zcur \mid \zpast, \zsum)$, from which the next event tokens are sampled.

Following the conditional flow-matching framework of \cref{eq:loss:fm:cond} with condition $c = (\zpast, \zsum)$, the velocity network $v_\theta$ is optimised with the loss:
\begin{equation}
\label{eq:loss:window}
   \Ls(\theta) = \E_{
      z_0 \sim p_0,\,
      z_1 \sim p_1,\,
      \tdiff \sim \mathcal{U}(0, 1)
   }
   \left\|
   v_\theta \left(z_\tdiff, \tdiff  | \zpast, \zsum \right)
   - v^*\left(z_\tdiff, \tdiff\right)
   \right\|_2^2.
\end{equation}

\subsection{Generation Process}
Generating a window of events with \modelname proceeds in three steps, (i) sampling a window of tokens from the conditional distribution $p(\zcur \mid \zpast, \zsum)$, (ii) matching each token embedding either to an existing node identifier or to a new one, and (iii) updating the memory module accordingly and iteratively repeating.
Specifically, let $\zcur \sim q(\zcur \mid \zpast, \zsum)$ be a generated window of $L$ tokens, and let $\hat{s}$ denote a node embedding extracted from token $z(t_k)$. The associated node identifier is sampled from a categorical distribution over $\gV$, $\hat{id} \sim p_{\text{match}}(\cdot \mid \hat{s})$, where the probability of selecting node $i$ is
\begin{equation}
\label{eq:generation:node_matching}
   p_{\text{match}}(i \mid \hat{s}) = \frac{e^{d(\hat{s},\, \getemb(i))/\temp}}
                       {\sum_{j \in \gV} e^{d(\hat{s},\, \getemb(j))/\temp}},
\end{equation}
with $d(\cdot , \cdot)$ a similarity score (we use cosine similarity) and $\temp$ a temperature parameter.

\textbf{New-Embedding Collapse.}
However, the formulation in \cref{eq:generation:node_matching} becomes problematic when node identifiers must be sampled from $\unseen$, the set of previously unseen nodes.
This induces two failure modes, one on each side of the matching score $d(\hat{s},\,\getemb(i))$, namely \emph{\memcollapse}, a collapse of the match targets $\getemb(i)$, and \emph{\wincollapse}, a collapse of the match queries $\hat{s}$.
\\
\textbullet~\textit{\Memcollapse}: For $i, j \in \unseen$ with $i \ne j$, the memory embeddings collapse, $\getemb(i) = \getemb(j) = \vzero$ (\cref{eq:getemb}), so the match targets are indistinguishable, and the same node recurring in the window may be mapped to different identifiers.
\\
\textbullet~\textit{\Wincollapse}: For $i, j \in \unseen$ with $i \ne j$, the generated embeddings collapse, $s_i = s_j$, so the match queries are indistinguishable, and different nodes may be mapped to the same identifier.

\textbf{Event-Class Information and Random Embeddings for New Nodes.}
To alleviate the aforementioned issues, we introduce \emph{\eventclass} attached to each token and initialise, during training, every unseen node in a window with a random Gaussian embedding.
To track node identity across the window, for each node participating in an event we associate a one-hot class vector $\vc$ indicating whether the node has already been seen, is a new node appearing for the first time in the window, or is a new node already seen earlier in the window.
The tokenization in \cref{eq:tokenization} thus becomes:
\begin{equation}
\label{eq:tokenization:with_class}
   z(t_k) = \concat\left[ \vc_{\src}, \vc_{\dst}, \getemb(\src), \getemb(\dst), \Delta t_k, \efeat_{\src,\dst}(t_k) \right] \in \R^{d}.
\end{equation}
This decomposition allows nodes to be sampled iteratively from the appropriate pool: a node assigned to the \emph{seen} class is sampled from $\seen$, a node assigned to the \emph{new} class is sampled from $\unseen$, and a new node already observed within the current window is sampled from the pool of previously matched new nodes.

Distinct embeddings remove the representational collapse underlying both. During training each new node receives a random Gaussian draw, giving distinct new nodes distinct memory targets, and hence distinct generated queries:
\begin{equation}
   \getemb(i) = \vs_i \text{ if } i \in \seen\text{, and } \xi_i \sim \mathcal{N}(0, \mI) \text{ otherwise.}
\end{equation}
These distinct targets resolve \wincollapse and give the class-vector routing the separation to re-identify a recurring new node, which resolves \memcollapse.
Both the class vectors and the random embeddings serve as input during training. At generation time, the class vectors are instead generated by the model and route the iterative sampling described above.
The contribution of these components, together with the conditioning signals $\zsum$ and $\zpast$, is examined in the ablation study of \cref{ablation}.

\section{Related Work}
\label{sec:related}
Dynamic graph generative models operate in either a transductive or an inductive setting; we review both, then prior couplings of diffusion models with graphs for signal generation. \\
\textbf{Transductive dynamic graph generation.}
Transductive models generate interactions only among existing nodes. They cannot introduce new nodes and often replicate the training interactions \citep{guptaTIGGERScalableGenerative2022}.
TagGen \citep{zhou2020data} builds a static graph with timestamped edges, samples synthetic random walks, keeps those a classifier judges realistic, and aggregates them into the dynamic graph.
Temporal edges restrict generation to known timesteps, and it supports neither node features nor new identities.
\citet{guptaTIGGERScalableGenerative2022} report that it regenerates 80\% of the training edges and, by modelling an adjacency matrix, captures only short-range dependencies, limiting it to roughly 200 timestamps.
Dymond \citep{zenoDYMONDDYnamicMOtifNoDes2021} models 3-node motifs that arrive at a constant rate and persist once formed.
Unlike TagGen, it does not replicate the training data but instead deviates from the source graph's statistics, while remaining confined to a similarly short horizon \citep{hosseiniDeepProbabilisticFramework2025}. \\
\textbf{Inductive dynamic graph generation.}
These models additionally introduce new nodes, enabling operation on unseen graphs while preventing leakage of source-graph node identities.
\tigger \citep{guptaTIGGERScalableGenerative2022} autoregressively models random walks with an RNN and merges them into the synthetic graph.
\tiggerind extends this to the inductive setting by reducing the continuous graph to a static one and applying GraphSAGE \citep{hamilton2017inductive} and a WGAN \citep{arjovsky2017wasserstein}.
The static reduction hampers scaling to large graphs, and the node-time-sequence representation prevents the use of node and edge features.
Furthermore, random-walk sampling and merging captures walk-level statistics but prevents precise conditioning on individual temporal interactions.
\dggen \citep{hosseiniDeepProbabilisticFramework2025} pairs a TGN~\citep{rossiTemporalGraphNetworks2020} encoder with a small decoder that factorises the interaction probability over edge components, enabling large-scale inductive autoregressive generation with node and edge features and arbitrary timestamps.
From the TGN node embeddings, the decoder samples, for each event, a source node, then a destination conditioned on the source, and finally the time and edge features via an RNN conditioned on the source--destination edge.
However, generation remains autoregressive and therefore slow. To accelerate it, \dggen samples batches of events, where each event in a batch is sampled independently of the others, sacrificing the coherence that sequential sampling provided. \\
\textbf{Other diffusion models on graphs and signal generation.}
Prior work on signal generation has coupled diffusion models with graphs, generating signals on a graph rather than the graph itself.
DiffSTG \citep{wen2023diffstg} performs probabilistic forecasting of node signals on a spatio-temporal graph, \ie a fixed sensor graph (\eg traffic, air quality) whose node values evolve over time. The topology is a static input, and only signal values are generated, which amounts to predicting new node values on a static graph.
TSGDiff \citep{shen2026tsgdiff} generates time series through a representation it calls ``dynamic graphs'', but these are a deterministic intermediate encoding. Each signal window becomes a graph whose nodes are time steps and whose edges link consecutive and Fourier-periodic neighbours. A VAE-style GNN embeds this graph, an MLP diffusion models the latent, and the decoder reconstructs only the series values. The adjacency is never decoded or sampled.
Both approaches contrast with our task, where the generated object is the graph itself: a continuous-time stream of interaction events, with node identities and inter-event times part of the sampled distribution.
To the best of our knowledge, this work is the first diffusion-based generative model for CTDGs.

\section{Results and Discussion}
This section shows that \modelname produces higher-quality features than existing graph generative models (\cref{feature_quality_twelve}); generates higher-quality topologies (\cref{feature_quality_twelve}); scales to more feature dimensions (\cref{scaling_to_more_features}), with fast training (\cref{trainingspeed}) and generation (\cref{speed_vs_other}); and remains efficient in speed and memory across generation window and context sizes (\cref{results:generation_speed:diff_conf,results:vram_consumption}).
\rebuttal{A controlled experiment on a synthetic dataset (\cref{appendix:synthetic}) further shows the advantage of jointly generating the events of a window over the existing paradigms, which generate events independently (\cref{appendix:synthetic:results,appendix:synthetic:indep}), and allows a careful ablation of the components of \modelname (\cref{appendix:synthetic:ablation}).}

\subsection{Experimental Methodology}
\textbf{Baselines and Datasets.}
For the \textit{generative task}, \modelname is evaluated against the state-of-the-art inductive CTDG generative models \tigger~\citep{guptaTIGGERScalableGenerative2022} and \dggen~\citep{hosseiniDeepProbabilisticFramework2025}, both detailed in \cref{sec:related}.
Recent studies~\citep{hosseini_quality_2024,guptaTIGGERScalableGenerative2022,rossiTemporalGraphNetworks2020} report that very few approaches can perform inductive CTDG generation, and to the best of our knowledge only \dggen can perform inductive generation jointly with message features.
All methods are evaluated on standard benchmarks for CTDG generation, namely the Reddit, Wikipedia, LastFM, and MOOC datasets~\citep{kumar2019predicting}.
\rebuttal{Generation is additionally evaluated on the Enron, UCI, SocialEvo, and Contacts datasets of the DyGLib benchmark~\citep{poursafaei2022towards,yuDyGFormer2023}.}
See \cref{appendix:impl_details} for dataset statistics and further implementation details.
Following \dggen, and for a fair comparison, all models are first trained and evaluated on the first 12 features of Reddit and Wikipedia. We additionally scale to the full 172-feature set (\cref{scaling_to_more_features}).
\\
\textbf{Evaluation Metrics.}
The quality of the generated graphs is evaluated using established metrics~\citep{hosseiniDeepProbabilisticFramework2025,guptaTIGGERScalableGenerative2022,hosseini_quality_2024}.
\textit{\underline{Feature quality}}:
is evaluated using the Jensen--Shannon (JS) distance, computed between the histograms of the marginal distributions of all interaction features and between the joint histograms of every pair of features.
The Wasserstein distance (WD) and Maximum Mean Discrepancy (MMD), both computed directly on the generated features without binning, are also reported as complementary measures.
\textit{\underline{Topological properties}}:
As in \citet{guptaTIGGERScalableGenerative2022,hosseiniDeepProbabilisticFramework2025}, the generated graphs are evaluated by computing static descriptors on discrete-time graph snapshots at multiple temporal resolutions and comparing them to those of the real graphs; the Median Absolute Error is reported for each descriptor.
For node-level metrics capturing local topology, we report Node Degree (\nodedegree), Closeness Centrality (\clocen), and Betweenness Centrality (\btwc).
For snapshot-level metrics capturing global structure, we use Wedge Count (\wedgecount), Power Law Exponent of the node degree (\powerlaw), Number of Connected Components (\nbcc), Largest Connected Component (\lcc), and Relative Edge Distribution Entropy (\rel).
These metrics probe four complementary axes: \nodedegree measures the graph density. \wedgecount, \powerlaw, and \rel capture the shape of the degree distribution. \clocen and \btwc assess the path structure. \nbcc and \lcc evaluate the connectivity of the generated graphs.
In addition, we report the global Johnson--Lindenstrauss (JL) similarity metric~\citep{hosseini_quality_2024}.
The JL metric applies a random matrix projection to the graph and summarises sensitivity to topology, features, and time within a single scalar value.

\subsection{Generation Quality}
\label{feature_quality_twelve}
\textbf{Feature Quality.}
\Cref{tab:feat_js} reports the \rebuttal{per-feature (1D) and pairwise (2D) JS distances} and the WD for each model. The \rebuttal{MMD is} detailed in \cref{appendix:feat_1d_mmd}.
As \tigger does not produce features, it is omitted from this table.
\rebuttal{The LastFM, Enron, and UCI datasets contain no features and are also omitted.}
\rebuttal{Contacts has a single feature, so it has no pairwise JS distance.}
We observe that \modelname produces high-quality features and consistently outperforms or matches \dggen across datasets and metrics\rebuttal{.
The per-feature JS distance improves on all five datasets, showing that the marginal feature distributions of \modelname are closer to the real data, and the pairwise JS distance further shows that this improvement extends to the dependencies between features and not only to the marginals.}
The gains are largest on the Wasserstein distance, indicating that \modelname places the feature mass correctly and does so uniformly across features rather than just on average.

On Reddit, \modelname reduces the Wasserstein distance by a factor of ${\sim}4.2$.
This suggests that \dggen more often generates out-of-distribution feature values, whereas \modelname fits the data distribution more tightly.
\rebuttal{
The DyGLib datasets show this behavior clearly.
On SocialEvo, most of the values generated by \dggen fall outside the range of the real feature, and \modelname reduces the Wasserstein distance by more than an order of magnitude on both SocialEvo and Contacts.
}
Results are closer on Wikipedia \rebuttal{ and MOOC}, where both models obtain similar \rebuttal{distances, although \modelname keeps a lower JS distance on both, with a much tighter cross-feature spread on MOOC.}

\begin{table}[t]
   \begin{minipage}[c]{0.44\textwidth}
      \caption{
         \textbf{Feature generation quality.}
         Mean and standard deviation of the Jensen--Shannon distance between the \rebuttal{marginal per-feature histograms and between the} joint pairwise feature histograms, and of the Wasserstein distance computed on the raw features.
         \rebuttal{The pairwise JS distance of SocialEvo covers its single pair of features. Contacts has a single feature, so it has no pairwise JS distance and no standard deviation over features.}
         The \rebuttal{MMD is} reported in \cref{tab:feat_js_full} (\cref{appendix:feat_1d_mmd}).
      }
      \label{tab:feat_js}
   \end{minipage}\hfill
   \begin{minipage}[c]{0.53\textwidth}
      
\providecommand{\currentrowstyle}{}
\providecommand{\resetrowstyle}{\global\let\currentrowstyle\relax}
\providecommand{\rowstyle}[1]{\gdef\currentrowstyle{#1}#1\ignorespaces}
\begin{tabular}{>{\resetrowstyle}c >{\currentrowstyle}l Z @{} Z @{}
 >{\rebuttalcolor}F @{$\,\pm\,$} >{\rebuttalcolor}F   %
 >{\currentrowstyle}F @{$\,\pm\,$} >{\currentrowstyle}F
 Z @{} Z @{}
 >{\currentrowstyle}F @{$\,\pm\,$} >{\currentrowstyle}F}
\toprule
{} & {\colmodel} & {\coldataset} & {\colwindow} & \multicolumn{2}{c}{\rebuttalcolor\colperfeatjsmean $\downarrow$} & \multicolumn{2}{c}{\colpairjsmean $\downarrow$} & {\colfeatureMmdMean} & & \multicolumn{2}{c}{\colfeatureWassersteinMean $\downarrow$} \\
\midrule

\multirow{2}{*}{\rotatebox{90}{\scriptsize Reddit}} & \dggen & Reddit & 512 & 0.17269200086593628 & 0.09335292875766754 & 0.2687458395957947 & 0.08474133908748627 & 0.05906833335757256 & 0.14211058616638184 & 0.26294252276420593 & 0.08516130596399307 \\
 & \modelname & Reddit & 512 & \bfseries 0.1025594174861908 & \bfseries 0.1182892769575119 & \bfseries 0.1906837970018386 & \bfseries 0.123350016772747 & \bfseries 0.043416604399681 & \bfseries 0.1464645117521286 & \bfseries 0.0632427260279655 & \bfseries 0.0323162041604518 \\

\midrule

\multirow{2}{*}{\rotatebox{90}{\scriptsize Wiki.}} & \dggen & Wikipedia & 512 & 0.10880722850561142 & 0.12273629009723663 & 0.18384771049022677 & 0.14339503645896912 & \bfseries 0.18352292478084564 & \bfseries 0.33252355456352234 & 0.10524817556142808 & 0.09992563724517822 \\
 & \modelname & Wikipedia & 512 & \bfseries 0.0922721102833748 & \bfseries 0.1487192213535308 & \bfseries 0.1720019876956939 & \bfseries 0.1779591292142868 & 0.2548881471157074 & 0.348622053861618 & \bfseries 0.0543153844773769 & \bfseries 0.0203846618533134 \\

\midrule

\multirow{2}{*}{\rotatebox{90}{\scriptsize MOOC}} & \dggen & MOOC & 512 & 0.08577833324670792 & 0.1096903458237648 & 0.15432703495025635 & 0.11482790112495422 & \bfseries 0.5640356540679932 & \bfseries 0.38273748755455017 & \bfseries 0.08524466305971146 & \bfseries 0.042651381343603134 \\
 & \modelname & MOOC & 512 & \bfseries 0.0651027262210846 & \bfseries 0.0437499545514583 & \bfseries 0.1101563796401024 & \bfseries 0.0376386642456054 & 0.567058801651001 & 0.3797310888767242 & 0.0980294719338417 & 0.0446901284158229 \\

\midrule

\rowstyle{\rebuttalcolor}\multirow{2}{*}{\rotatebox{90}{\scriptsize SocEvo}} & \dggen & SocialEvo & {-} & 0.7017863988876343 & 0.1183253824710846 & 0.8122607469558716 & {-} & {-} & {-} & 0.5266754031181335 & 0.5696733593940735 \\
\rowstyle{\rebuttalcolor} & \modelname & SocialEvo & {-} & \bfseries 0.43044042587280273 & \bfseries 0.20596516132354736 & \bfseries 0.6087855696678162 & \bfseries {-} & {-} & {-} & \bfseries 0.03247492015361786 & \bfseries 0.0006472781533375382 \\

\midrule

\rowstyle{\rebuttalcolor}\multirow{2}{*}{\rotatebox{90}{\scriptsize Cont.}} & \dggen & Contacts & {-} & 0.33462733030319214 & {-} & {-} & {-} & {-} & {-} & 11.689751625061035 & {-} \\
\rowstyle{\rebuttalcolor} & \modelname & Contacts & {-} & \bfseries 0.06351087242364883 & \bfseries {-} & {-} & {-} & {-} & {-} & \bfseries 0.619920551776886 & \bfseries {-} \\

\bottomrule
\end{tabular}

   \end{minipage}
\end{table}

\textbf{Topology Quality.}
\Cref{tab:topo_main} evaluates the topological properties of the generated graphs.
Following previous works \citep{guptaTIGGERScalableGenerative2022,hosseini_quality_2024,hosseiniDeepProbabilisticFramework2025}, we compute per-window graph statistics and compare them against those of the corresponding real window.
As a complement, we report the JL similarity \citep{hosseini_quality_2024}, which embeds the dynamic graph and jointly captures temporal structure, features, and topology---information that feature-only or topology-only window-level statistics cannot see.
As \tigger cannot generate features, its JL similarity is omitted on all datasets except LastFM, which has no features.
Notably, \modelname achieves the best JL similarity on all \rebuttal{eight} datasets. \rebuttal{On the JODIE datasets, it} outperforms \dggen on 21 of the 36 dataset--metric comparisons (with 4 ties), and \tigger on 32 out of 36. \rebuttal{On the DyGLib datasets, it outperforms \dggen on 19 of the 32 topology comparisons (with 1 tie), and \tigger on 26 of 32 (with 1 tie).}
This shows that \modelname generates topologically faithful graphs, with the density, degree shape, and path structure of its windows remaining within the scale of the real statistics, whereas both baselines produce spurious hubs and inflated components.
Because the JL similarity embeds the full event stream rather than marginal per-window statistics, this gain further shows that \modelname reproduces the real joint temporal structure, not merely individually plausible windows.
In \cref{appendix:topo_window_size}, we verify the robustness of these results to different time-window sizes, confirming the advantages of \modelname.
\rebuttaltwo{The median error pairs each generated window with the real window at the same position, so it is biased by the natural variability of a single real stream and is not fully representative of the quality of a generator. The Wasserstein distance and the KS statistic instead compare the distributions of the per-window statistics over all windows, and with them the advantage of \modelname over \dggen becomes clearer (\cref{appendix:topo_dist}).}

Reddit has the largest dynamic range (its real windows exhibit high \colwedgeCountMedae, \colnComponentsMedae, and \collccMedae values) and is where \modelname performs best, outperforming \tigger and \dggen on 8 of 9 metrics and coming very close to \dggen on \colplmedae.
Both baselines approximate the real data density (their \coldegreeMedaeMean remains within the real-data range) but concentrate the edges on only a few nodes, creating large hubs and clusters that are absent from the real data:
their \colwedgeCountMedae errors reach $23$--$55\times$ the real values, their \collccMedae errors $7$--$25\times$ the real sizes, and their \colclosenessMedaeMean and \colbetweennessMedaeMean errors $15$--$91\times$ the real values.
In contrast, \modelname is the only model whose errors remain below the real-data scale on density, degree shape, path structure, and connectivity.

On Wikipedia and LastFM, both \dggen and \modelname clearly outperform \tigger on the topological metrics and match the real window statistics, the exception being \dggen's degree-tail shape (\colplmedae), a poor fit, whereas \modelname stays within 25\% of the real value.
This error should be read with caution, as the power-law exponent is highly sensitive and becomes ill-posed when almost all nodes have degree one.
The JL similarity, which sees topology and features jointly, still discriminates between the models and favors \modelname over \tigger and \dggen on both datasets.

On MOOC, \dggen and \modelname perform comparably on most topological metrics, except \colnComponentsMedae, where \modelname's error is large (${\sim}7.5\times$ the real-data scale), over-fragmenting its graphs into too many connected components.
Despite this, the JL similarity rates \dggen five times worse than \modelname, indicating that even though each individual \dggen window is plausible, the windows do not compose into a realistic stream of events.

\rebuttal{
On the DyGLib datasets, \modelname leads on SocialEvo and Contacts, with the best value on five of the eight topology metrics.
Both \tigger and \dggen generate too many wedges on SocialEvo, and the windows of \tigger on Contacts contain spurious hubs (\colbetweennessMedaeMean error of $101\times$ the real value).
On Enron, \modelname is better on four metrics (betweenness, wedges, and connectivity).
On UCI, \dggen obtains the best value on more metrics, while \modelname better fits the degree tail.
}

\begin{table}[t]
   \centering
   \caption{
      \textbf{Topology generation quality.}
      Each cell reports the median absolute error between the generated and real per-snapshot statistics. The \textcolor{\pctofcolor}{(percentage)} beside it rescales this error by the magnitude of the real statistic (its median over windows).
      The \emph{Real Data} rows give the median of each real statistic over windows, \ie the scale against which each error should be judged.
   }
   
\providecommand{\currentrowstyle}{}
\providecommand{\resetrowstyle}{\global\let\currentrowstyle\relax}
\providecommand{\rowstyle}[1]{\gdef\currentrowstyle{#1}#1\ignorespaces}
\newcommand{\ncols}{21}
{\setlength{\tabcolsep}{3pt}
\begin{tabular}{>{\resetrowstyle}c >{\currentrowstyle}l Z Z
  >{\currentrowstyle}T@{\,}>{\currentrowstyle}l   %
  >{\currentrowstyle}T@{\,}>{\currentrowstyle}l   %
  >{\currentrowstyle}T@{\,}>{\currentrowstyle}l   %
  >{\currentrowstyle}S[round-mode=places,round-precision=0,table-format=5.0]@{\,}>{\currentrowstyle}l   %
  >{\currentrowstyle}T@{\,}>{\currentrowstyle}l   %
  >{\currentrowstyle}S[round-mode=places,round-precision=0,table-format=3.0]@{\,}>{\currentrowstyle}l   %
  >{\currentrowstyle}S[round-mode=places,round-precision=0,table-format=3.0]@{\,}>{\currentrowstyle}l   %
  >{\currentrowstyle}T@{\,}>{\currentrowstyle}l   %
  >{\currentrowstyle}J}  %
\toprule
{} & {\colmodel} & {\coldataset} & {\colwindow} & \multicolumn{2}{c}{\coldegreeMedaeMean $\downarrow$} & \multicolumn{2}{c}{\colclosenessMedaeMean $\downarrow$} & \multicolumn{2}{c}{\colbetweennessMedaeMean $\downarrow$} & \multicolumn{2}{c}{\colwedgeCountMedae $\downarrow$} & \multicolumn{2}{c}{\colplmedae $\downarrow$} & \multicolumn{2}{c}{\colnComponentsMedae $\downarrow$} & \multicolumn{2}{c}{\collccMedae $\downarrow$} & \multicolumn{2}{c}{\colrelEdgeDistrEntropyMedae $\downarrow$} & {\coljlSimilarity $\uparrow$} \\
\midrule

\multirow{4}{*}{\rotatebox{90}{Reddit}} & Real Data & Reddit & 512 & 1.3246753215789795 & & 0.013884562067687511 & & 9.621482604416089e-05 & & 178 & & 3.1764779090881348 & & 58 & & 14 & & 0.9566155672073364 & & \\
\cdashline{2-\ncols}

 & \tigger & Reddit & 512 & 1.1650450229644775 & \pctof{87.94947742936536} & 0.20860935747623444 & \pctof{1502.4554354632126} & 0.0041036298498511314 & \pctof{4265.070175325826} & 9855 & \pctof{5536.516853932584} & 0.8541185855865479 & \pctof{26.88885646403686} & 29 & \pctof{50.0} & 351 & \pctof{2507.1428571428573} & 0.10907500982284546 & \pctof{11.402178007751843} & {-} \\  %

 & \dggen & Reddit & 512 & 0.6005522012710571 & \pctof{45.33580353525527} & 0.2832840085029602 & \pctof{2040.2804720951597} & 0.00879451259970665 & \pctof{9140.496284502064} & 4207 & \pctof{2363.483146067416} & \bfseries 0.25750184059143066 & \bfseries \pctof{8.106520742823337} & 48 & \pctof{82.75862068965517} & 104 & \pctof{742.8571428571429} & 0.19661521911621097 & \pctof{20.55321132711576} & 0.22936931252479553 \\

 & \modelname & Reddit & 512 & \bfseries 0.148996353149414 & \bfseries \pctof{11.247763940511414} & \bfseries 0.0038975095376372 & \bfseries \pctof{28.07081360317163} & \bfseries 7.208663737401366e-05 & \bfseries \pctof{74.92258764873428} & \bfseries 146 & \bfseries \pctof{82.02247191011236} & 0.2990965843200683 & \pctof{9.4159818793114} & \bfseries 15 & \bfseries \pctof{25.86206896551724} & \bfseries 7 & \bfseries \pctof{50.0} & \bfseries 0.0274417400360107 & \bfseries \pctof{2.868627793306968} & \bfseries 0.2574158012866974 \\

\midrule

\multirow{4}{*}{\rotatebox{90}{Wikipedia}} & Real Data & Wikipedia & 512 & 1.037037014961243 & & 0.035125453025102615 & & 6.167128594825044e-05 & & 1 & & 5.088466644287109 & & 15 & & 3 & & 0.9963162541389464 &  &  \\

\cdashline{2-\ncols}

 & \tigger & Wikipedia & 512 & 0.31343281269073486 & \pctof{30.22387900999356} & 0.01199326477944851 & \pctof{34.144085688738166} & 0.0011347298277541995 & \pctof{1839.9645966623318} & 53 & \pctof{5300.0} & 2.234300136566162 & \pctof{43.909104505473} & 9 & \pctof{60.0} & 13 & \pctof{433.3333333333333} & 0.034503161907196045 & \pctof{3.4630732725539} & {-} \\  %

 & \dggen & Wikipedia & 512 & \bfseries 0.032258033752441406 & \bfseries \pctof{3.110596178058985} & \bfseries 0.009696969762444496 & \bfseries \pctof{27.606675294734274} & \bfseries 6.210114224813879e-05 & \bfseries \pctof{100.69701205881947} & \bfseries 1 & \bfseries \pctof{100.0} & 89.58309936523438 & \pctof{1760.5126578909687} & \bfseries 5 & \bfseries \pctof{33.33333333333333} & \bfseries 1 & \bfseries \pctof{33.33333333333333} & \bfseries 0.0033708810806274414 & \bfseries \pctof{0.33833444617850605} & 0.12313487380743028 \\

 & \modelname & Wikipedia & 512 & 0.0368342399597168 & \pctof{3.551873214582741} & 0.0112863313406705 & \pctof{32.13148975645878} & 8.517721289535984e-05 & \pctof{138.11486429330122} & \bfseries 1 & \bfseries \pctof{100.0} & \bfseries 1.255626916885376 & \bfseries \pctof{24.67593883699887} & \bfseries 5 & \bfseries \pctof{33.33333333333333} & \bfseries 1 & \bfseries \pctof{33.33333333333333} & 0.0038322806358337 & \pctof{0.3846449980027415} & \bfseries 0.1339774280786514 \\

\midrule

\multirow{4}{*}{\rotatebox{90}{MOOC}} & Real Data & MOOC & 512 & 1.9666666984558103 & & 0.12778709828853607 & & 0.0224645733833313 & & 133 & & 2.087773084640503 & & 6 & & 36 & & 0.9536862373352052 &  &  \\
\cdashline{2-\ncols}

 & \tigger & MOOC & 512 & 8.1505765914917 & \pctof{414.4360911735261} & 0.2453535795211792 & \pctof{192.00183962796044} & 0.035333868116140366 & \pctof{157.28706489640297} & 634991 & \pctof{477436.84210526315} & \bfseries 0.4681823253631592 & \bfseries \pctof{22.424962214884395} & \bfseries 5 & \bfseries \pctof{83.33333333333334} & 1052 & \pctof{2922.222222222222} & 0.10286420583724976 & \pctof{10.78595892551343} & {-} \\  %

 & \dggen & MOOC & 512 & \bfseries 0.3989248275756836 & \bfseries \pctof{20.28431293868513} & \bfseries 0.0748901218175888 & \bfseries \pctof{58.60538569276464} & \bfseries 0.01879216358065605 & \bfseries \pctof{83.6524391538182} & 203 & \pctof{152.63157894736844} & 0.6331453323364258 & \pctof{30.32634805929822} & 8 & \pctof{133.33333333333331} & \bfseries 27 & \bfseries \pctof{75.0} & 0.04963552951812744 & \pctof{5.204597442531968} & 0.06484098732471466 \\

 & \modelname & MOOC & 512 & 0.5 & \pctof{25.423728402610905} & 0.1060688719153404 & \pctof{83.00436690082975} & 0.0204569343477487 & \pctof{91.06308852910507} & \bfseries 129 & \bfseries \pctof{96.99248120300751} & 0.6165134906768799 & \pctof{29.529717343924776} & 45 & \pctof{750.0} & \bfseries 27 & \bfseries \pctof{75.0} & \bfseries 0.0202569365501403 & \bfseries \pctof{2.1240671991600104} & \bfseries 0.3212104737758636 \\

\midrule

\multirow{4}{*}{\rotatebox{90}{LastFM}} & Real Data & LastFM & 512 & 1.0526316165924072 & & 0.1428571492433548 & & 0.0003418803389649838 & & 1 & & 4.654970645904541 & & 5 & & 3 & & 0.993847131729126 &  &  \\
\cdashline{2-\ncols}

 & \tigger & LastFM & 512 & 1.3659729957580566 & \pctof{129.76742995617045} & 0.14864332973957062 & \pctof{104.05032616628738} & 0.02369505539536476 & \pctof{6930.803762246083} & 419 & \pctof{41900.0} & 2.319615125656128 & \pctof{49.830929174535896} & \bfseries 2 & \bfseries \pctof{40.0} & 63 & \pctof{2100.0} & 0.08921158313751221 & \pctof{8.976388851905135} & 0.10652248561382294 \\

 & \dggen & LastFM & 512 & 0.058823585510253906 & \pctof{5.58824042362307} & \bfseries 0.08888888359069824 & \bfseries \pctof{62.22221573193897} & 0.0004901961074210703 & \pctof{143.38236264334503} & 1 & \pctof{100.0} & 90.34503173828124 & \pctof{1940.8292470709157} & \bfseries 2 & \bfseries \pctof{40.0} & 1 & \pctof{33.33333333333333} & 0.007004678249359131 & \pctof{0.7048043935260119} & 0.14681603014469147 \\

 & \modelname & LastFM & 512 & \bfseries 0.0 & \bfseries \pctof{0.0} & 0.2666666805744171 & \pctof{186.66666805744163} & \bfseries 0.0 & \bfseries \pctof{0.0} & \bfseries 0.0 & \bfseries \pctof{0.0} & \bfseries 0.0 & \bfseries \pctof{0.0} & 3.0 & \pctof{60.0} & \bfseries 0.0 & \bfseries \pctof{0.0} & \bfseries 0.0001128315925598 & \bfseries \pctof{0.0113530128485159} & \bfseries 0.1742327362298965 \\

\midrule

\rowstyle{\rebuttalcolor}\multirow{4}{*}{\rotatebox{90}{Enron}} & Real Data & Enron & 6 h & 1.454545497894287 &  & 0.2113971710205078 &  & 0.007521367631852627 &  & 6 &  & 2.689798355102539 &  & 3 &  & 4 &  & 0.9492163062095642 &  & \\
\cdashline{2-\ncols}

\rowstyle{\rebuttalcolor} & \tigger & Enron & 6 h & 0.4017094373703003 & \pctof{27.61752299614182} & 0.3207327723503113 & \pctof{151.72046570064873} & 0.12449496239423752 & \pctof{1655.2170893363514} & 19 & \pctof{316.66666666666663} & 0.5376443862915039 & \pctof{19.98827849944938} & 5 & \pctof{166.66666666666669} & 7 & \pctof{175.0} & 0.05061793327331543 & \pctof{5.332602584066882} & 0.07564064115285873 \\
\rowstyle{\rebuttalcolor} & \dggen & Enron & 6 h & \bfseries 0.26190483570098877 & \bfseries \pctof{18.005956917823646} & \bfseries 0.1507054567337036 & \bfseries \pctof{71.29019560961085} & 0.013028891757130625 & \pctof{173.22503558998898} & 15 & \pctof{250.0} & \bfseries 0.5345854759216309 & \bfseries \pctof{19.874555834548858} & 5 & \pctof{166.66666666666669} & 5 & \pctof{125.0} & \bfseries 0.028676867485046387 & \bfseries \pctof{3.021109866892154} & 0.1346103399991989 \\
\rowstyle{\rebuttalcolor} & \modelname & Enron & 6 h & 0.3333333730697632 & \pctof{22.916668715576268} & 0.15328799188137054 & \pctof{72.51184637021466} & \bfseries 0.011111111380159857 & \bfseries \pctof{147.72727413435877} & \bfseries 6 & \bfseries \pctof{100.0} & 0.9711508750915528 & \pctof{36.10496947658856} & \bfseries 3 & \bfseries \pctof{100.0} & \bfseries 2 & \bfseries \pctof{50.0} & 0.03504282236099243 & \pctof{3.6917636298228333} & \bfseries 0.13532505929470062 \\

\midrule

\rowstyle{\rebuttalcolor}\multirow{4}{*}{\rotatebox{90}{UCI}} & Real Data & UCI & 3 h & 1.1428571939468384 &  & 0.14946125447750092 &  & 0.0029304029885679483 &  & 1 &  & 3.5145955085754395 &  & 4 &  & 3 &  & 0.9795815348625184 &  & \\
\cdashline{2-\ncols}

\rowstyle{\rebuttalcolor} & \tigger & UCI & 3 h & 6.248483657836914 & \pctof{546.7422956194447} & 0.2991858720779419 & \pctof{200.17620829148046} & 0.006060739979147911 & \pctof{206.82274768323654} & 24628 & \pctof{2462800.0} & 3.0868868827819824 & \pctof{87.83050212322104} & \bfseries 5 & \bfseries \pctof{125.0} & 266 & \pctof{8866.666666666668} & 0.12258923053741456 & \pctof{12.514448892160837} & 0.07926905900239944 \\
\rowstyle{\rebuttalcolor} & \dggen & UCI & 3 h & \bfseries 0.1230769157409668 & \bfseries \pctof{10.769229645912514} & 0.11544010788202286 & \pctof{77.23748090138008} & \bfseries 0.0017974881920963526 & \bfseries \pctof{61.33928333777611} & \bfseries 1 & \bfseries \pctof{100.0} & 3.051054000854492 & \pctof{86.81095714741771} & 9 & \pctof{225.0} & \bfseries 1 & \bfseries \pctof{33.33333333333333} & \bfseries 0.011893689632415771 & \bfseries \pctof{1.2141602520187376} & 0.06558875739574432 \\
\rowstyle{\rebuttalcolor} & \modelname & UCI & 3 h & 0.125 & \pctof{10.93749951105567} & \bfseries 0.08811283856630325 & \bfseries \pctof{58.95363241419018} & 0.002779676578938961 & \pctof{94.85646137350395} & 3 & \pctof{300.0} & \bfseries 1.3219938278198242 & \bfseries \pctof{37.61439473174721} & 8 & \pctof{200.0} & \bfseries 1 & \bfseries \pctof{33.33333333333333} & 0.015148818492889404 & \pctof{1.5464581511344535} & \bfseries 0.1132771223783493 \\

\midrule

\rowstyle{\rebuttalcolor}\multirow{4}{*}{\rotatebox{90}{SocialEvo}} & Real Data & SocialEvo & 1 h & 4.0 &  & 0.2874262034893036 &  & 0.04005618020892143 &  & 300 &  & 1.6153825521469116 &  & 3 &  & 29 &  & 0.9491733312606812 &  & \\
\cdashline{2-\ncols}

\rowstyle{\rebuttalcolor} & \tigger & SocialEvo & 1 h & 7.483870506286621 & \pctof{187.09676265716553} & 0.2914128005504608 & \pctof{101.38699847569939} & 0.02014654316008091 & \pctof{50.2957173025545} & 5986 & \pctof{1995.3333333333333} & 0.2172269821166992 & \pctof{13.4474017828158} & 2 & \pctof{66.66666666666666} & 28 & \pctof{96.55172413793103} & 0.04127663373947144 & \pctof{4.348692949963975} & {-} \\
\rowstyle{\rebuttalcolor} & \dggen & SocialEvo & 1 h & 3.335597038269043 & \pctof{83.38992595672607} & 0.18170544505119324 & \pctof{63.21812098038421} & 0.02314912341535091 & \pctof{57.79163987832037} & 1621 & \pctof{540.3333333333334} & \bfseries 0.12612152099609375 & \bfseries \pctof{7.807532700441325} & \bfseries 1 & \bfseries \pctof{33.33333333333333} & \bfseries 27 & \bfseries \pctof{93.10344827586206} & 0.025004565715789795 & \pctof{2.6343519031006712} & 0.1697990447282791 \\
\rowstyle{\rebuttalcolor} & \modelname & SocialEvo & 1 h & \bfseries 1.2727272510528564 & \bfseries \pctof{31.81818127632141} & \bfseries 0.09679850935935974 & \bfseries \pctof{33.677691241871074} & \bfseries 0.016408301889896393 & \bfseries \pctof{40.96322166595877} & \bfseries 210 & \bfseries \pctof{70.0} & 0.1830512285232544 & \pctof{11.33175719150684} & 2 & \pctof{66.66666666666666} & 31 & \pctof{106.89655172413792} & \bfseries 0.01752549409866333 & \bfseries \pctof{1.846395544572051} & \bfseries 0.2345074862241745 \\

\midrule

\rowstyle{\rebuttalcolor}\multirow{4}{*}{\rotatebox{90}{Contacts}} & Real Data & Contacts & 10 min & 2.076923131942749 &  & 0.022593744099140167 &  & 0.00016181962564587593 &  & 403 &  & 2.148925304412842 &  & 43 &  & 14 &  & 0.93998384475708 &  & \\
\cdashline{2-\ncols}

\rowstyle{\rebuttalcolor} & \tigger & Contacts & 10 min & 2.462000846862793 & \pctof{118.54077837535775} & 0.09595243632793428 & \pctof{424.68585953217854} & 0.016412390395998955 & \pctof{10142.397951108618} & 3646 & \pctof{904.7146401985111} & \bfseries 0.4329680204391479 & \bfseries \pctof{20.1481186689013} & 36 & \pctof{83.72093023255815} & 48 & \pctof{342.85714285714283} & \bfseries 0.030699610710144043 & \bfseries \pctof{3.265972163391568} & {-} \\
\rowstyle{\rebuttalcolor} & \dggen & Contacts & 10 min & 1.0176990032196045 & \pctof{49.000321079175} & 0.21201029419898987 & \pctof{938.3583936717164} & 0.0004683425650000572 & \pctof{289.4225982359966} & 423 & \pctof{104.96277915632754} & 2.0470900535583496 & \pctof{95.26110792937416} & 41 & \pctof{95.34883720930233} & \bfseries 9 & \bfseries \pctof{64.28571428571429} & 0.04825723171234131 & \pctof{5.133836286815379} & 0.16811734437942505 \\
\rowstyle{\rebuttalcolor} & \modelname & Contacts & 10 min & \bfseries 0.7678486108779907 & \bfseries \pctof{36.9704876925198} & \bfseries 0.012437012046575546 & \bfseries \pctof{55.04626409860441} & \bfseries 0.00027282099472358823 & \bfseries \pctof{168.59573962964564} & \bfseries 381 & \bfseries \pctof{94.54094292803971} & 1.0333391427993774 & \pctof{48.08632206421527} & \bfseries 28 & \bfseries \pctof{65.11627906976744} & 21 & \pctof{150.0} & 0.03633362054824829 & \pctof{3.8653452132082107} & \bfseries 0.2779332399368286 \\

\bottomrule
\end{tabular}}

   \label{tab:topo_main}
\end{table}

\subsection{Scaling to More Features}
\label{scaling_to_more_features}
To evaluate the ability of \modelname to scale to a higher number of features, additional trainings are performed on the Reddit and Wikipedia datasets using all 172 features instead of the first 12 (\cref{feature_quality_twelve}).
\Cref{tab:scale_full} reports the feature and topology metrics and the JL similarity for this setting.
To save space, the MMD and per-feature JS distance are omitted. Because the added features are far more atomic than the first 12 (the most frequent value covers a median of $89\%$ of the samples on Reddit and $97\%$ on Wikipedia), the MMD saturates at the same value for both models, as on MOOC (\cref{appendix:feat_1d_mmd}).
Likewise, only a reduced set of topological metrics that still covers the four structural axes is reported (\nodedegree; \wedgecount; \clocen and \btwc; \nbcc), omitting \powerlaw, \rel, and \lcc, which overlap with a reported metric on the same axis.

\modelname scales better to a high number of features, as illustrated by the pairwise JS distance ($7\%$ and $38\%$ lower on Reddit and Wikipedia, respectively) and by the Wasserstein distance ($2.3\times$ and $1.7\times$ lower), the latter indicating that \dggen produces out-of-distribution feature values that heavily penalize the distance.
The topology of \dggen remains plausible on both datasets, on par with \modelname.
This can be explained by its sequential generation, where the nodes of an event are sampled first and the associated features are then generated conditionally, so the feature dimensionality does not directly affect edge selection.
The JL similarity is the most telling indicator. From 12 (\cref{tab:topo_main}) to 172 features, that of \dggen drops by $44\%$ on Reddit (resp. $65\%$ on Wikipedia), whereas that of \modelname drops by only $14\%$ (resp. $16\%$).
This widens the gap between the two models from ${\sim}1.1\times$ to $1.7\times$ (resp. $2.6\times$).
Since the per-window topology of \dggen is preserved, this drop is explained by the degradation of its feature quality within the joint stream.
Together, these results show that \modelname scales better to a high number of features.

\begin{table}[t]
   \centering
   \caption{
      Feature and topology generation quality, using all 172 features of Reddit and Wikipedia.
   }
   
\newcommand{\ncols}{29}
{\setlength{\tabcolsep}{3pt}
\begin{tabular}{c l Z Z
  ZZ %
  F @{$\,\pm\,$} F   %
  ZZ %
  F @{$\,\pm\,$} F   %
  |
  T@{\,}l   %
  T@{\,}l   %
  T@{\,}l   %
  S[round-mode=places,round-precision=0,table-format=5.0]@{\,}l   %
  ZZ %
  S[round-mode=places,round-precision=0,table-format=3.0]@{\,}l   %
  ZZ %
  ZZ %
  |
  J}  %
\toprule
{} & {\colmodel} & {\coldataset} & {\colwindow} &

  \multicolumn{2}{Z}{\colperfeatjsmean $\downarrow$} &
  \multicolumn{2}{c}{\colpairjsmean $\downarrow$} &
  \multicolumn{2}{Z}{\colfeatureMmdMean $\downarrow$} &
  \multicolumn{2}{c|}{\colfeatureWassersteinMean $\downarrow$} &

  \multicolumn{2}{c}{\coldegreeMedaeMean $\downarrow$} &
  \multicolumn{2}{c}{\colclosenessMedaeMean $\downarrow$} &
  \multicolumn{2}{c}{\colbetweennessMedaeMean $\downarrow$} &
  \multicolumn{2}{c}{\colwedgeCountMedae $\downarrow$} &
  \multicolumn{2}{Z}{\colplmedae $\downarrow$} &
  \multicolumn{2}{c}{\colnComponentsMedae $\downarrow$} &
  \multicolumn{2}{Z}{\collccMedae $\downarrow$} &
  \multicolumn{2}{Z|}{\colrelEdgeDistrEntropyMedae $\downarrow$} &
  {\coljlSimilarity $\uparrow{}$} \\
\midrule

\multirow{3}{*}{\rotatebox{90}{Reddit}} & Real Data & Reddit & 512 & {-} & {-} & {-} & {-} & {-} & {-} & {-} & {-} & 1.3246753215789795 & & 0.013884562067687511 & & 9.621482604416089e-05 & & 178 & & 3.1764779090881348 & & 58 & & 14 & & 0.9566155672073364 & & \\

\cdashline{2-\ncols}

 & \dggen & Reddit & 512 & 0.11286170780658722 & 0.09946580976247787 & 0.18742753565311432 & 0.1054646298289299 & 0.6790286302566528 & 0.30737969279289246 & 0.24386295676231384 & 0.13156379759311676 & 0.15480709075927734 & \pctof{11.686417663065633} & 0.007755046244710684 & \pctof{55.85373313832069} & 8.544742013327777e-05 & \pctof{88.8089950857043} & \bfseries 102 & \bfseries \pctof{57.30337078651685} & 0.3457951545715332 & \pctof{10.886118665651288} & 34 & \pctof{58.620689655172406} & \bfseries 7 & \bfseries \pctof{50.0} & 0.02899181842803955 & \pctof{3.0306655486148784} & 0.12817582488059998 \\

 & \modelname & Reddit & 512 & \bfseries 0.1046560183167457 & \bfseries 0.1057332083582878 & \bfseries 0.1740528792142868 & \bfseries 0.1168287098407745 & \bfseries 0.6780153512954712 & \bfseries 0.3092963695526123 & \bfseries 0.1044520437717437 & \bfseries 0.034816775470972 & \bfseries 0.0854156017303466 & \bfseries \pctof{6.448040537852961} & \bfseries 0.0056113544851541 & \bfseries \pctof{40.414342618792645} & \bfseries 7.359051960520446e-05 & \bfseries \pctof{76.48563389952784} & 130 & \pctof{73.03370786516854} & \bfseries 0.1773781776428223 & \bfseries \pctof{5.584114944899519} & \bfseries 13 & \bfseries \pctof{22.41379310344828} & 8 & \pctof{57.14285714285714} & \bfseries 0.0276286602020263 & \bfseries \pctof{2.8881675303155654} & \bfseries 0.2212187647819519 \\

\midrule

\multirow{3}{*}{\rotatebox{90}{Wiki.}} & Real Data & Wikipedia & 512 & {-} & {-} & {-} & {-} & {-} & {-} & {-} & {-} & 1.037037014961243 & & 0.035125453025102615 & & 6.167128594825044e-05 & & 1 & & 5.088466644287109 & & 15 & & 3 & & 0.9963162541389464 &  &  \\

\cdashline{2-\ncols}

 & \dggen & Wikipedia & 512 & 0.1133900135755539 & 0.09885967522859572 & 0.176575168967247 & 0.11495837569236755 & \bfseries 0.7832493185997009 & \bfseries 0.27085164189338684 & 0.19144727289676663 & 0.15242710709571838 & \bfseries 0.04040396213531494 & \bfseries \pctof{3.89609643170017} & \bfseries 0.010204341262578964 & \bfseries \pctof{29.051130686588927} & 0.00011396011541364716 & \pctof{184.78634531680314} & \bfseries 2 & \bfseries \pctof{200.0} & \bfseries 1.1518888473510742 & \bfseries \pctof{22.637248662017182} & \bfseries 4 & \bfseries \pctof{26.666666666666668} & \bfseries 1 & \bfseries \pctof{33.33333333333333} & \bfseries 0.0044002532958984375 & \bfseries \pctof{0.4416522642904487} & 0.042888447642326355 \\

 & \modelname & Wikipedia & 512 & \bfseries 0.0680560767650604 & \bfseries 0.0829098299145698 & \bfseries 0.1089576706290245 & \bfseries 0.1067437604069709 & 0.8259096145629883 & 0.2723456621170044 & \bfseries 0.1102070361375808 & \bfseries 0.0337075591087341 & 0.0588235855102539 & \pctof{5.672274437808019} & 0.0141364410519599 & \pctof{40.24557645379634} & \bfseries 0.0001091345620807 & \bfseries \pctof{176.96170981792554} & 3 & \pctof{300.0} & 1.334155559539795 & \pctof{26.21920615393381} & 8 & \pctof{53.333333333333336} & \bfseries 1 & \bfseries \pctof{33.33333333333333} & 0.0055849552154541 & \pctof{0.5605604839078759} & \bfseries 0.1119727045297622 \\

\bottomrule
\end{tabular}}

   \label{tab:scale_full}
\end{table}

\subsection{Quality vs. Generation Efficiency}
\label{speed_vs_other}

In \cref{tab:speed}, we report the average number of events generated per second on Reddit with both 12 (\reddits) and 172 (\redditf) features for \modelname and \dggen (\tigger does not generate features).
For \modelname, the number of flow-matching sampling steps (\nbsteps) is listed alongside.
All models were run under the same setting using an A100, with speeds averaged over 10 runs.
\tigger's inductive generation runs differently from \dggen and \modelname, as it requires sampling random walks and then merging them in a discrete fashion to match window statistics to real-graph statistics.
Note that \modelname's generation speed is largely unaffected by the feature count, since its cost is dominated by the transformer operating over the fixed-length event sequence and the feature dimension enters only through a lightweight projection.
On \reddits, \modelname is faster than \dggen at every step count, from $2\times$ at 100-\nbsteps up to $11\times$ at 10-\nbsteps, which the flow-matching framework enables with minimal loss in generation quality (Pairwise JS stays almost the same).
Scaling to 172 features on \redditf slows \dggen by $14\times$ relative to its 12-feature counterpart, as its RNN does not generate in a single step. The gap to \modelname thus widens to $25\times$ at 100-\nbsteps and $175\times$ at 10-\nbsteps while matching \dggen's generation quality.
This highlights the potential of the \modelname framework not only to train faster (\cref{trainingspeed}) but also to generate dynamic graphs in real time at scale.

\begin{table}[t]
   \centering
   \caption{
      \textbf{Generation Speed} on the Reddit dataset for 12 features (\reddits) and 172 features (\redditf).
      \modelname lines with a lower Pairwise JS distance than \dggen are highlighted in green.
      For \modelname, the number of generation steps used is reported next to the name.
      Speeds are averaged over 10 runs.
   }
   
{\setlength{\tabcolsep}{3pt}
\begin{tabular}{
  c %
  l %
  Z %
  Z %
  ZZ %
  F @{$\,\pm\,$} F   %
  ZZ %
  F @{$\,\pm\,$} F   %
  |
  T@{\,}l   %
  T@{\,}l   %
  T@{\,}l   %
  S[round-mode=places,round-precision=0,table-format=5.0]@{\,}l   %
  ZZ %
  S[round-mode=places,round-precision=0,table-format=3.0]@{\,}l   %
  ZZ %
  ZZ %
  |
  J   %
  S[round-mode=places,round-precision=0,table-format=4,group-digits=true,group-separator={\,},group-minimum-digits=4]   %
}
\toprule
  {} & %
  {\colmodel} &
  {\coldataset} &
  {\colwindow} &

  \multicolumn{2}{Z}{\colperfeatjsmean $\downarrow$} &
  \multicolumn{2}{c}{\colpairjsmean $\downarrow$} &
  \multicolumn{2}{Z}{\colfeatureMmdMean $\downarrow$} &
  \multicolumn{2}{c|}{\colfeatureWassersteinMean $\downarrow$} &

  \multicolumn{2}{c}{\coldegreeMedaeMean $\downarrow$} &
  \multicolumn{2}{c}{\colclosenessMedaeMean $\downarrow$} &
  \multicolumn{2}{c}{\colbetweennessMedaeMean $\downarrow$} &
  \multicolumn{2}{c}{\colwedgeCountMedae $\downarrow$} &
  \multicolumn{2}{Z}{\colplmedae $\downarrow$} &
  \multicolumn{2}{c}{\colnComponentsMedae $\downarrow$} &
  \multicolumn{2}{Z}{\collccMedae $\downarrow$} &
  \multicolumn{2}{Z|}{\colrelEdgeDistrEntropyMedae $\downarrow$} &
  {\coljlSimilarity $\uparrow{}$} & {\coleventsPerSecond $\uparrow{}$} \\
\midrule

\multirow{8}{*}{\rotatebox{90}{Reddit 12F.}} &
\tigger & reddit & 512 & {-} & {-} & {-} & {-} & {-} & {-} & {-} & {-} & 1.1650450229644775 & \pctof{87.94947742936536} & 0.20860935747623444 & \pctof{1502.4554354632126} & 0.0041036298498511314 & \pctof{4265.070175325826} & 9855 & \pctof{5536.516853932584} & 0.8541185855865479 & \pctof{26.88885646403686} & 29 & \pctof{50.0} & 351 & \pctof{2507.1428571428573} & 0.10907500982284546 & \pctof{11.402178007751843} & {-} & 424.575765501 \\

& \dggen & reddit & 512 & 0.17269200086593628 & 0.09335292875766754 & 0.2687458395957947 & 0.08474133908748627 & 0.05906833335757256 & 0.14211058616638184 & 0.26294252276420593 & 0.08516130596399307 & 0.6005522012710571 & \pctof{45.33580353525527} & 0.2832840085029602 & \pctof{2040.2804720951597} & 0.00879451259970665 & \pctof{9140.496284502064} & 4207 & \pctof{2363.483146067416} & 0.25750184059143066 & \pctof{8.106520742823337} & 48 & \pctof{82.75862068965517} & 104 & \pctof{742.8571428571429} & 0.19661521911621097 & \pctof{20.55321132711576} & 0.22936931252479553 & 325.55842465052433 \\

\cdashline{2-30}[1pt/1pt]

& \modelname 1 & reddit & 512 & 0.554884135723114 & 0.1589133441448211 & 0.7008004784584045 & 0.0677219480276107 & 0.2447866201400756 & 0.1034252196550369 & 0.6126219630241394 & 0.0707843974232673 & 0.1928741931915283 & \pctof{14.560110696531067} & 0.0090383179485797 & \pctof{65.09616871254428} & 9.014691022457556e-05 & \pctof{93.69336715653338} & 119 & \pctof{66.85393258426966} & 0.5876660346984863 & \pctof{18.50055475018828} & 55 & \pctof{94.82758620689656} & 8 & \pctof{57.14285714285714} & 0.0346354246139526 & \pctof{3.6206210520976994} & 0.1226597279310226 & 10191.639453125 \\
& \modelname 5 & reddit & 512 & 0.1876860111951828 & 0.1007194221019744 & 0.2851443290710449 & 0.0981509760022163 & 0.0588512606918811 & 0.1425716876983642 & 0.1965579986572265 & 0.0188544858247041 & 0.1827971935272216 & \pctof{13.799396014212151} & 0.0033040223643183 & \pctof{23.79637433439526} & 6.161211058497429e-05 & \pctof{64.03598397267322} & 154 & \pctof{86.51685393258427} & 0.5437741279602051 & \pctof{17.118775685624247} & 16 & \pctof{27.58620689655172} & 8 & \pctof{57.14285714285714} & 0.0294158458709716 & \pctof{3.0749913423263475} & 0.1789285838603973 & 5693.983813476562 \\
& \cellcolor{green!15}\modelname 10 & \cellcolor{green!15}reddit & \cellcolor{green!15}512 & \cellcolor{green!15}0.1409797370433807 & \cellcolor{green!15}0.1064087077975273 & \cellcolor{green!15}0.2289282530546188 & \cellcolor{green!15}0.1091488152742386 & \cellcolor{green!15}0.0478608347475528 & \cellcolor{green!15}0.1458411961793899 & \cellcolor{green!15}0.1233165487647056 & \cellcolor{green!15}0.0196683909744024 & \cellcolor{green!15}0.201107382774353 & \cellcolor{green!15}\pctof{15.18163579394218} & \cellcolor{green!15}0.004327418282628 & \cellcolor{green!15}\pctof{31.167121163287764} & \cellcolor{green!15}6.705807754769921e-05 & \cellcolor{green!15}\pctof{69.69619995666858} & \cellcolor{green!15}158 & \cellcolor{green!15}\pctof{88.76404494382022} & \cellcolor{green!15}0.6332011222839355 & \cellcolor{green!15}\pctof{19.93406346294117} & \cellcolor{green!15}22 & \cellcolor{green!15}\pctof{37.93103448275862} & \cellcolor{green!15}9 & \cellcolor{green!15}\pctof{64.28571428571429} & \cellcolor{green!15}0.0304492712020874 & \cellcolor{green!15}\pctof{3.183020666387278} & \cellcolor{green!15}0.1865038871765136 & \cellcolor{green!15}3535.790698242188  \\
& \cellcolor{green!15}\modelname 20 & \cellcolor{green!15}reddit & \cellcolor{green!15}512 & \cellcolor{green!15}0.1194481328129768 & \cellcolor{green!15}0.1121085882186889 & \cellcolor{green!15}0.206331193447113 & \cellcolor{green!15}0.1166561469435691 & \cellcolor{green!15}0.0447024665772914 & \cellcolor{green!15}0.146875724196434 & \cellcolor{green!15}0.0897488817572593 & \cellcolor{green!15}0.0221033487468957 & \cellcolor{green!15}0.197155237197876 & \cellcolor{green!15}\pctof{14.883287548746054} & \cellcolor{green!15}0.0060354843735694 & \cellcolor{green!15}\pctof{43.46902944541127} & \cellcolor{green!15}9.582219354342668e-05 & \cellcolor{green!15}\pctof{99.59192099920858} & \cellcolor{green!15}155 & \cellcolor{green!15}\pctof{87.07865168539325} & \cellcolor{green!15}0.5906109809875488 & \cellcolor{green!15}\pctof{18.593265808578984} & \cellcolor{green!15}25 & \cellcolor{green!15}\pctof{43.103448275862064} & \cellcolor{green!15}9 & \cellcolor{green!15}\pctof{64.28571428571429} & \cellcolor{green!15}0.0284624695777893 & \cellcolor{green!15}\pctof{2.975329960485617} & \cellcolor{green!15}0.2188581675291061 & \cellcolor{green!15}2061.1695190429687  \\
& \cellcolor{green!15}\modelname 50 & \cellcolor{green!15}reddit & \cellcolor{green!15}512 & \cellcolor{green!15}0.1068933978676796 & \cellcolor{green!15}0.1166643947362899 & \cellcolor{green!15}0.1942984759807586 & \cellcolor{green!15}0.1219480708241462 & \cellcolor{green!15}0.0435875840485096 & \cellcolor{green!15}0.146716222167015 & \cellcolor{green!15}0.0694340020418167 & \cellcolor{green!15}0.024688621982932 & \cellcolor{green!15}0.2107534408569336 & \cellcolor{green!15}\pctof{15.909818611682203} & \cellcolor{green!15}0.0059895906597375 & \cellcolor{green!15}\pctof{43.13849173303569} & \cellcolor{green!15}9.001931175589561e-05 & \cellcolor{green!15}\pctof{93.5607488544212} & \cellcolor{green!15}164 & \cellcolor{green!15}\pctof{92.13483146067416} & \cellcolor{green!15}0.580388069152832 & \cellcolor{green!15}\pctof{18.27143414069714} & \cellcolor{green!15}23 & \cellcolor{green!15}\pctof{39.6551724137931} & \cellcolor{green!15}9 & \cellcolor{green!15}\pctof{64.28571428571429} & \cellcolor{green!15}0.0330805778503418 & \cellcolor{green!15}\pctof{3.458084834110997} & \cellcolor{green!15}0.1591815501451492 & \cellcolor{green!15}915.034176635742  \\

& \cellcolor{green!15}\modelname 100 & \cellcolor{green!15}reddit & \cellcolor{green!15}512 & \cellcolor{green!15}0.1025594174861908 & \cellcolor{green!15}0.1182892769575119 & \cellcolor{green!15}0.1906837970018386 & \cellcolor{green!15}0.123350016772747 & \cellcolor{green!15}0.043416604399681 & \cellcolor{green!15}0.1464645117521286 & \cellcolor{green!15}0.0632427260279655 & \cellcolor{green!15}0.0323162041604518 & \cellcolor{green!15}0.148996353149414 & \cellcolor{green!15}\pctof{11.247763940511414} & \cellcolor{green!15}0.0038975095376372 & \cellcolor{green!15}\pctof{28.07081360317163} & \cellcolor{green!15}7.208663737401366e-05 & \cellcolor{green!15}\pctof{74.92258764873428} & \cellcolor{green!15}146 & \cellcolor{green!15}\pctof{82.02247191011236} & \cellcolor{green!15}0.2990965843200683 & \cellcolor{green!15}\pctof{9.4159818793114} & \cellcolor{green!15}15 & \cellcolor{green!15}\pctof{25.86206896551724} & \cellcolor{green!15}7 & \cellcolor{green!15}\pctof{50.0} & \cellcolor{green!15}0.0274417400360107 & \cellcolor{green!15}\pctof{2.868627793306968} & \cellcolor{green!15}0.2574158012866974 & \cellcolor{green!15}498.43504638671874  \\

\midrule

\multirow{8}{*}{\rotatebox{90}{Reddit 172F.}}
& \dggen & reddit & 512 & 0.11286170780658722 & 0.09946580976247787 & 0.18742753565311432 & 0.1054646298289299 & 0.6790286302566528 & 0.30737969279289246 & 0.24386295676231384 & 0.13156379759311676 & 0.15480709075927734 & \pctof{11.686417663065633} & 0.007755046244710684 & \pctof{55.85373313832069} & 8.544742013327777e-05 & \pctof{88.8089950857043} & 102 & \pctof{57.30337078651685} & 0.3457951545715332 & \pctof{10.886118665651288} & 34 & \pctof{58.620689655172406} & 7 & \pctof{50.0} & 0.02899181842803955 & \pctof{3.0306655486148784} & 0.12817582488059998 & 23.773251957718887 \\

\cdashline{2-30}[1pt/1pt]

& \modelname 1 & reddit & 512 & 0.2825990319252014 & 0.2575068771839142 & 0.4341231286525726 & 0.2671393752098083 & 0.7118657827377319 & 0.2471007406711578 & 0.4095383584499359 & 0.1820437908172607 & 0.2767319679260254 & \pctof{20.890550568736185} & 0.0095158517360687 & \pctof{68.53548343605482} & 9.37954755499959e-05 & \pctof{97.48547017789696} & 158 & \pctof{88.76404494382022} & 1.6762425899505615 & \pctof{52.77047843319512} & 63 & \pctof{108.6206896551724} & 10 & \pctof{71.42857142857143} & 0.0397372245788574 & \pctof{4.153938733702919} & 0.113508328795433 & 9538.58076171875 \\
& \modelname 5 & reddit & 512 & 0.1445774435997009 & 0.1105311661958694 & 0.2223166674375534 & 0.1222138255834579 & 0.6815582513809204 & 0.3025586903095245 & 0.1736384928226471 & 0.0459483079612255 & 0.0884211063385009 & \pctof{6.674926670567491} & 0.0031438497826457 & \pctof{22.642772363430502} & 4.721328514278866e-05 & \pctof{49.07069636140963} & 84 & \pctof{47.19101123595505} & 0.4224650859832764 & \pctof{13.299796128742877} & 10 & \pctof{17.24137931034483} & 6 & \pctof{42.85714285714285} & 0.0164791941642761 & \pctof{1.722655863983492} & 0.1521432101726532 & 6559.387451171875 \\
& \modelname 10 & reddit & 512 & 0.1199720725417137 & 0.1042510941624641 & 0.1895816922187805 & 0.1172012835741043 & 0.6786579489707947 & 0.3079867959022522 & 0.1302918493747711 & 0.032470628619194 & 0.0759472846984863 & \pctof{5.733275426914354} & 0.0028137024492025 & \pctof{20.26497080344114} & 4.636694939108565e-05 & \pctof{48.19106503378601} & 94 & \pctof{52.80898876404494} & 0.4155733585357666 & \pctof{13.08283483876217} & 12 & \pctof{20.689655172413797} & 7 & \pctof{50.0} & 0.0171015858650207 & \pctof{1.7877177051326576} & 0.1850873529911041 & 4169.3412109375 \\
& \cellcolor{green!15}\modelname 20 & \cellcolor{green!15}reddit & \cellcolor{green!15}512 & \cellcolor{green!15}0.1101231575012207 & \cellcolor{green!15}0.1038273945450782 & \cellcolor{green!15}0.1781457811594009 & \cellcolor{green!15}0.1164722964167594 & \cellcolor{green!15}0.6781013011932373 & \cellcolor{green!15}0.309096097946167 & \cellcolor{green!15}0.1122548058629036 & \cellcolor{green!15}0.0307221878319978 & \cellcolor{green!15}0.0678837299346923 & \cellcolor{green!15}\pctof{5.124556095283537} & \cellcolor{green!15}0.0053934063762426 & \cellcolor{green!15}\pctof{38.8446272194231} & \cellcolor{green!15}7.052662840578705e-05 & \cellcolor{green!15}\pctof{73.30120658682748} & \cellcolor{green!15}128 & \cellcolor{green!15}\pctof{71.91011235955057} & \cellcolor{green!15}0.1961071491241455 & \cellcolor{green!15}\pctof{6.173729354864036} & \cellcolor{green!15}12 & \cellcolor{green!15}\pctof{20.689655172413797} & \cellcolor{green!15}8 & \cellcolor{green!15}\pctof{57.14285714285714} & \cellcolor{green!15}0.0261719226837158 & \cellcolor{green!15}\pctof{2.735887181944984} & \cellcolor{green!15}0.1939403861761093 & \cellcolor{green!15}2426.310693359375  \\
& \cellcolor{green!15}\modelname 50 & \cellcolor{green!15}reddit & \cellcolor{green!15}512 & \cellcolor{green!15}0.1056699976325035 & \cellcolor{green!15}0.1044786870479583 & \cellcolor{green!15}0.1740602999925613 & \cellcolor{green!15}0.1162303835153579 & \cellcolor{green!15}0.6779216527938843 & \cellcolor{green!15}0.3093065619468689 & \cellcolor{green!15}0.1051884144544601 & \cellcolor{green!15}0.0320929214358329 & \cellcolor{green!15}0.0806162357330322 & \cellcolor{green!15}\pctof{6.085735456816672} & \cellcolor{green!15}0.0052832467481493 & \cellcolor{green!15}\pctof{38.05123072944947} & \cellcolor{green!15}7.161425310187042e-05 & \cellcolor{green!15}\pctof{74.43161937329779} & \cellcolor{green!15}134 & \cellcolor{green!15}\pctof{75.28089887640449} & \cellcolor{green!15}0.1872329711914062 & \cellcolor{green!15}\pctof{5.894357730482528} & \cellcolor{green!15}11 & \cellcolor{green!15}\pctof{18.96551724137931} & \cellcolor{green!15}8 & \cellcolor{green!15}\pctof{57.14285714285714} & \cellcolor{green!15}0.0275691151618957 & \cellcolor{green!15}\pctof{2.881942977614166} & \cellcolor{green!15}0.1853614449501037 & \cellcolor{green!15}1069.7513305664063  \\
& \cellcolor{green!15}\modelname 100 & \cellcolor{green!15}reddit & \cellcolor{green!15}512 & \cellcolor{green!15}0.1046560183167457 & \cellcolor{green!15}0.1057332083582878 & \cellcolor{green!15}0.1740528792142868 & \cellcolor{green!15}0.1168287098407745 & \cellcolor{green!15}0.6780153512954712 & \cellcolor{green!15}0.3092963695526123 & \cellcolor{green!15}0.1044520437717437 & \cellcolor{green!15}0.034816775470972 & \cellcolor{green!15}0.0854156017303466 & \cellcolor{green!15}\pctof{6.448040537852961} & \cellcolor{green!15}0.0056113544851541 & \cellcolor{green!15}\pctof{40.414342618792645} & \cellcolor{green!15}7.359051960520446e-05 & \cellcolor{green!15}\pctof{76.48563389952784} & \cellcolor{green!15}130 & \cellcolor{green!15}\pctof{73.03370786516854} & \cellcolor{green!15}0.1773781776428223 & \cellcolor{green!15}\pctof{5.584114944899519} & \cellcolor{green!15}13 & \cellcolor{green!15}\pctof{22.41379310344828} & \cellcolor{green!15}8 & \cellcolor{green!15}\pctof{57.14285714285714} & \cellcolor{green!15}0.0276286602020263 & \cellcolor{green!15}\pctof{2.8881675303155654} & \cellcolor{green!15}0.2212187647819519 & \cellcolor{green!15}582.7856567382812  \\

\bottomrule
\end{tabular}}

   \label{tab:speed}
\end{table}

\section{Conclusion}
In this paper, we introduced \modelname, a framework for dynamic graph generation.
Unlike prior dynamic graph generation approaches, \modelname uses a diffusion-based method to generate both the topology and the event features.
This design enables faster training, better scaling, and generation of large graphs, in terms of both features and topology.
As a result, \modelname is a promising approach for dynamic graph generation in real-world applications, where the number of events and features can be very large.

\subsubsection*{Acknowledgments}
This work has been supported by the ANR-23-CE38-0008 grant.
We thank CRIANN, who provided us with the computation resources necessary for our experiments.
This work was granted access to the HPC resources of IDRIS under the allocation 2026-AD011017917 made by GENCI.

\subsection*{Reproducibility statement}

We provide the information needed to reproduce our results throughout the paper and appendix.
The full experimental setup of \modelname is described in \cref{appendix:impl_details}, and all model, training, and generation hyperparameters are reported in \cref{appendix:impl_details:hp} and \cref{tab:hp}.
All experiments use only publicly available benchmark datasets, so no private or newly collected data is required to reproduce our results.
Upon acceptance, we will release our source code and the trained model checkpoints in a public GitHub repository, whose link will be added in the final version of the paper.

\bibliography{iclr2027_conference}
\bibliographystyle{iclr2027_conference}

\clearpage
\appendix
\section{Appendix}

\subsection{Limitations and Future Work}

While we have made significant progress, our model is still far from addressing every challenge in dynamic graph generation.
We believe that \modelname can achieve significantly better results in both quality and efficiency.
First, our model currently uses fixed condition and generation window sizes, which may not be suitable for all applications.
Furthermore, decoupling these parameters from the training dynamics could help achieve better generalization.
Second, our model still requires multiple steps to generate a window of events.
Diffusion methods that require fewer generation steps would be better suited to resource-constrained devices.
\rebuttal{Future work could also explore the design space of the node encoder.
This work uses a Temporal Graph Network (TGN) encoder, and other node-level encoders or adaptations of edge-based encoders could yield a more discriminative space for event generation.
}

\subsection{Implementation Details}
\label{appendix:impl_details}

\subsubsection{Datasets and Baselines}

Dataset statistics for Reddit, Wikipedia, LastFM, and MOOC~\citep{kumar2019predicting}\rebuttal{, and for the Enron, UCI, SocialEvo, and Contacts datasets of the DyGLib benchmark~\citep{poursafaei2022towards,yuDyGFormer2023},} are summarized in \cref{tab:dataset-stats}.

\begin{table}[h]
   \centering
   \caption{Dataset statistics.}
   \label{tab:dataset-stats}
   \resizebox{\textwidth}{!}{%
   \begin{tabular}{lrrrrrrr}
         \toprule
         Dataset & \#Nodes & \#Edges & \#Features & \#Train & \#Valid & \#Test & Time span (days) \\
         \midrule
         Reddit & 10,984 & 672,447 & 172 & 537,957 & 67,245 & 67,245 & 31.0 \\
         Wikipedia & 9,227 & 157,474 & 172 & 125,979 & 15,747 & 15,748 & 31.0 \\
         MOOC & 7,144 & 411,749 & 4 & 329,400 & 41,174 & 41,175 & 29.8 \\
         LastFM & 1,980 & 1,293,103 & 0 & 1,034,482 & 129,310 & 129,311 & 1586.9 \\
         \midrule
         \rebuttalcolor Enron & \rebuttalcolor 184 & \rebuttalcolor 125,235 & \rebuttalcolor 0 & \rebuttalcolor 87,664 & \rebuttalcolor 18,786 & \rebuttalcolor 18,785 & \rebuttalcolor 1316.4 \\
         \rebuttalcolor UCI & \rebuttalcolor 1,899 & \rebuttalcolor 59,835 & \rebuttalcolor 0 & \rebuttalcolor 41,884 & \rebuttalcolor 8,975 & \rebuttalcolor 8,976 & \rebuttalcolor 193.7 \\
         \rebuttalcolor SocialEvo & \rebuttalcolor 74 & \rebuttalcolor 2,099,519 & \rebuttalcolor 2 & \rebuttalcolor 1,469,665 & \rebuttalcolor 314,930 & \rebuttalcolor 314,924 & \rebuttalcolor 242.3 \\
         \rebuttalcolor Contacts & \rebuttalcolor 692 & \rebuttalcolor 2,426,279 & \rebuttalcolor 1 & \rebuttalcolor 1,698,494 & \rebuttalcolor 364,005 & \rebuttalcolor 363,780 & \rebuttalcolor 28.0 \\
         \bottomrule
   \end{tabular}%
   }
\end{table}

Following \citet{hosseiniDeepProbabilisticFramework2025}, training is performed on a 70/15/15 train/val/test split.
\rebuttal{On the DyGLib datasets, the split is taken at the 70\% and 85\% quantiles of the timestamps.}
For baselines, we use the existing code for \tigger and \dggen.

\subsubsection{Model Implementation}
In \modelname, the event encoder $\getemb$ (\cref{eq:getemb}) is a pre-trained TGN~\citep{rossiTemporalGraphNetworks2020}, with all hyperparameters fixed to those of \citet{hosseiniDeepProbabilisticFramework2025}.
The denoiser $v_\theta$ is an encoder--decoder transformer, where the encoder processes $\zpast$ and $\zsum$, and the decoder predicts the denoised $\zcur$.
Both $\zpast$ and $\zsum$ are encoded by 4-layer transformer blocks with hidden size 384, 4 heads, and dropout rate $0.1$.
The decoder is a small DiT~\citep{peebles2023scalable} adapted for event processing, with depth 12, hidden size 384, 6 heads, and full cross-attention.
We use RoPE~\citep{su2024roformer} positional embeddings in the decoder and in the $\zpast$ encoder.

\subsubsection{Training and Generation Hyperparameters}
\label{appendix:impl_details:hp}
The pre-trained TGN encoder $\getemb$ reuses the per-dataset configuration of \citet{hosseiniDeepProbabilisticFramework2025}, namely the memory, embedding, and time-encoding dimensions, the neighbour budget, and whether the inter-event interval $\Delta t$ is log-compressed before entering the TGN.
TGN encoder is kept frozen during diffusion training.
\rebuttal{It is pretrained alone on the training split with the original self-supervised objective of TGN, a binary cross-entropy that scores each event against an event with a uniformly sampled negative destination. Its link-prediction head is then discarded, and the diffusion model consumes only the raw memory states.}

All datasets share the same diffusion training setup, with the velocity network optimised with AdamW ($\beta = (0.9, 0.999)$, weight decay $0$), a linear learning-rate warm-up over the first $5$ epochs (from $0.3\times$ to $1\times$ the base rate), an exponential moving average of the weights (decay $0.999$), and gradient-norm clipping at $1.0$.
The TGN memory is updated in chunks of $100$ events, and the diffusion model consumes $\log(1 + \Delta t)$ in place of the raw inter-event interval.
We found the log-encoding to be necessary for LastFM mainly, as the inter-event $\Delta t$ can be very large.

At generation time we integrate the flow-matching ODE with a fixed-step Euler solver for $100$ steps, and match each generated embedding to a node identifier by cosine similarity (\cref{eq:generation:node_matching}) with stochastic sampling.

All per-dataset hyperparameters are summarised in \cref{tab:hp}.

\begin{table}[h]
   \centering
   \caption{
      \textbf{Training and generation hyperparameters of \modelname.}
      Pre-trained TGN encoder dimensions follow \citet{hosseiniDeepProbabilisticFramework2025}. $L$ and $M$ are the sizes of the generated window $\zcur$ and the past context $\zpast$, and $\temp$ is the matching temperature.
      Generation always uses a $100$-step Euler solver with cosine node matching (\cref{eq:generation:node_matching}) and stochastic sampling.
      The last two rows use the full $172$-dimensional feature set (\cref{scaling_to_more_features}).
   }
   \label{tab:hp}
   \resizebox{\textwidth}{!}{%
   \begin{tabular}{lccccccccc}
      \toprule
      & \multicolumn{5}{c}{Pre-trained TGN encoder} & \multicolumn{3}{c}{Diffusion training} & \multicolumn{1}{c}{Generation} \\
      \cmidrule(lr){2-6}\cmidrule(lr){7-9}\cmidrule(lr){10-10}
      Dataset & $d_{\text{mem}}$ & $d_{\text{emb}}$ & $d_{\text{time}}$ & \#Nbr & $\log\Delta t$ & $L$ & $M$ & LR & $\temp$ \\
      \midrule
      Reddit    & 76  & 76  & 100 & 100 & Yes & 4096 & 4096 & $10^{-4}$ & 0.05 \\
      Wikipedia & 76  & 100 & 300 & 200 & Yes & 2048 & 4096 & $10^{-4}$ & 0.3  \\
      MOOC      & 300 & 200 & 200 & 10  & Yes & 2048 & 4096 & $10^{-4}$ & 0.3  \\
      LastFM    & 300 & 76  & 800 & 40  & No  & 4096 & 4096 & $10^{-4}$ & 0.05 \\
      \midrule
      Reddit (172F)    & 76 & 76  & 100 & 100 & Yes & 4096 & 4096 & $10^{-3}$ & 0.3  \\
      Wikipedia (172F) & 76 & 100 & 300 & 200 & Yes & 512  & 4096 & $10^{-4}$ & 0.05 \\
      \bottomrule
   \end{tabular}%
   }
\end{table}

\subsubsection[\dggen Adaptations]{\dggen Adaptations}
\label{appendix:impl_details:dggen_call}

\textbf{\dggen Full-Feature Adaptation:}
For the feature-scaling comparison (\cref{scaling_to_more_features}), we train \dggen on the full feature set ($172$ features on Reddit and Wikipedia) from the authors' released code.

The only change we make to \dggen's released configuration concerns its feature-noise schedule.
\dggen adds Gaussian noise of scale $\epsilon$ to the message features and anneals it during training, rescaling $\epsilon$ by $0.1$ whenever the feature loss improves by less than a relative threshold over the last $T$ epochs, down to a floor of $\epsilon/100$, yielding three successive noise regimes.
This schedule was tuned for the 12-feature setting over $50$ epochs, where on Reddit it spends roughly $22\%/20\%/58\%$ of training in the three regimes.
With $172$ features, however, the per-epoch cost grows sharply and only about $9$ epochs fit the same compute budget of 48h.

Since the first rescale cannot fire before epoch $T{+}1$ (with the released $T{=}10$ on Reddit and $T{=}20$ on Wikipedia), the entire run stays pinned at the initial, highest noise level, where the features are buried in noise and no feature learning occurs.

To restore a fair comparison, we keep the initial noise $\epsilon$ and its floor unchanged (final noise level matches the 12-feature) and only shorten the annealing window $T$ so that each noise regime receives the same \emph{proportion} of the (now shorter) epoch budget as before, namely $T{=}1$ on Reddit (rescaling at epochs $2$ and $3$, \ie $\approx 22\%/11\%/67\%$) and $T{=}5$ on Wikipedia.
We additionally clip the gradient norm to $1.0$ for stability and, on Wikipedia, lower the learning rate to $10^{-4}$.

\rebuttal{
\textbf{\dggen DyGLib Implementation:}
No \dggen configuration exists for these datasets, so the TGN hyperparameters of DyGLib~\citep{yuDyGFormer2023} are used, and \dggen's released code is otherwise unchanged.
Only the noise schedule changes, along with three fixes for numerical stability.
The noise schedule takes the initial noise and threshold of \dggen's MOOC configuration, and training lasts $5T$ epochs, so the three noise regimes keep the proportions of the released runs.
With $T$ the largest value that fits a 24h budget, \ie $T{=}58$ on Enron, $T{=}122$ on UCI, and $T{=}3$ on SocialEvo and Contacts.
The gradient norm is clipped to $1.0$ on SocialEvo, and the TGN memory and the GNN receive $\log(1+\Delta t)$ on Enron.
On Contacts, where most consecutive events share a timestamp, the inter-event time head models $\Delta t + 1$.
}

\subsection{Feature Generation Quality: Maximum Mean Discrepancy}
\label{appendix:feat_1d_mmd}

\Cref{tab:feat_js_full} extends \cref{tab:feat_js} with the MMD.
On Reddit, \modelname obtains a lower MMD, with a smaller gain than on the other metrics.

On Wikipedia and MOOC, \dggen performs better on the MMD.
This lower MMD performance of \modelname is explained by the near-discrete nature of the features, on which the median-heuristic kernel bandwidth collapses, making the MMD a less suitable metric in this case.
On Wikipedia, this follows from the atom-like structure of the dataset. For each of the first 12 features, the single most frequent value alone accounts for at least $47\%$ of the real samples.
This causes the median-heuristic bandwidth of the Gaussian kernel to collapse, so the MMD heavily penalizes values that lie near, but not exactly on, these point masses.
In \dggen, each feature is produced by an RNN with a GMM head trained by maximum likelihood. When $47\%$ of a feature's samples concentrate on a single value, the likelihood is maximized by placing one component's mean on that atom and shrinking its scale. GMMs can undergo component collapse and, on this dataset, are encouraged to do so.
By contrast, \modelname uses a diffusion process, which may be less subject to variance collapse.
The low Wasserstein distance indicates that \modelname does place the feature mass correctly as a continuous generator, but it does not reproduce the repeated values as closely as the GMM of \dggen.
The four MOOC features are even more atomic (the most frequent value covers $87\%$, $83\%$, $64\%$, and $94\%$ of the data, respectively).
This makes the MMD near-saturated for both \dggen and \modelname, as the bandwidth collapse turns it into an exact-match statistic that neither continuous generator can satisfy, though we still report it for completeness.

\begin{table}[t]
   \centering
   \caption{
      \textbf{Feature generation quality (full metric set).}
      Mean and standard deviation of the Jensen--Shannon distance between the marginal per-feature histograms and between the joint pairwise histograms.
      The Maximum Mean Discrepancy (MMD) and the Wasserstein distance are computed on the raw features.
   }
   
\providecommand{\currentrowstyle}{}
\providecommand{\resetrowstyle}{\global\let\currentrowstyle\relax}
\providecommand{\rowstyle}[1]{\gdef\currentrowstyle{#1}#1\ignorespaces}
\providecommand{\rebuttalcolor}{\color{black}}
\begin{tabular}{>{\resetrowstyle}c >{\currentrowstyle}l Z
 >{\currentrowstyle}F @{$\,\pm\,$} >{\currentrowstyle}F
 >{\currentrowstyle}F @{$\,\pm\,$} >{\currentrowstyle}F
 >{\currentrowstyle}F @{$\,\pm\,$} >{\currentrowstyle}F
 >{\currentrowstyle}F @{$\,\pm\,$} >{\currentrowstyle}F}
\toprule
{\colmodel} & {\coldataset} & {\colwindow} & \multicolumn{2}{c}{\colperfeatjsmean $\downarrow$} & \multicolumn{2}{c}{\colpairjsmean $\downarrow$} & \multicolumn{2}{c}{\colfeatureMmdMean $\downarrow$} & \multicolumn{2}{c}{\colfeatureWassersteinMean $\downarrow$} \\
\midrule

\dggen & Reddit & 512 & 0.17269200086593628 & 0.09335292875766754 & 0.2687458395957947 & 0.08474133908748627 & 0.05906833335757256 & 0.14211058616638184 & 0.26294252276420593 & 0.08516130596399307 \\
\modelname & Reddit & 512 & \bfseries 0.1025594174861908 & \bfseries 0.1182892769575119 & \bfseries 0.1906837970018386 & \bfseries 0.123350016772747 & \bfseries 0.043416604399681 & \bfseries 0.1464645117521286 & \bfseries 0.0632427260279655 & \bfseries 0.0323162041604518 \\

\midrule

\dggen & Wikipedia & 512 & 0.10880722850561142 & 0.12273629009723663 & 0.18384771049022677 & 0.14339503645896912 & \bfseries 0.18352292478084564 & \bfseries 0.33252355456352234 & 0.10524817556142808 & 0.09992563724517822 \\
\modelname & Wikipedia & 512 & \bfseries 0.0922721102833748 & \bfseries 0.1487192213535308 & \bfseries 0.1720019876956939 & \bfseries 0.1779591292142868 & 0.2548881471157074 & 0.348622053861618 & \bfseries 0.0543153844773769 & \bfseries 0.0203846618533134 \\

\midrule

\dggen & MOOC & 512 & 0.08577833324670792 & 0.1096903458237648 & 0.15432703495025635 & 0.11482790112495422 & \bfseries 0.5640356540679932 & \bfseries 0.38273748755455017 & \bfseries 0.08524466305971146 & \bfseries 0.042651381343603134 \\
\modelname & MOOC & 512 & \bfseries 0.0651027262210846 & \bfseries 0.0437499545514583 & \bfseries 0.1101563796401024 & \bfseries 0.0376386642456054 & 0.567058801651001 & 0.3797310888767242 & 0.0980294719338417 & 0.0446901284158229 \\

\midrule

\rowstyle{\rebuttalcolor}\dggen & SocialEvo & {-} & 0.7017863988876343 & 0.1183253824710846 & 0.8122607469558716 & {-} & 1.037302851676941 & 0.05868881195783615 & 0.5266754031181335 & 0.5696733593940735 \\
\rowstyle{\rebuttalcolor}\modelname & SocialEvo & {-} & \bfseries 0.43044042587280273 & \bfseries 0.20596516132354736 & \bfseries 0.6087855696678162 & \bfseries {-} & \bfseries 0.5029605627059937 & \bfseries 0.7029798030853271 & \bfseries 0.03247492015361786 & \bfseries 0.0006472781533375382 \\

\midrule

\rowstyle{\rebuttalcolor}\dggen & Contacts & {-} & 0.33462733030319214 & {-} & {-} & {-} & 0.1990657150745392 & {-} & 11.689751625061035 & {-} \\
\rowstyle{\rebuttalcolor}\modelname & Contacts & {-} & \bfseries 0.06351087242364883 & \bfseries {-} & {-} & {-} & \bfseries 0.0005543231964111328 & \bfseries {-} & \bfseries 0.619920551776886 & \bfseries {-} \\

\bottomrule
\end{tabular}

   \label{tab:feat_js_full}
\end{table}

\subsection{Training Speed}
\label{trainingspeed}

\subsubsection{With 12 Features}

\begin{figure}[t]
   \centering
   \resizebox{\textwidth}{!}{\input{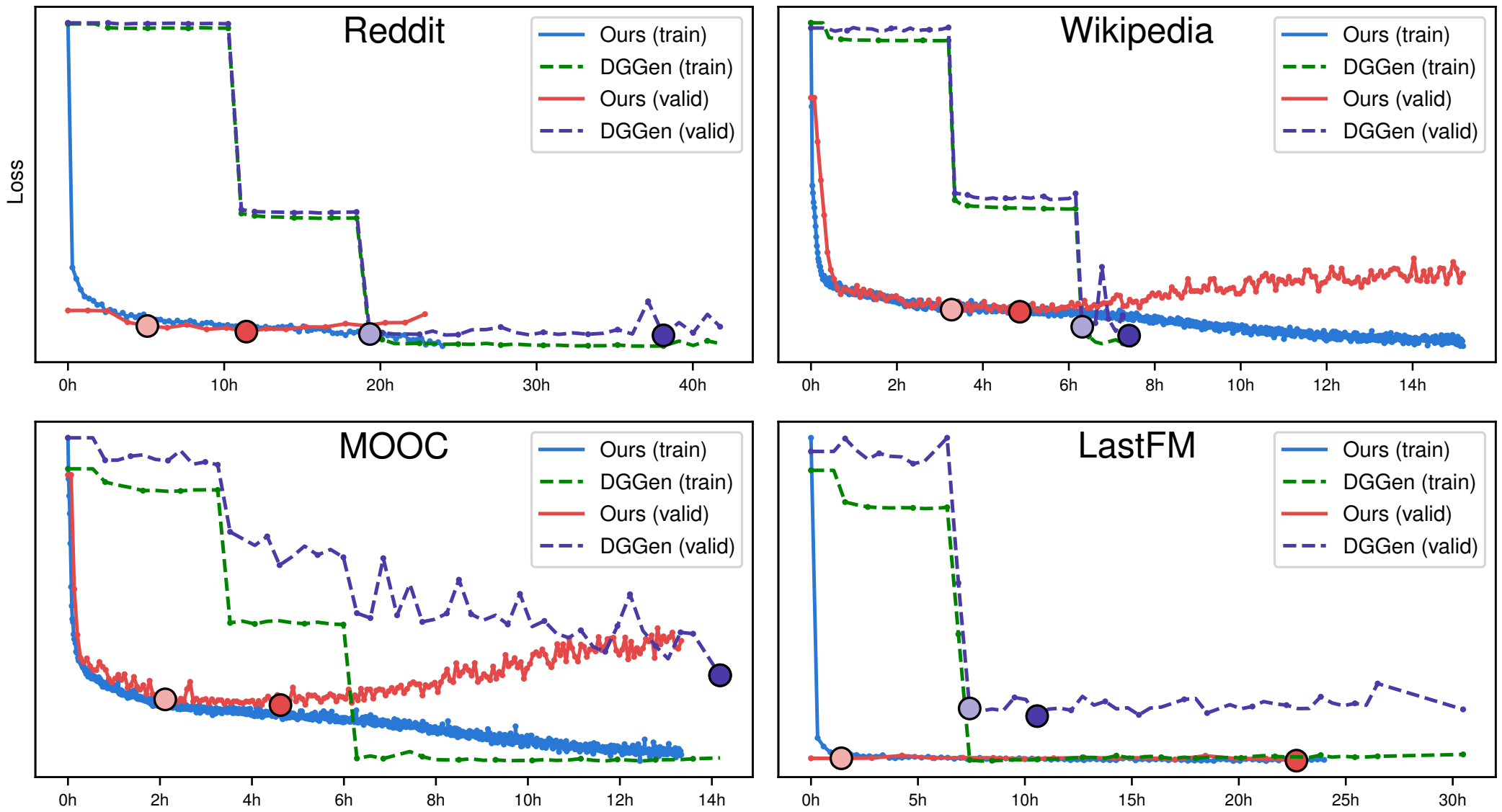}}
   \vspace{-20pt}
   \caption{
     \textbf{Convergence speed}.
      Colored dots (\protect\capdot{oursValidDot}\,\protect\capdot{dggenValidDot}) indicate the best validation loss, while the lighter dots (\protect\capdot{oursWithinDot}\,\protect\capdot{dggenWithinDot}) indicate the first value within 5\% of the best validation loss.
      Loss magnitudes should not be compared across models.
   }
   \label{fig:speed_by_time_c12}
\end{figure}

\begin{figure}[t]
   \centering
   \resizebox{\textwidth}{!}{\input{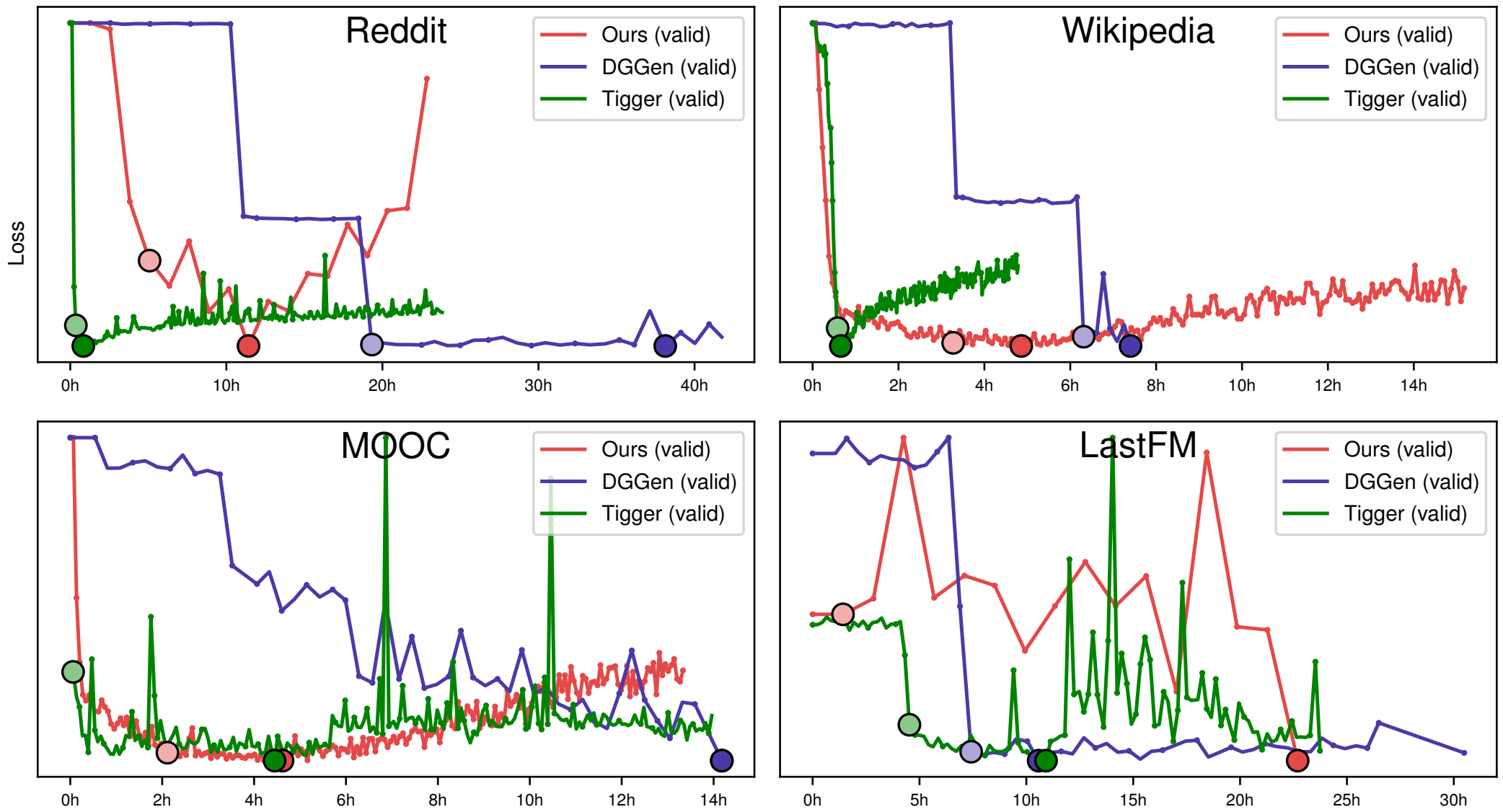}}
   \vspace{-20pt}
   \caption{
     \textbf{Convergence speed}.
      Colored dots (\protect\capdot{oursValidDot}\,\protect\capdot{dggenValidDot}\,\protect\capdot{tiggerValidDot}) indicate the best validation loss, while the lighter dots (\protect\capdot{oursWithinDot}\,\protect\capdot{dggenWithinDot}\,\protect\capdot{tiggerWithinDot}) indicate the first value within 5\% of the best validation loss.
      Loss magnitudes should not be compared across models.
      For visibility, we omit the training loss and show only the validation loss.
   }
   \label{fig:speed_by_time_c12_tigger}
\end{figure}

\Cref{fig:speed_by_time_c12} compares the convergence speed of \modelname and \dggen trained on the first 12 features.
\modelname reaches its best validation loss substantially earlier in training time than \dggen, and, unlike \dggen, does not require a noise scheduler to stabilize feature-generation convergence \citep{hosseiniDeepProbabilisticFramework2025}.
This training-speed gap widens further when scaling to a higher number of features (\cref{scaling_to_more_features}).
\modelname exploits two levels of parallelism, generating all events and all features simultaneously, whereas \dggen relies on an RNN to generate features sequentially and processes only a small number of events in parallel during training.
\Cref{fig:speed_by_time_c12_tigger} additionally reports \tigger for comparison.
Note that \tigger does not generate features and follows a multi-component training pipeline. We report only its main training stage, excluding the GraphSAGE and WGAN components of \tiggerind.

\subsubsection{Scaling to More Features}
When scaling to a higher number of features, the convergence-speed gap between \modelname and \dggen widens further, as shown in \cref{fig:speed_by_time_call}.
As above, this is due to \modelname's two levels of parallelism, generating all windows and all features simultaneously, whereas \dggen relies on an RNN to generate features sequentially and processes only a small number of events in parallel during training.

\begin{figure}[t]
   \centering
   \resizebox{\textwidth}{!}{\input{figures/ours_vs_dggen_call_by_time.pgf}}
   \vspace{-20pt}
   \caption{
     \textbf{Convergence speed}.
      Colored dots (\protect\capdot{oursValidDot}\,\protect\capdot{dggenValidDot}) indicate the best validation loss, while the lighter dots (\protect\capdot{oursWithinDot}\,\protect\capdot{dggenWithinDot}) indicate the first value within 5\% of the best validation loss.
      Loss magnitudes should not be compared across models.
   }
   \label{fig:speed_by_time_call}
\end{figure}

\subsection{Generation Speed vs.\ Window Size}
\label{results:generation_speed:diff_conf}

In \cref{fig:speed_by_gen_size}, we compare the generation speed of \modelname for different $L$ on the \reddits dataset, fixing $M=4096$ and, more importantly, initializing $\zsum$ to all $10{,}984$ nodes of \reddits (\cref{tab:dataset-stats}).
Initializing $\zsum$ to such a high value gives a worst-case scenario, since earlier in inference the model typically operates with a smaller $\zsum$.
This also explains why, in \cref{speed_vs_other}, \reddits is slower than its \redditf counterpart. We find that on \redditf \modelname tends to reuse past nodes more, whereas on \reddits it creates more new nodes early in the generation, causing $\zsum$ to grow sooner.
Even for smaller window sizes, \modelname remains highly efficient in generation speed.

\subsection{VRAM Consumption}
\label{results:vram_consumption}

In \cref{fig:vram_by_config}, we report the peak VRAM consumption of \modelname on \reddits across $z$ generation window sizes $L$ and $\zpast$ context sizes $M$, both ranging from $256$ to $4096$, keeping $\zsum$ full (\cref{results:generation_speed:diff_conf}).

Across the entire grid, the peak footprint remains modest, from $0.74$~GB for the smallest configuration ($L=M=256$) to $3.52$~GB for the largest ($L=M=4096$), so that even the most demanding setting fits comfortably on a single consumer-grade GPU.

The generation window size $L$ is the dominant driver, as it sets the transformer sequence length. At fixed $M=256$, increasing $L$ from $256$ to $4096$ raises the footprint by roughly $4.4\times$, whereas increasing $M$ over the same range at fixed $L=256$ raises it by only about $2.5\times$.

Overall, \modelname keeps memory requirements low even in its worst-case configuration, confirming its scalability to large generation windows and long context lengths.

\begin{figure}[t]
   \centering
   \setlength{\speedfigheight}{0.39\textwidth}%
   \begin{minipage}[t]{0.53\textwidth}
      \centering
      \resizebox{!}{\speedfigheight}{\input{figures/event_reddit_gen_speed_only.pgf}}
      \caption{\textbf{Generation speed} (events/s) of \modelname on \reddits as a function of the $z$ generation window size $L$, with $M=4096$ and $\zsum$ initialized to all $10{,}984$ nodes (worst case). \modelname stays fast even for small window sizes.}
      \label{fig:speed_by_gen_size}
   \end{minipage}
   \hfill
   \begin{minipage}[t]{0.45\textwidth}
      \centering
      \resizebox{!}{\speedfigheight}{\input{figures/event_reddit_gen_speed_only_vram_grid.pgf}}
      \caption{\textbf{Peak VRAM consumption} of \modelname on \reddits as a function of the $z$ generation window size $L$ and $\zpast$ the context size $M$ with $\zsum$ full. Even the largest configuration stays below $3.5$~GB.}
      \label{fig:vram_by_config}
   \end{minipage}
\end{figure}

\subsection{Topology Quality vs.\ Window Size}
\label{appendix:topo_window_size}
\Cref{tab:topo_all_ws} extends \cref{tab:topo_main} by reporting the topological metrics of \tigger, \dggen, and \modelname across time-snapshot window sizes from 512 to 4096\rebuttal{\,s on the JODIE datasets, and from 6\,h to 2\,d on Enron, 3\,h to 1\,d on UCI, 1\,h to 8\,h on SocialEvo, and 10 to 80\,min on Contacts}.
The results are consistent with \cref{tab:topo_main} at every resolution, with \modelname often outperforming \tigger and \dggen. On Reddit, its errors remain within the scale of the real statistics, and it obtains the best value on at least eight of the nine metrics.
\rebuttal{On the DyGLib datasets (\cref{tab:topo_all_ws_dgb}), \modelname outperforms \tigger at every window size except the coarsest one on SocialEvo. Its advantage over \dggen grows with the window size on Enron and UCI, and narrows on SocialEvo and Contacts, where its windows aggregate into too many wedges and an oversized component at coarse resolutions.}

\begin{table}[t]
   \centering
   \caption{
      \textbf{Topology generation quality across snapshot window sizes.}
      Extension of \cref{tab:topo_main} to window sizes 512--4096\rebuttal{\,s on the JODIE datasets}.
      Each cell reports the median absolute error between the generated and real per-snapshot statistics. The \textcolor{\pctofcolor}{(percentage)} beside it rescales this error by the magnitude of the real statistic, and the \emph{Real Data} rows give the median of each real statistic over windows.
   }
   
\providecommand{\currentrowstyle}{}
\providecommand{\resetrowstyle}{\global\let\currentrowstyle\relax}
\providecommand{\rowstyle}[1]{\gdef\currentrowstyle{#1}#1\ignorespaces}
\newcommand{\ncols}{22}
{\setlength{\tabcolsep}{3pt}
\begin{tabular}{
  >{\resetrowstyle}c %
  >{\currentrowstyle}l %
  Z %
  >{\currentrowstyle}l %
  Z %
  >{\currentrowstyle}T@{\,}>{\currentrowstyle}l   %
  >{\currentrowstyle}T@{\,}>{\currentrowstyle}l   %
  >{\currentrowstyle}T@{\,}>{\currentrowstyle}l   %
  >{\currentrowstyle}S[round-mode=places,round-precision=0,table-format=5.0]@{\,}>{\currentrowstyle}l   %
  >{\currentrowstyle}T@{\,}>{\currentrowstyle}l   %
  >{\currentrowstyle}S[round-mode=places,round-precision=0,table-format=3.0]@{\,}>{\currentrowstyle}l   %
  >{\currentrowstyle}S[round-mode=places,round-precision=0,table-format=3.0]@{\,}>{\currentrowstyle}l   %
  >{\currentrowstyle}T@{\,}>{\currentrowstyle}l   %
  Z}  %
\toprule
  {} & {\colmodel} & {\coldataset} & {\colwindow} & \multicolumn{1}{Z}{\colskiprate} & \multicolumn{2}{c}{\coldegreeMedaeMean $\downarrow$} & \multicolumn{2}{c}{\colclosenessMedaeMean $\downarrow$} & \multicolumn{2}{c}{\colbetweennessMedaeMean $\downarrow$} & \multicolumn{2}{c}{\colwedgeCountMedae $\downarrow$} & \multicolumn{2}{c}{\colplmedae $\downarrow$} & \multicolumn{2}{c}{\colnComponentsMedae $\downarrow$} & \multicolumn{2}{c}{\collccMedae $\downarrow$} & \multicolumn{2}{c}{\colrelEdgeDistrEntropyMedae $\downarrow$} & {\coljlSimilarity $\uparrow$} \\
\midrule

\multirow{18}{*}{\rotatebox{90}{Reddit}}
& Real Data & reddit & 512 & {-} & 1.3246753215789795 & {-} & 0.013884562067687511 & {-} & 9.621482604416089e-05 & {-} & 178 & {-} & 3.1764779090881348 & {-} & 58 & {-} & 14 & {-} & 0.9566155672073364 & {-} & {-} \\
\cdashline{2-\ncols}

& \tigger & reddit & 512 & 0.8574938774108887 & 1.1650450229644775 & \pctof{87.94947742936536} & 0.20860935747623444 & \pctof{1502.4554354632126} & 0.0041036298498511314 & \pctof{4265.070175325826} & 9855 & \pctof{5536.516853932584} & 0.8541185855865479 & \pctof{26.88885646403686} & 29 & \pctof{50.0} & 351 & \pctof{2507.1428571428573} & 0.10907500982284546 & \pctof{11.402178007751843} & 0.11496290564537048 \\
\bottomrule
\end{tabular}}

   \label{tab:topo_all_ws}
\end{table}

\begin{table}[t]
   \centering
   \caption{
      \rebuttal{\textbf{Topology generation quality across snapshot window sizes on the DyGLib datasets.}
      Extension of \cref{tab:topo_main} to four window sizes per dataset, with the same layout as \cref{tab:topo_all_ws}.}
   }
   
\providecommand{\currentrowstyle}{}
\providecommand{\resetrowstyle}{\global\let\currentrowstyle\relax}
\providecommand{\rowstyle}[1]{\gdef\currentrowstyle{#1}#1\ignorespaces}
\newcommand{\ncols}{22}
{\setlength{\tabcolsep}{3pt}
\begin{tabular}{
  >{\resetrowstyle}c %
  >{\currentrowstyle}l %
  Z %
  >{\currentrowstyle}l %
  Z %
  >{\currentrowstyle}T@{\,}>{\currentrowstyle}l   %
  >{\currentrowstyle}T@{\,}>{\currentrowstyle}l   %
  >{\currentrowstyle}T@{\,}>{\currentrowstyle}l   %
  >{\currentrowstyle}S[round-mode=places,round-precision=0,table-format=5.0]@{\,}>{\currentrowstyle}l   %
  >{\currentrowstyle}T@{\,}>{\currentrowstyle}l   %
  >{\currentrowstyle}S[round-mode=places,round-precision=0,table-format=3.0]@{\,}>{\currentrowstyle}l   %
  >{\currentrowstyle}S[round-mode=places,round-precision=0,table-format=3.0]@{\,}>{\currentrowstyle}l   %
  >{\currentrowstyle}T@{\,}>{\currentrowstyle}l   %
  Z}  %
\toprule
  {} & {\colmodel} & {\coldataset} & {\colwindow} & \multicolumn{1}{Z}{\colskiprate} & \multicolumn{2}{c}{\coldegreeMedaeMean $\downarrow$} & \multicolumn{2}{c}{\colclosenessMedaeMean $\downarrow$} & \multicolumn{2}{c}{\colbetweennessMedaeMean $\downarrow$} & \multicolumn{2}{c}{\colwedgeCountMedae $\downarrow$} & \multicolumn{2}{c}{\colplmedae $\downarrow$} & \multicolumn{2}{c}{\colnComponentsMedae $\downarrow$} & \multicolumn{2}{c}{\collccMedae $\downarrow$} & \multicolumn{2}{c}{\colrelEdgeDistrEntropyMedae $\downarrow$} & {\coljlSimilarity $\uparrow$} \\
\midrule

\rowstyle{\rebuttalcolor}\multirow{16}{*}{\rotatebox{90}{Enron}}
& Real Data & Enron & 6\,h & {-} & 1.454545497894287 & {-} & 0.2113971710205078 & {-} & 0.007521367631852627 & {-} & 6 & {-} & 2.689798355102539 & {-} & 3 & {-} & 4 & {-} & 0.9492163062095642 & {-} & {-} \\
\cdashline{2-\ncols}

\rowstyle{\rebuttalcolor}& \tigger & Enron & 6\,h & {-} & 0.4017094373703003 & \pctof{27.61752299614182} & 0.3207327723503113 & \pctof{151.72046570064873} & 0.12449496239423752 & \pctof{1655.2170893363514} & 19 & \pctof{316.66666666666663} & 0.5376443862915039 & \pctof{19.98827849944938} & 5 & \pctof{166.66666666666669} & 7 & \pctof{175.0} & 0.05061793327331543 & \pctof{5.332602584066882} & {-} \\
\bottomrule
\end{tabular}}

   \label{tab:topo_all_ws_dgb}
\end{table}

\subsection{Ablation Study}
\label{ablation}
In \cref{tab:ablation} we ablate the components of \modelname by removing them one at a time from the token representation and the conditioning.
Starting from the full model, we successively strip the random Gaussian embedding for new nodes ($-\xi$), the event-class information ($-\vc$), the graph-level summary ($-\zsum$), and the past-event context ($-\zpast$).
Each row omits every component listed in its label.
Feature-quality metrics are essentially unchanged across all variants, as expected. Events can be generated almost independently, one by one, without information about the rest of the dynamic graph.
The ablation should instead surface in the global JL similarity and in the per-window topology metrics.
Two distinct behaviours emerge. On Reddit and MOOC, the JL similarity declines as components are removed while the topology metrics stay plausible, whereas on Wikipedia and LastFM the JL similarity is largely unchanged but the power-law exponent of the node degree (\powerlaw) collapses.

On Reddit and MOOC, every feature and topology metric remains plausible, yet the JL similarity degrades as more components are removed.
Since JL embeds the full event stream whereas the topology metrics are computed per window, this indicates that the generated windows stay individually realistic but no longer compose into a coherent continuous graph. Each window statistic is plausible, but their joint distribution deteriorates with every removed component.

On Wikipedia and LastFM the picture is reversed, as at the ablation steps where \powerlaw collapses, the JL similarity stays similar to the full-model value.
On Wikipedia the collapse coincides with removing the event-class information $\vc$ ($-\vc-\xi$ row), confirming that $\vc$ is what mitigates \memcollapse by routing each generated embedding to the correct pool, an already-seen node or a new one.
\rebuttal{The same ablation on the synthetic dataset of \cref{appendix:synthetic:ablation} isolates this mechanism. Without $\xi$ the recall swaps the new nodes of a window, without $\vc$ the new events reuse existing nodes, and without both the generation collapses onto a single node per window.}
On LastFM the same collapse appears only once the graph-level summary $\zsum$ is removed, suggesting that on this dataset most nodes are matched to identifiers outside the recent past-event context, so the model must fall back on the global summary to resolve them.
Because \powerlaw collapses while the JL similarity barely moves, JL alone does not capture this failure mode, a limitation of the metric and a reminder that the global and per-window statistics are complementary rather than interchangeable.

\begin{table}[t]
   \centering
   \caption{
      \textbf{Ablation study on \modelname.}
      Feature quality (Jensen--Shannon distance), per-snapshot topology, and the global JL similarity as the components of \modelname are removed one at a time.
      Each row omits every component listed in its label, namely the random embedding $\xi$, the event-class information $\vc$, the graph-level summary $\zsum$, and the past context $\zpast$.
      Topology cells report the median absolute error to the real per-snapshot statistic, with the \textcolor{\pctofcolor}{(percentage)} rescaling it by the magnitude of that statistic. The \emph{Real Data} rows give the median of each statistic over windows.
   }
   
\newcommand{\ncols}{29}
{\setlength{\tabcolsep}{3pt}
\begin{tabular}{c l Z Z
  ZZ %
  F @{$\,\pm\,$} F   %
  ZZ %
  ZZ %
  |
  T@{\,}l   %
  T@{\,}l   %
  T@{\,}l   %
  S[round-mode=places,round-precision=0,table-format=5.0]@{\,}l   %
  T@{\,}l   %
  S[round-mode=places,round-precision=0,table-format=3.0]@{\,}l   %
  S[round-mode=places,round-precision=0,table-format=3.0]@{\,}l   %
  T@{\,}l   %
  |
  J}  %
\toprule
{} & {\colmodel} & {\coldataset} & {\colwindow} &

  \multicolumn{2}{Z}{\colperfeatjsmean $\downarrow$} &
  \multicolumn{2}{c}{\colpairjsmean $\downarrow$} &
  \multicolumn{2}{Z}{\colfeatureMmdMean $\downarrow$} &
  \multicolumn{2}{Z|}{\colfeatureWassersteinMean $\downarrow$} &

  \multicolumn{2}{c}{\coldegreeMedaeMean $\downarrow$} &
  \multicolumn{2}{c}{\colclosenessMedaeMean $\downarrow$} &
  \multicolumn{2}{c}{\colbetweennessMedaeMean $\downarrow$} &
  \multicolumn{2}{c}{\colwedgeCountMedae $\downarrow$} &
  \multicolumn{2}{c}{\colplmedae $\downarrow$} &
  \multicolumn{2}{c}{\colnComponentsMedae $\downarrow$} &
  \multicolumn{2}{c}{\collccMedae $\downarrow$} &
  \multicolumn{2}{c|}{\colrelEdgeDistrEntropyMedae $\downarrow$} &
  {\coljlSimilarity $\uparrow{}$} \\
\midrule

\multirow{6}{*}{\rotatebox{90}{Reddit}} & Real Data & Reddit & 512 & {-} & {-} & {-} & {-} & {-} & {-} & {-} & {-} & 1.3246753215789795 & & 0.013884562067687511 & & 9.621482604416089e-05 & & 178 & & 3.1764779090881348 & & 58 & & 14 & & 0.9566155672073364 & & \\

\cdashline{2-\ncols}

& \modelname & reddit & 512 & 0.1025594174861908 & 0.1182892769575119 & 0.1906837970018386 & 0.123350016772747 & 0.043416604399681 & 0.1464645117521286 & 0.0632427260279655 & 0.0323162041604518 & 0.148996353149414 & \pctof{11.247763940511414} & 0.0038975095376372 & \pctof{28.07081360317163} & 7.208663737401366e-05 & \pctof{74.92258764873428} & 146 & \pctof{82.02247191011236} & 0.2990965843200683 & \pctof{9.4159818793114} & 15 & \pctof{25.86206896551724} & 7 & \pctof{50.0} & 0.0274417400360107 & \pctof{2.868627793306968} & \multicolumn{1}{>{\cellcolor{green!100!red!40!white}}J}{0.2574158012866974} \\

& \modelname - $\xi$ & Reddit & 512 & 0.1122628152370452 & 0.1150602623820304 & 0.2006172388792038 & 0.1202331334352493 & 0.0437731184065341 & 0.1469084173440933 & 0.0731079280376434 & 0.0376615114510059 & 0.0796676874160766 & \pctof{6.014129358212456} & 0.0047316746786236 & \pctof{34.078674253870375} & 6.689869769616053e-05 & \pctof{69.53054996477904} & 129 & \pctof{72.47191011235955} & 0.1704728603363037 & \pctof{5.366725827010112} & 16 & \pctof{27.58620689655172} & 7 & \pctof{50.0} & 0.0206507444381713 & \pctof{2.158729707740116} & \multicolumn{1}{>{\cellcolor{green!48.8!red!40!white}}J}{0.19024658203125} \\

& \modelname - $\vc$ - $\xi$ & Reddit & 512 & 0.0998357608914375 & 0.1213684603571891 & 0.1882880181074142 & 0.1268174201250076 & 0.0434438996016979 & 0.1472025662660598 & 0.0541990585625171 & 0.0258980505168437 & 0.225171685218811 & \pctof{16.99825470821122} & 0.0071343090385198 & \pctof{51.38303249133796} & 8.435251220362261e-05 & \pctof{87.67101253699332} & 150 & \pctof{84.26966292134831} & 0.9204301834106444 & \pctof{28.97643899166516} & 27 & \pctof{46.55172413793103} & 7 & \pctof{50.0} & 0.0341014862060546 & \pctof{3.5648056936401025} & \multicolumn{1}{>{\cellcolor{green!58.8!red!40!white}}J}{0.2033393532037735} \\

& \modelname - $\vc$ - $\xi$ - $\zsum$ & reddit & 512 & 0.1022274494171142 & 0.11215490847826 & 0.1875669956207275 & 0.115741916000843 & 0.0428219698369503 & 0.1459431052207946 & 0.0499803274869918 & 0.020114479586482 & 0.1900045871734619 & \pctof{14.343483575053035} & 0.0055079795420169 & \pctof{39.66981108345712} & 6.832882354501635e-05 & \pctof{71.01693819376095} & 145 & \pctof{81.46067415730337} & 0.5644552707672119 & \pctof{17.769847199385968} & 13 & \pctof{22.41379310344828} & 7 & \pctof{50.0} & 0.0318713784217834 & \pctof{3.331680929553142} & \multicolumn{1}{>{\cellcolor{green!19.3!red!40!white}}J}{0.1514575183391571} \\

& \modelname - $\vc$ - $\xi$ - $\zsum$ - $\zpast$ & reddit & 512 & 0.0989647805690765 & 0.120869368314743 & 0.1879329830408096 & 0.127447172999382 & 0.0436065085232257 & 0.1474248319864273 & 0.0536326467990875 & 0.0286633390933275 & 0.1393781900405883 & \pctof{10.521686919814668} & 0.006172249559313 & \pctof{44.454045645971554} & 6.582296191481873e-05 & \pctof{68.41249381317508} & 126 & \pctof{70.78651685393258} & 0.2880749702453613 & \pctof{9.069005939602407} & 22 & \pctof{37.93103448275862} & 6 & \pctof{42.85714285714285} & 0.0292650461196899 & \pctof{3.059227460109589} & \multicolumn{1}{>{\cellcolor{green!0!red!40!white}}J}{0.126164898276329} \\

\midrule

\multirow{6}{*}{\rotatebox{90}{Wiki.}} & Real Data & Wikipedia & 512 & {-} & {-} & {-} & {-} & {-} & {-} & {-} & {-} & 1.037037014961243 & & 0.035125453025102615 & & 6.167128594825044e-05 & & 1 & & 5.088466644287109 & & 15 & & 3 & & 0.9963162541389464 &  &  \\
\cdashline{2-\ncols}

& \modelname & Wikipedia & 512 & 0.0922721102833748 & 0.1487192213535308 & 0.1720019876956939 & 0.1779591292142868 & 0.2548881471157074 & 0.348622053861618 & 0.0543153844773769 & 0.0203846618533134 & 0.0368342399597168 & \pctof{3.551873214582741} & 0.0112863313406705 & \pctof{32.13148975645878} & 8.517721289535984e-05 & \pctof{138.11486429330122} & 1 & \pctof{100.0} & 1.255626916885376 & \pctof{24.67593883699887} & 5 & \pctof{33.33333333333333} & 1 & \pctof{33.33333333333333} & 0.0038322806358337 & \pctof{0.3846449980027415} & 0.1339774280786514 \\
& \modelname  - $\xi$ & wikipedia & 512 & 0.1114971563220024 & 0.1721076667308807 & 0.2039113193750381 & 0.2036808580160141 & 0.232744038105011 & 0.337967574596405 & 0.0887663289904594 & 0.0439266711473465 & 0.0952380895614624 & \pctof{9.183673117494434} & 0.0103058293461799 & \pctof{29.340061005945856} & 0.0002013490448007 & \pctof{326.48750825415226} & 5 & \pctof{500.0} & 1.5596256256103516 & \pctof{30.65020829725521} & 5 & \pctof{33.33333333333333} & 1 & \pctof{33.33333333333333} & 0.0089092254638671 & \pctof{0.8942166131342374} & 0.0604263171553611 \\
& \modelname - $\vc$ - $\xi$ & Wikipedia & 512 & 0.0993034616112709 & 0.1538864821195602 & 0.1823306530714035 & 0.1839354485273361 & 0.2243028134107589 & 0.3441550135612488 & 0.063096597790718 & 0.0175915565341711 & 0.0322580337524414 & \pctof{3.110596178058985} & 0.0112539902329444 & \pctof{32.039416615927365} & 6.10948191024363e-05 & \pctof{99.06525891758142} & 1 & \pctof{100.0} & \multicolumn{1}{>{\cellcolor{red!40!white}}T}{89.51190948486328} & \pctof{1759.1136140267229} & 6 & \pctof{40.0} & 1 & \pctof{33.33333333333333} & 0.0033724904060363 & \pctof{0.3384959737459075} & 0.1273448914289474 \\

& \modelname - $\vc$ - $\xi$ - $\zsum$ & wikipedia & 512 & 0.1140952482819557 & 0.166060522198677 & 0.2018931210041046 & 0.1977711617946624 & 0.2143279314041137 & 0.3464490175247192 & 0.0700465217232704 & 0.0187098979949951 & 0.0322580337524414 & \pctof{3.110596178058985} & 0.0143589712679386 & \pctof{40.87910626427192} & 6.10948191024363e-05 & \pctof{99.06525891758142} & 1 & \pctof{100.0} & \multicolumn{1}{>{\cellcolor{red!40!white}}T}{89.44422912597656} & \pctof{1757.7835402811734} & 9 & \pctof{60.0} & 1 & \pctof{33.33333333333333} & 0.0033716559410095 & \pctof{0.3384122187109586} & 0.1514523178339004 \\
& \modelname - $\vc$ - $\xi$ - $\zsum$ -$\zpast$ & wikipedia & 512 & 0.1008586585521698 & 0.1634153872728347 & 0.1864599734544754 & 0.1953510791063308 & 0.43600594997406 & 0.352239578962326 & 0.0915868356823921 & 0.0462510474026203 & 0.034433364868164 & \pctof{3.3203602543975674} & 0.0117113217711448 & \pctof{33.34141132009117} & 6.10948191024363e-05 & \pctof{99.06525891758142} & 1 & \pctof{100.0} & \multicolumn{1}{>{\cellcolor{red!40!white}}T}{89.6581039428711} & \pctof{1761.9866692755363} & 6 & \pctof{40.0} & 1 & \pctof{33.33333333333333} & 0.00337415933609 & \pctof{0.3386634838158052} & 0.1456869095563888 \\

\midrule

\multirow{6}{*}{\rotatebox{90}{MOOC}} & Real Data & MOOC & 512 & {-} & {-} & {-} & {-} & {-} & {-} & {-} & {-} & 1.9666666984558103 & & 0.12778709828853607 & & 0.0224645733833313 & & 133 & & 2.087773084640503 & & 6 & & 36 & & 0.9536862373352052 &  &  \\
\cdashline{2-\ncols}

& \modelname & MOOC & 512 & 0.0651027262210846 & 0.0437499545514583 & 0.1101563796401024 & 0.0376386642456054 & 0.567058801651001 & 0.3797310888767242 & 0.0980294719338417 & 0.0446901284158229 & 0.5 & \pctof{25.423728402610905} & 0.1060688719153404 & \pctof{83.00436690082975} & 0.0204569343477487 & \pctof{91.06308852910507} & 129 & \pctof{96.99248120300751} & 0.6165134906768799 & \pctof{29.529717343924776} & 45 & \pctof{750.0} & 27 & \pctof{75.0} & 0.0202569365501403 & \pctof{2.1240671991600104} & \multicolumn{1}{>{\cellcolor{green!100!red!40!white}}J}{0.3212104737758636} \\
& \modelname - $\xi$ & mooc & 512 & 0.069129504263401 & 0.0456628873944282 & 0.1170213893055915 & 0.0422805212438106 & 0.5687575340270996 & 0.3779482245445251 & 0.1349517703056335 & 0.1174800097942352 & 0.7034631967544556 & \pctof{35.76931451103543} & 0.106255255639553 & \pctof{83.15022178501518} & 0.0221734158694744 & \pctof{98.70392591531282} & 148 & \pctof{111.2781954887218} & 0.8350462913513184 & \pctof{39.99698518458036} & 55 & \pctof{916.6666666666666} & 40 & \pctof{111.11111111111111} & 0.025583803653717 & \pctof{2.6826227172160344} & \multicolumn{1}{>{\cellcolor{green!56!red!40!white}}J}{0.2422850728034973} \\
& \modelname - $\vc$ - $\xi$ & mooc & 512 & 0.056776225566864 & 0.0419961251318454 & 0.0909146070480346 & 0.0362424999475479 & 0.5724009871482849 & 0.3688097894191742 & 0.148957222700119 & 0.1369787156581878 & 0.6226092576980591 & \pctof{31.658097337333277} & 0.1188864558935165 & \pctof{93.03478792912068} & 0.0227357968688011 & \pctof{101.20733868763814} & 110 & \pctof{82.70676691729322} & 0.7964985370635986 & \pctof{38.15062771540371} & 67 & \pctof{1116.6666666666663} & 27 & \pctof{75.0} & 0.0250440835952758 & \pctof{2.626029674629067} & \multicolumn{1}{>{\cellcolor{green!2.3!red!40!white}}J}{0.145823672413826} \\

& \modelname - $\vc$ - $\xi$ -$\zsum$ & mooc & 512 & 0.0737310200929641 & 0.0411136262118816 & 0.1201851889491081 & 0.031036140397191 & 0.5690380334854126 & 0.3766962885856629 & 0.1303839087486267 & 0.1209490373730659 & 0.6666666269302368 & \pctof{33.89830251631814} & 0.1146857738494873 & \pctof{89.74753741612732} & 0.0223398245871067 & \pctof{99.44468655560064} & 102 & \pctof{76.69172932330827} & 0.8614041805267334 & \pctof{41.25947340082028} & 63 & \pctof{1050.0} & 27 & \pctof{75.0} & 0.0265510082244873 & \pctof{2.7840401994975057} & \multicolumn{1}{>{\cellcolor{green!22.3!red!40!white}}J}{0.181798979640007} \\
& \modelname - $\vc$ - $\xi$ -$\zsum$ -$\zpast$ & mooc & 512 & 0.0544684268534183 & 0.0343302749097347 & 0.0876466855406761 & 0.0234232917428016 & 0.577008843421936 & 0.3628275990486145 & 0.1677072644233703 & 0.0930826440453529 & 0.6713178157806396 & \pctof{34.13480364048193} & 0.1169465854763984 & \pctof{91.5167392034677} & 0.022358126938343 & \pctof{99.5261586179632} & 103 & \pctof{77.44360902255639} & 0.851738452911377 & \pctof{40.79650509806429} & 62 & \pctof{1033.3333333333337} & 28 & \pctof{77.77777777777779} & 0.0268114805221557 & \pctof{2.8113523580955238} & \multicolumn{1}{>{\cellcolor{green!0!red!40!white}}J}{0.1416743099689483} \\

\midrule

\multirow{6}{*}{\rotatebox{90}{LastFM}} & Real Data & LastFM & 512 & {-} & {-} & {-} & {-} & {-} & {-} & {-} & {-} & 1.0526316165924072 & & 0.1428571492433548 & & 0.0003418803389649838 & & 1 & & 4.654970645904541 & & 5 & & 3 & & 0.993847131729126 &  &  \\
\cdashline{2-\ncols}

& \modelname & LastFM & 512 & {-} & {-} & {-} & {-} & {-} & {-} & {-} & {-} & 0.0 & \pctof{0.0} & 0.2666666805744171 & \pctof{186.66666805744163} & 0.0 & \pctof{0.0} & 0.0 & \pctof{0.0} & 0.0 & \pctof{0.0} & 3.0 & \pctof{60.0} & 0.0 & \pctof{0.0} & 0.0001128315925598 & \pctof{0.0113530128485159} & 0.1742327362298965 \\

& \modelname  - $\xi$ & LastFM & 512 & {-} & {-} & {-} & {-} & {-} & {-} & {-} & {-} & 0.04347825050354 & \pctof{4.1304336501204855} & 0.6666666269302368 & \pctof{466.6666179895423} & 0.0002224694035248 & \pctof{65.0723010859273} & 0 & \pctof{0.0} & 0.6636345386505127 & \pctof{14.256470966887418} & 3 & \pctof{60.0} & 0 & \pctof{0.0} & 0.0052927732467651 & \pctof{0.5325540596526758} & 0.0866998583078384 \\

& \modelname - $\vc$ - $\xi$ & LastFM & 512 & {-} & {-} & {-} & {-} & {-} & {-} & {-} & {-} & 0.0 & \pctof{0.0} & 0.6666666269302368 & \pctof{466.6666179895423} & 0.0 & \pctof{0.0} & 0 & \pctof{0.0} & 0.0 & \pctof{0.0} & 3 & \pctof{60.0} & 0 & \pctof{0.0} & 0.0001041293144226 & \pctof{0.0104773974888311} & 0.1871466785669326 \\

& \modelname - $\vc$ - $\xi$ -$\zsum$ & lastfm & 512 & {-} & {-} & {-} & {-} & {-} & {-} & {-} & {-} & 0.0526316165924072 & \pctof{5.000003397464631} & 0.6666666269302368 & \pctof{466.6666179895423} & 0.0002801120572257 & \pctof{81.93277743721566} & 1 & \pctof{100.0} & \multicolumn{1}{>{\cellcolor{red!40!white}}T}{90.34503173828124} & \pctof{1940.829247070916} & 3 & \pctof{60.0} & 1 & \pctof{33.33333333333333} & 0.0060085654258728 & \pctof{0.6045764216694891} & 0.1851364523172378 \\
& \modelname - $\vc$ - $\xi$ -$\zsum$ -$\zpast$ & lastfm & 512 & {-} & {-} & {-} & {-} & {-} & {-} & {-} & {-} & 0.0370370149612426 & \pctof{3.518516295486105} & 0.6666666269302368 & \pctof{466.6666179895423} & 0.0001139601154136 & \pctof{33.33333404273921} & 1 & \pctof{100.0} & \multicolumn{1}{>{\cellcolor{red!40!white}}T}{89.82171630859375} & \pctof{1929.5871690967415} & 3 & \pctof{60.0} & 1 & \pctof{33.33333333333333} & 0.0039610862731933 & \pctof{0.3985609201589925} & 0.1694341897964477 \\

\bottomrule
\end{tabular}}

   \label{tab:ablation}
\end{table}

\begingroup\rebuttalcolor
\subsection{Synthetic Dataset for Node Assignment}
\label{appendix:synthetic}
To assess the ability of \modelname to accurately create new nodes and reference them later, we create a deterministic dataset that serves as a synthetic test bed, illustrated in \cref{fig:synthetic}.
In each window, new nodes are created in a controlled way and then referenced again, both later in the same window and in future windows.
This makes it possible to check whether seen nodes, new nodes, and recalled new nodes are matched to the right identifier, covering the three cases of the event class $\vc$ (\cref{eq:tokenization:with_class}).
Additionally, random features control the order in which nodes reappear, which makes it possible to evaluate feature validity, the interaction between nodes and features, and stochastic generation.

\begin{figure}[ht]
   \centering
   {\color{black}\begin{tikzpicture}[
    font=\sffamily,
    >={Stealth[length=9,width=7]},
    box/.style={draw=ink, line width=1pt, rounded corners=3, inner sep=0pt},
    eventbox/.style={box, dash pattern=on 4pt off 4pt},
    event/.style={circle, draw=ink, line width=1pt, minimum size=9mm,
                  inner sep=0pt, font=\Large},
    edge/.style={draw=ink, line width=1pt},
    timeline/.style={->, draw=ink, line width=0.8},
    lbl/.style={font=\sffamily\footnotesize, inner sep=1pt, scale=1.5},
    blocklbl/.style={font=\sffamily\footnotesize, inner sep=1pt, scale=2.0, anchor=north},
    rowlbl/.style={font=\sffamily\footnotesize, scale=2.2, anchor=south},
    bracket/.style={draw=ink, line width=0.8pt},
    recur/.style={->, draw=ink, line width=0.9pt, dash pattern=on 3pt off 2pt},
    feat/.style={font=\sffamily\footnotesize, inner sep=1pt, scale=1.35, anchor=west},
    sep/.style={draw=ink!60, line width=0.3pt, dash pattern=on 1pt off 1.5pt},
]

\def\L{10}          %
\def\K{4}           %
\def\N{2}           %
\def\R{4}           %
\def\evDx{1.6}      %
\def\winGap{0.9}    %
\def\pairDx{0}      %
\def\pairRise{1.45} %
\def\axisY{-1.1}    %
\def\evTick{0.13}   %

\newcommand{\ev}[6]{%
  \pgfmathsetmacro\ex{#1*\evDx + floor(#1/\L)*\winGap}%
  \coordinate (s#1) at (\ex,0);%
  \coordinate (d#1) at ($(s#1)+(\pairDx,\pairRise)$);%
  \draw[edge] (s#1) -- (d#1);%
  \node[event, fill=#4] at (s#1) {#2};%
  \node[event, fill=#5] at (d#1) {#3};%
  \node[feat] (f#1) at ($(s#1)!0.5!(d#1)+(0.12,0)$) {$#6$};%
  \draw[edge] ($(s#1)+(0.5*\pairDx,\axisY-\evTick)$) -- ($(s#1)+(0.5*\pairDx,\axisY+\evTick)$);%
}
\definecolor{hueA}{HTML}{4C78A8}
\definecolor{hueB}{HTML}{F58518}
\definecolor{hueC}{HTML}{54A24B}
\definecolor{hueD}{HTML}{B279A2}
\definecolor{hueE}{HTML}{E45756}
\definecolor{hueF}{HTML}{72B7B2}
\newcommand{\pairhue}[2]{\colorlet{#1s}{#2!70!white}\colorlet{#1d}{#2!30!white}}
\pairhue{p2n1}{hueE}
\pairhue{p2n2}{hueF}
\pairhue{p1n1}{hueA}
\pairhue{p1n2}{hueB}
\pairhue{p0n1}{hueC}
\pairhue{p0n2}{hueD}
\newcommand{\nodeu}[2]{\scalebox{0.8}{$u^{#1}_{#2}$}}
\newcommand{\nodev}[2]{\scalebox{0.8}{$v^{#1}_{#2}$}}

\foreach \j/\m/\f in {0/2/-1,1/1/+1,2/2/-1,3/1/+1} {\ev{\j}{\nodeu{w-2}{\m}}{\nodev{w-2}{\m}}{p2n\m s}{p2n\m d}{\f}}
\ev{4}{\nodeu{w-1}{1}}{\nodev{w-1}{1}}{p1n1s}{p1n1d}{\boldsymbol{-1}}
\ev{5}{\nodeu{w-1}{2}}{\nodev{w-1}{2}}{p1n2s}{p1n2d}{\boldsymbol{+1}}
\foreach \j/\m/\f in {6/1/-1,7/2/+1,8/1/-1,9/2/+1} {\ev{\j}{\nodeu{w-1}{\m}}{\nodev{w-1}{\m}}{p1n\m s}{p1n\m d}{\f}}

\foreach \j/\m/\f in {10/1/-1,11/2/+1,12/1/-1,13/2/+1} {\ev{\j}{\nodeu{w-1}{\m}}{\nodev{w-1}{\m}}{p1n\m s}{p1n\m d}{\f}}
\ev{14}{\nodeu{w}{1}}{\nodev{w}{1}}{p0n1s}{p0n1d}{\boldsymbol{+1}}
\ev{15}{\nodeu{w}{2}}{\nodev{w}{2}}{p0n2s}{p0n2d}{\boldsymbol{-1}}
\foreach \j/\m/\f in {16/2/-1,17/1/+1,18/2/-1,19/1/+1} {\ev{\j}{\nodeu{w}{\m}}{\nodev{w}{\m}}{p0n\m s}{p0n\m d}{\f}}
\foreach \j in {4,15} {\node[circle, draw=ink, fill=gray!20, line width=0.6pt, inner sep=-1.5pt, xshift=1pt, fit=(f\j)] {};
                        \node[feat, anchor=center] at (f\j.center) {$\boldsymbol{-1}$};}

\node[eventbox, inner sep=0.65cm, fit=(s0) (d9) (f9)]   (win0) {};
\node[eventbox, inner sep=0.65cm, fit=(s10) (d19) (f19)] (win1) {};
\def\topY{0.25}
\newcommand{\tblk}[3]{%
  \coordinate (ta) at ($(s#1 |- win0.north)+(-0.35,\topY)$);%
  \coordinate (tb) at ($(s#2 |- win0.north)+(\pairDx+0.35,\topY)$);%
  \draw[bracket] ($(ta)+(0,-0.18)$) -- (ta) -- (tb) -- ($(tb)+(0,-0.18)$);%
  \node[blocklbl, anchor=south] at ($(ta)!0.5!(tb)+(0,0.05)$) {#3};%
}
\tblk{0}{3}{$K = \K$}
\tblk{4}{5}{$n = \N$}
\tblk{6}{9}{$r = \R$}
\node[rowlbl] at ($(win0.north)+(0,2.6)$) {window $w-1$};
\node[rowlbl] at ($(win1.north)+(0,2.6)$) {window $w$};

\node[font=\large] at ($(win0.west)+(-0.7,0)$) {$\cdots$};
\node[font=\large] at ($(win1.east)+(0.7,0)$) {$\cdots$};

\path let \p1 = (win0.north), \p2 = (win0.south) in
  \pgfextra{\xdef\sepYloA{\y2}\xdef\sepYhiA{\y1}};
\path let \p1 = (win1.north), \p2 = (win1.south) in
  \pgfextra{\xdef\sepYloB{\y2}\xdef\sepYhiB{\y1}};
\foreach \sepIdx/\sepYlo/\sepYhi in {3/\sepYloA/\sepYhiA, 5/\sepYloA/\sepYhiA, 13/\sepYloB/\sepYhiB, 15/\sepYloB/\sepYhiB} {
  \pgfmathsetmacro\sepX{\sepIdx*\evDx + floor(\sepIdx/\L)*\winGap + 0.5*\evDx}
  \draw[sep] (\sepX,\sepYlo) -- (\sepX,\sepYhi);
}

\draw[timeline] ($(win0.west |- 0,\axisY)+(-1.3,0)$) -- ($(win1.east |- 0,\axisY)+(1.5,0)$);

\def\brY{-1.65}
\newcommand{\blk}[4]{%
  \coordinate (#3a) at ($(s#1)+(-0.35,\brY)$);%
  \coordinate (#3b) at ($(s#2)+(\pairDx+0.35,\brY)$);%
  \draw[bracket] ($(#3a)+(0,0.18)$) -- (#3a) -- (#3b) -- ($(#3b)+(0,0.18)$);%
  \coordinate (#3) at ($(#3a)!0.5!(#3b)$);%
  \node[blocklbl] at ($(#3)+(0,-0.08)$) {#4};%
}
\blk{0}{3}{seenA}{seen}
\blk{4}{5}{newA}{new}
\blk{6}{9}{againA}{recall}
\blk{10}{13}{seenB}{seen}
\blk{14}{15}{newB}{new}
\blk{16}{19}{againB}{recall}

\def\arY{-1.0}
\draw[recur] ($(newA |- win0.north)+(0.4,1.05)$) to[out=25, in=155] node[lbl, above, scale=1.3] {recalled, event with feature $-1$ first}
            ($(againA |- win0.north)+(0,1.05)$);
\draw[recur] ($(newA)+(-0.2,\arY)$) .. controls ($(newA)+(0.5,\arY-2.8)$) and ($(seenB)+(-1.5,\arY-2.6)$)
            .. node[lbl, below, scale=1.3] {replayed, event with feature $-1$ first, as the seen block of the next window}
            ($(seenB)+(0,\arY)$);
\draw[recur] ($(newB |- win1.north)+(0.4,0.25)$) to[out=25, in=155] node[lbl, above, scale=1.3] {recalled, event with feature $-1$ first}
            ($(againB |- win1.north)+(0,0.25)$);
\end{tikzpicture}}
   \caption{\rebuttal{
      \textbf{Two consecutive windows $w-1$ and $w$ of the synthetic dataset} (\cref{eq:synthetic}), with $L = 10$, $K = 4$, $n = 2$ and $r = 4$.
      The source node $u^w_i$ and the destination node $v^w_i$ are the $i$-th new nodes created in window $w$, so the superscript is the window where the node was created and the subscript its order of appearance inside that window.
      The numbers next to the events are features.
      The recall and seen blocks start with the new node of feature $-1$, which appears second in window $w-2$ (not shown), first in window $w-1$ and second in window $w$.
   }}
   \label{fig:synthetic}
\end{figure}
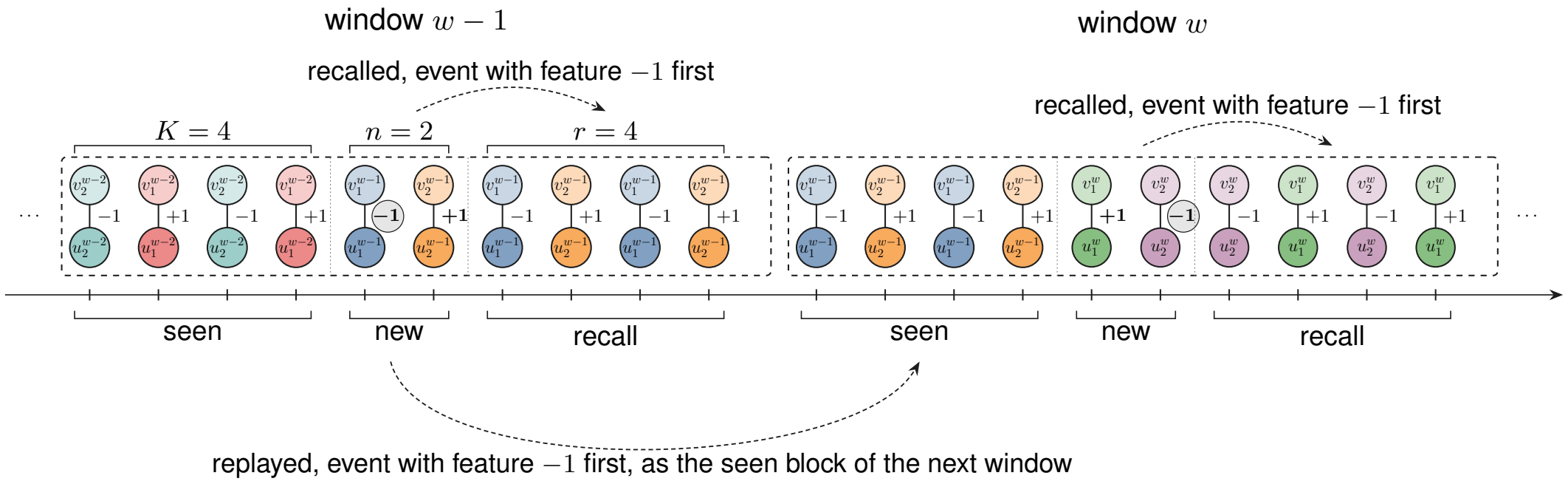

\subsubsection{Synthetic Dataset Description}
The dataset is a stream of events split into windows of size $L = K + n + r$ ($L=30$ events), each composed of three parts.
The first part covers the reappearance of previously seen nodes ($K=10$ events), the second the creation of new nodes ($n=2$ events), and the third the reappearance of the nodes created in the window ($r=18$ events).
Controlled random features are assigned to each event so that generation is not fully deterministic, while the event order still depends on the feature values, which keeps the evaluation controlled.

Formally, window $w$ creates source nodes $u^w_1, \dots, u^w_n$ and destination nodes $v^w_1, \dots, v^w_n$, in this order of appearance, and gives the edge $(u^w_i, v^w_i)$ a feature $f^w_i$.
The features $(f^w_1, \dots, f^w_n)$ are a random permutation of $n$ values evenly spaced in $[-1, 1]$ ($-1$ and $1$ for $n = 2$), drawn in the new block of each window.
Let $\rho_w$ be the permutation of $\{1, \dots, n\}$ that sorts the new nodes of window $w$ by increasing feature value, $f^w_{\rho_w(1)} < \dots < f^w_{\rho_w(n)}$, and let $\pi(j) = (j \bmod n) + 1$ cycle through the ranks of the $n$ created nodes.
The $j$-th event of a seen or recall block references the edge with the $\pi(j)$-th smallest feature.
The $j$-th event of window $w$ is
\begin{equation}
\label{eq:synthetic}
   x(t_{(w \times L + j)}) =
   \left\{\begin{array}{@{(}l@{,\,}l@{,\,}l@{,\,}l@{)}l@{\quad}l@{}}
      u^{w-1}_{i} & v^{w-1}_{i} & t_{(w \times L + j)} & f^{w-1}_{i} & ,\, i = \rho_{w-1}(\pi(j)) & \text{if } 0 \le j < K \quad \text{(seen)}, \\
      u^{w}_{i} & v^{w}_{i} & t_{(w \times L + j)} & f^{w}_{i} & ,\, i = j-K+1 & \text{if } K \le j < K+n \quad \text{(new)}, \\
      u^{w}_{i} & v^{w}_{i} & t_{(w \times L + j)} & f^{w}_{i} & ,\, i = \rho_{w}(\pi(j-K-n)) & \text{if } K+n \le j < L \quad \text{(recall)},
   \end{array}\right.
\end{equation}
with $t_i = i$, so the events of a window are split into three blocks.
The experiments use $L = 30$, $n = 2$, $K = 10$, $r = 18$ and $30{,}000$ events ($3{,}000$ windows). Training uses the first $21{,}000$ events ($700$ windows), and $4{,}500$ events ($150$ windows) are generated for evaluation.

\subsubsection{Controlled Generation Results}
\label{appendix:synthetic:results}

\textbf{Evaluation protocol:}
Given its own past, the synthetic stream is fixed up to the random order of the features of the new nodes, so a generated stream can be scored against the rule of \cref{eq:synthetic} without being aligned to real data.
Each generated window is checked against the nodes and features that the generated stream itself produced.

Let $B$ be the set of indices $k$ of the generated events of one block (seen, new or recall) in a window.
We write $\hat{x}(t_k) = (\hat{\src}_k, \hat{\dst}_k, t_k, \hat{\efeat}_k)$ for the $k$-th generated event and $x^\star(t_k) = (\src^\star_k, \dst^\star_k, t_k, \efeat^\star_k)$ for the event that \cref{eq:synthetic} expects given the generated past.

For the seen and recall blocks, where the rules of \cref{eq:synthetic} are fully deterministic, $\mathrm{Acc}$ verifies that the nodes are referenced in the right order.
The node accuracy $\mathrm{Acc}(B)$ and the feature accuracy $\mathrm{Acc}_f(B)$ are defined as
\begin{equation}
\label{eq:synthetic:metrics}
   \mathrm{Acc}(B) = \frac{1}{|B|} \sum_{k \in B} \mathbf{1}\big[\hat{\src}_k = \src^\star_k\big],
   \qquad
   \mathrm{Acc}_f(B) = \frac{1}{|B|} \sum_{k \in B} \mathbf{1}\big[\operatorname{sign}(\hat{\efeat}_k) = \efeat^\star_k\big],
\end{equation}
where the expected feature $\efeat^\star_k \in \{-1, 1\}$ is compared with the sign of the generated one.

The new block creates new nodes and introduces randomness into the stream, so it has no ground truth for its nodes and features and the metrics of \cref{eq:synthetic:metrics} do not apply.
The fresh accuracy $\mathrm{Acc}_{\text{fresh}}(B)$ is instead the fraction of new events whose node appears nowhere earlier in the generated stream, so the new nodes must also be distinct, and the feature accuracy of the block is $1$ if its two features have opposite signs and $0$ otherwise,
\begin{equation}
\label{eq:synthetic:fresh}
   \mathrm{Acc}_{\text{fresh}}(B) = \frac{1}{|B|} \sum_{k \in B} \mathbf{1}\big[\hat{\src}_k \notin \{\hat{\src}_{k'} : k' < k\}\big],
   \qquad
   \mathrm{Acc}_f(B) = \mathbf{1}\big[\operatorname{sign}(\hat{\efeat}_{k_1}) \neq \operatorname{sign}(\hat{\efeat}_{k_2})\big].
\end{equation}

On the seen and recall blocks, a low $\mathrm{Acc}$ does not tell whether the model references the wrong nodes or the right nodes but in the wrong order.
We therefore introduce two metrics that ignore positions within a block, so that errors on node identity are separated from errors on node order.

On the seen block, the pair accuracy $\mathrm{Acc}_{\text{pair}}(B)$ is the fraction of events whose $\hat{\src}_k$ is one of the two new nodes of the previous window,
\begin{equation}
\label{eq:synthetic:pair}
   \mathrm{Acc}_{\text{pair}}(B) = \frac{1}{|B|} \sum_{k \in B} \mathbf{1}\big[\hat{\src}_k \in \hat{N}\big],
\end{equation}
where $\hat{N} = \{\hat{\src}_k : k \in B'\}$ gathers the generated nodes of the new block $B'$ of the previous window.

On the recall block, the set accuracy $\mathrm{Acc}_{\text{set}}(B)$ is the fraction of events matched to an expected node once the order is discarded, each new node being expected as many times as the rule reuses it,
\begin{equation}
\label{eq:synthetic:set}
   \mathrm{Acc}_{\text{set}}(B) = \frac{1}{|B|} \sum_{v} \min\big(\hat{c}_v, c^\star_v\big),
\end{equation}
where the sum runs over the new nodes $v$ of the window.
For each such node, $\hat{c}_v = |\{k \in B : \hat{\src}_k = v\}|$ is the number of times $v$ is generated in the block, and $c^\star_v = |\{k \in B : \src^\star_k = v\}|$ is the number of times the rule expects it.
Taking their minimum counts each generated occurrence of $v$ as correct only up to the number of times it is expected.
All node accuracies are computed on both source nodes $\hat{\src}_k$ and destination nodes $\hat{\dst}_k$ and averaged over the two sides.

In the data, the features $(f^w_1, f^w_2)$ of the two new events are a random permutation of $\{-1, 1\}$, so either new event may carry the feature $+1$.
A model could instead always give the feature $+1$ to the first new event and still pass these checks.
The share of windows whose first new event has the feature $+1$ measures this and should be close to $0.5$, as should the share of the feature $+1$ over all events.

\textbf{Results:}
Experiments on this simple synthetic dataset show that, unlike other methods, the formulation of \modelname enables fine-grained control over generation.
Models that generate a bag of independent events, such as \dggen, cannot solve the task, since no event attends to the others, which breaks the strong conditioning the task requires.

\Cref{tab:synthetic} scores the windows generated by \modelname and \dggen.
\tigger generates no features and no order within a window, so the permutation $\rho_w$ of its windows is undefined and it is only compared through the window statistics of \cref{tab:synthetic:window}.
\modelname follows the rule in every window, as the real stream does, with all its node and feature accuracies equal to $1$.
It creates fresh nodes, reuses them in the right order within the window and recalls them in the right order in the next one, with every feature of the right sign.
It also randomizes which new event carries the feature $+1$, giving it to the first one in $55\%$ of the windows ($45\%$ in the data), so it reads $\rho_w$ from the features instead of learning a fixed order.
\dggen does not follow the rule.
More than half of its new events reuse a node already in the stream ($\mathrm{Acc}_{\text{fresh}} = 0.43$), and fewer than one seen event in ten recalls the right node at the right position ($0.10$ order accuracy), against all of them for \modelname.
Its feature accuracy $\mathrm{Acc}_f$ on the seen and recall blocks is at chance ($0.50$ and $0.51$), and the feature $+1$ makes only a quarter of its events ($0.24$).
This failure stems from the training and sampling scheme of \dggen, which generates events independently as a bag, without information about one another. This independent generation breaks the conditioning, whereas the joint generation of \modelname enables precise conditioning.

\Cref{tab:synthetic:window} compares the three models with statistics that need neither the features or the order of the events.
The number of events, nodes and fresh nodes per window, and the share of the new nodes of the previous window that the window recalls.
\tigger generated and merge random walks and cannot reproduce the structure of the stream.
It generates about $8$ events per window instead of $30$, and creates about $2$ fresh nodes per window instead of $4$.
It also recalls few of the new nodes of the previous window ($0.12$ against $1$ in the data).
\dggen keeps the number of events but spreads them over about $33$ nodes per window instead of $8$, most of them fresh, and almost never recalls the new nodes of the previous window.
\modelname matches the data on every statistic.

\begin{table}[ht]
   \centering
   \caption{\rebuttal{
      \textbf{Rule accuracy on the synthetic dataset} (\cref{eq:synthetic:metrics}), on the $150$ generated windows.
      Topology columns give the node accuracy of each block.
      Feature columns give the feature accuracy $\mathrm{Acc}_f$ of each block (\cref{eq:synthetic:metrics,eq:synthetic:fresh}), then \emph{$+1$ first}, the share of windows whose first new event has the feature $+1$, and \emph{$+1$ share}, the share of events with the feature $+1$.
   }}
   
{\setlength{\tabcolsep}{4pt}
\sisetup{round-mode=places,round-precision=2,table-format=1.2,detect-weight=true}
\begin{tabular}{l SSSSS | SSS SS}
\toprule
& \multicolumn{5}{c|}{Topology, node accuracy $\uparrow$} & \multicolumn{5}{c}{Features} \\
\cmidrule(lr){2-6}\cmidrule(lr){7-11}
& \multicolumn{2}{c}{seen} & {new} & \multicolumn{2}{c|}{recall}
& {seen $\uparrow$} & {new $\uparrow$} & {recall $\uparrow$}
& {$+1$ first} & {$+1$ share} \\
& {order} & {pair} & {fresh} & {order} & {set}
& \multicolumn{3}{c}{$\mathrm{Acc}_f$}
& \multicolumn{2}{c}{target $0.5$} \\
\midrule
Real data & 1.0000 & 1.0000 & 1.0000 & 1.0000 & 1.0000 & 1.0000 & 1.0000 & 1.0000 & 0.4533 & 0.5000 \\
\midrule
\dggen & 0.0993 & 0.1775 & 0.4300 & 0.1767 & 0.2724 & 0.5000 & 0.2600 & 0.5115 & 0.2067 & 0.2424 \\
\modelname & \bfseries 1.0000 & \bfseries 1.0000 & \bfseries 1.0000 & \bfseries 1.0000 & \bfseries 1.0000 & \bfseries 1.0000 & \bfseries 1.0000 & \bfseries 1.0000 & \bfseries 0.5467 & \bfseries 0.5000 \\
\bottomrule
\end{tabular}}

   \label{tab:synthetic}
\end{table}

\begin{table}[ht]
   \centering
   \caption{\rebuttal{
      \textbf{Window statistics on the synthetic dataset}.
      A window is $30$ consecutive events, or one timestamp for \tigger.
      Nodes are counted on both sides, and a node is fresh if it appears nowhere earlier in the generated stream.
      Previous new nodes recalled is the share of the fresh nodes of the previous window that appear in the window.
   }}
   
{\setlength{\tabcolsep}{5pt}
\sisetup{round-mode=places,round-precision=2,table-format=2.2,detect-weight=true}
\begin{tabular}{l SSS S[table-format=1.2] c}
\toprule
& {events} & {nodes} & {fresh nodes} & {previous new} & {features} \\
& {per window} & {per window} & {per window} & {nodes recalled} & \\
\midrule
Real data & 30.0000 & 8.0000 & 4.0267 & 0.9966 & yes \\
\midrule
\tigger & 8.0933 & 6.1667 & 2.1400 & 0.1231 & no \\
\dggen & \bfseries 30.0000 & 32.8400 & 26.7933 & 0.0073 & yes \\
\modelname & \bfseries 30.0000 & \bfseries 8.0933 & \bfseries 4.1200 & \bfseries 0.9945 & yes \\
\bottomrule
\end{tabular}}

   \label{tab:synthetic:window}
\end{table}

\subsubsection{Independent Event Generation with \modelname}
\label{appendix:synthetic:indep}
\modelname generates the $L$ events of a window jointly.
To assess the contribution of this joint generation, two variants generate each event independently of the other events of its window, with the same trained model.
In both, $L$ windows are generated from the same context, each from a different initial noise, and a single event is kept from each of them, so that, as in \dggen, no kept event attends to another kept event (\cref{fig:synthetic:indep}).
The \emph{column independent} variant keeps the first event of every window.
The \emph{diagonal independent} variant keeps the $i$-th event of the $i$-th window, so each kept event retains its position in the window, which removes the bias of always sampling the first position and is closer to the joint generation.
\Cref{tab:synthetic:indep} evaluates both variants with the metrics of \cref{appendix:synthetic:results}.
As these metrics score each window against the nodes of the generated stream itself, a stream collapsed onto a single node can still reach high seen and recall accuracies.
We therefore add the mean number of distinct source nodes per window, which is $4$ in the data (the two new nodes of the window and the two of the previous window) and exposes such collapse.

\begin{figure}[ht]
   \centering
   {\color{black}\begin{tikzpicture}[
    font=\sffamily,
    cell/.style={draw=ink, line width=0.6pt, minimum size=6mm, inner sep=0pt},
    drop/.style={cell, fill=black!6},
    keep/.style={cell, fill=red!55, text=white, font=\sffamily\bfseries\small},
    ctx/.style={draw=ink, line width=1pt, rounded corners=3, fill=beige,
                minimum width=12mm, minimum height=36mm, align=center, font=\sffamily\small},
    arr/.style={->, >={Stealth[length=5,width=4]}, draw=ink, line width=0.6pt},
    lbl/.style={font=\sffamily\small},
    title/.style={font=\sffamily\bfseries},
]
\def\L{6}
\def\cs{0.6}
\newcommand{\panel}[4]{%
  \begin{scope}[xshift=#1cm]
    \node[ctx] (#2ctx) at (-2.35,{-(\L-1)*\cs/2}) {same\\context};
    \foreach \k in {1,...,\L} {
      \pgfmathsetmacro\y{-(\k-1)*\cs}
      \node[lbl, anchor=east] (#2eps\k) at (-0.45,\y) {$\boldsymbol{\epsilon}_{\k}$};
      \draw[arr] (#2ctx.east) -- (#2eps\k.west);
      \foreach \j in {1,...,\L} {
        \pgfmathsetmacro\x{(\j-1)*\cs}
        \pgfmathtruncatemacro\hit{(#4==1 && \j==\k) || (#4==0 && \j==1) ? 1 : 0}
        \ifnum\hit=1
          \node[keep] (#2c\k-\j) at (\x,\y) {};
        \else
          \node[drop] (#2c\k-\j) at (\x,\y) {};
        \fi
      }
    }
    \foreach \j in {1,...,\L} {
      \node[lbl, font=\sffamily\scriptsize] at ({(\j-1)*\cs},0.5) {\j};
    }
    \node[lbl, anchor=south] at ({(\L-1)*\cs/2},0.75) {event position $j$ in the window};
    \node[lbl, font=\sffamily\scriptsize, anchor=east] at (-0.45,0.5) {window $k$};
    \node[title, anchor=south] at ({(\L-1)*\cs/2-0.9},1.3) {#3};
    \pgfmathsetmacro\ky{-(\L-1)*\cs-1.4}
    \foreach \k in {1,...,\L} {
      \pgfmathtruncatemacro\pos{#4==1 ? \k : 1}
      \node[keep] (#2k\k) at ({(\k-1)*\cs},\ky) {\pos};
    }
    \draw[arr, line width=0.9pt] ({(\L-1)*\cs/2},{-(\L-1)*\cs-0.4}) -- ({(\L-1)*\cs/2},\ky+0.38);
    \node[lbl, anchor=north] at ({(\L-1)*\cs/2},\ky-0.4) {generated window};
  \end{scope}
}
\panel{0}{col}{Column independent}{0}
\panel{7.6}{dia}{Diagonal independent}{1}
\end{tikzpicture}}
   \caption{\rebuttal{
      \textbf{Independent event generation}, drawn with $L = 6$.
      The $L$ windows (rows) are generated independently, each from its own initial Gaussian noise $\boldsymbol{\epsilon}_k$, with the same context as conditioning.
      Only the red event of each window is kept, the first one for the column independent variant (left) and the $k$-th one of the $k$-th window for the diagonal independent variant (right).
      The kept events form the generated window, where each number is the position at which the event was generated.
   }}
   \label{fig:synthetic:indep}
\end{figure}
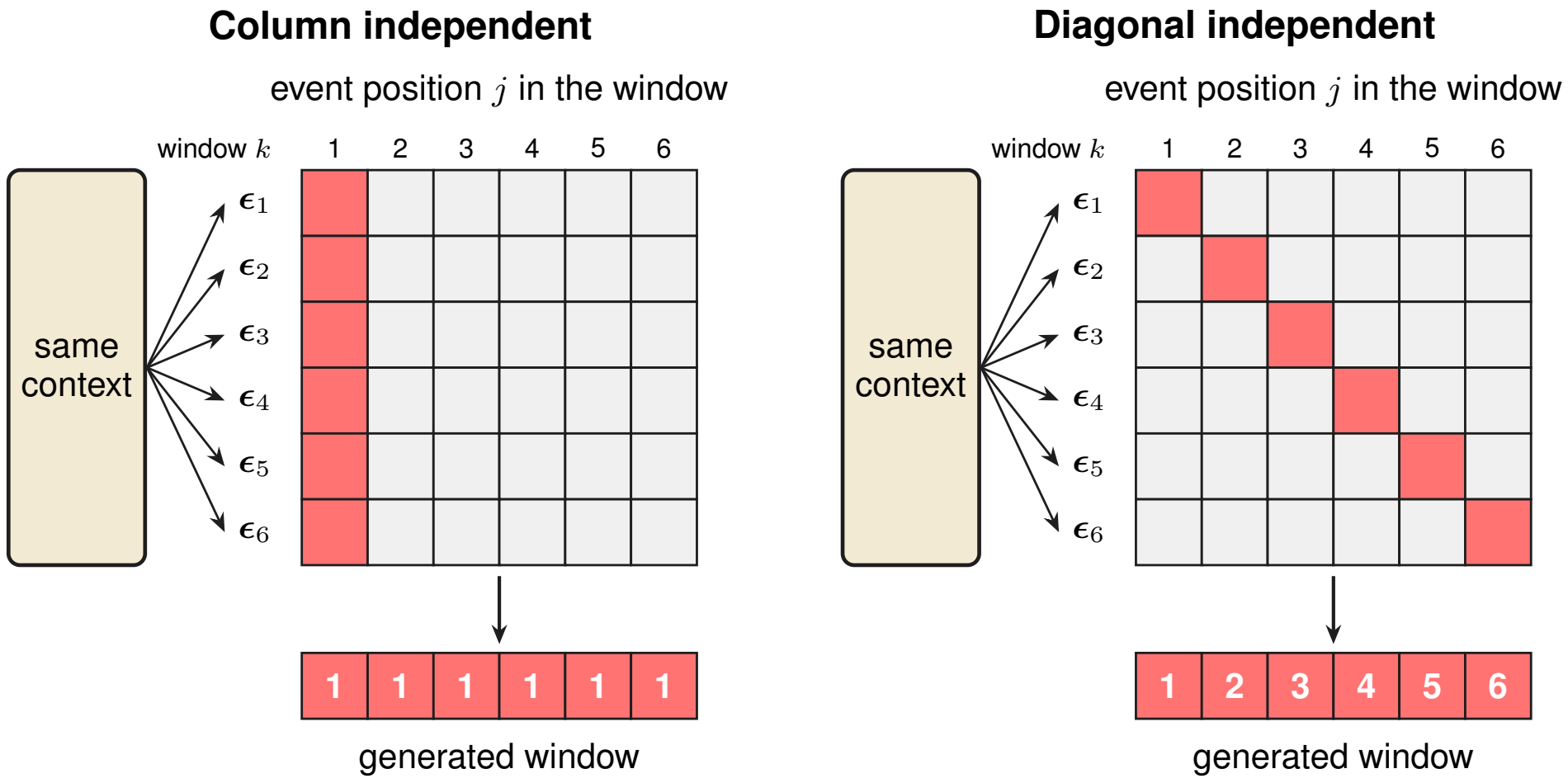
Both experiments show the benefit of joint event-window diffusion over an event-independent diffusion strategy, as independently sampled events cannot achieve the precise conditioning needed to solve the task.

\textbf{Column independent event generation:}
Every kept event is generated at the first position of a window. Consequently, the $L$ kept events are the same event, namely the seen event that recalls the new node of feature $-1$ of the previous window.
The whole stream therefore repeats this node, with $1.12$ distinct nodes per window against $4$ in the data, and only $1\%$ of events with the feature $+1$.
No new node is created ($\mathrm{Acc}_{\text{fresh}} = 0.003$), and the generated windows hold a single node, so the new nodes they are scored against are the same node.
The high seen and recall accuracies (above $0.99$, in grey) come from this degenerate stream and do not reflect a correct recall.

\textbf{Diagonal independent event generation:}
Each kept event is generated at its own position, so the stream keeps the block structure, with $3.61$ distinct nodes per window.
The new events create fresh nodes ($\mathrm{Acc}_{\text{fresh}} = 1$), and the seen events reuse the new nodes of the previous window ($\mathrm{Acc}_{\text{pair}} = 0.99$), with their features alternating in sign as expected.
The two new events of a window, however, no longer see each other.
They draw their features independently and take opposite signs in only half of the windows ($\mathrm{Acc}_f = 0.51$ on the new block), and each seen or recall event chooses between the two new nodes without knowing which one received which feature.
The two new nodes are therefore not recalled exactly as many times each, which lowers $\mathrm{Acc}_{\text{set}}$ to $0.90$, the value expected from independent fair choices.
The order accuracy of the seen and recall blocks thus falls to chance ($0.50$ and $0.51$, against $1$ for the joint generation).
Matching each node to its feature, as the order requires, therefore needs the events of a window to be generated jointly.

\begin{table}[ht]
   \centering
   \caption{\rebuttal{
      \textbf{Joint versus independent event generation on the synthetic dataset} (\cref{eq:synthetic:metrics}), with the same \modelname model and the columns of \cref{tab:synthetic}.
      Nodes per window is the mean number of distinct source nodes in a window.
      Grey values come from a stream collapsed onto a single node per window.
      Best values in bold, taken as the closest to $4$ for nodes per window and to $0.5$ for the two balance columns.
   }}
   
{\setlength{\tabcolsep}{4pt}
\sisetup{round-mode=places,round-precision=2,table-format=1.2,detect-weight=true}
\begin{tabular}{l SSSSS S | SSS SS}
\toprule
& \multicolumn{6}{c|}{Topology} & \multicolumn{5}{c}{Features} \\
\cmidrule(lr){2-7}\cmidrule(lr){8-12}
& \multicolumn{2}{c}{seen $\uparrow$} & {new $\uparrow$} & \multicolumn{2}{c}{recall $\uparrow$} & {nodes per}
& {seen $\uparrow$} & {new $\uparrow$} & {recall $\uparrow$}
& {$+1$ first} & {$+1$ share} \\
& {order} & {pair} & {fresh} & {order} & {set} & {window}
& \multicolumn{3}{c}{$\mathrm{Acc}_f$}
& \multicolumn{2}{c}{target $0.5$} \\
\midrule
Real data & 1.0000 & 1.0000 & 1.0000 & 1.0000 & 1.0000 & 4.0000 & 1.0000 & 1.0000 & 1.0000 & 0.4533 & 0.5000 \\
\midrule
\modelname, joint window & \bfseries 1.0000 & \bfseries 1.0000 & \bfseries 1.0000 & \bfseries 1.0000 & \bfseries 1.0000 & \bfseries 4.0467 & \bfseries 1.0000 & \bfseries 1.0000 & \bfseries 1.0000 & \bfseries 0.5467 & \bfseries 0.5000 \\
\modelname, column independent & \color{gray} 0.9916 & \color{gray} 0.9933 & 0.0033 & \color{gray} 0.9915 & \color{gray} 0.9924 & 1.1200 & 0.4946 & 0.0333 & 0.5004 & 0.0067 & 0.0109 \\
\modelname, diagonal independent & 0.4966 & 0.9852 & \bfseries 1.0000 & 0.5083 & 0.8987 & 3.6133 & \bfseries 1.0000 & 0.5067 & 0.9996 & 0.4267 & 0.4980 \\
\bottomrule
\end{tabular}}

   \label{tab:synthetic:indep}
\end{table}

\subsubsection{Synthetic \modelname ablation}
\label{appendix:synthetic:ablation}
As in \cref{ablation}, the synthetic dataset is used to ablate the two components of \modelname that identify the nodes of a window, the random Gaussian embedding $\xi$ of the new nodes and the event-class information $\vc$.
Four models are trained, with or without each component.
\Cref{tab:synthetic:ablation} evaluates them with the metrics and the nodes per window of \cref{appendix:synthetic:results,appendix:synthetic:indep}.
The features are generated perfectly by every variant, as on the real datasets, so the ablation only affects the topology.
Both components are needed to solve the task, and only the full model follows the rule exactly.

\textbf{Without the Gaussian embedding ($-\xi$):}
The event class still tells a new node from a seen one, so the new events create fresh nodes ($\mathrm{Acc}_{\text{fresh}} = 1.00$).
However, all the new nodes now share the same training target, the memory of a node never seen ($\vzero$ embedding for the TGN encoder).
The recall events are matched against the embeddings generated for the new events of their window, which differ only through their initial noise, so a recall event can follow its new node only by copying this difference.
It mostly does so thanks to the joint generation, but exchanges the two new nodes in some windows ($0.94$ order accuracy).
These errors are written to the memory of the two nodes, so the seen events of the next window still recall the right pair ($\mathrm{Acc}_{\text{pair}} = 1.00$) but in the wrong order in about one window in ten ($0.90$ order accuracy).

\textbf{Without the event class ($-\vc$):}
The new nodes keep their Gaussian embedding, but nothing tells the model which events must create a node, and each generated embedding is matched against all the nodes.
Most new events then reuse a node already in the stream ($\mathrm{Acc}_{\text{fresh}} = 0.17$), and the two new events of a window often fall onto the same node. This lowers the number of distinct nodes per window to $3.28$ and breaks the seen block of the next window, whose order accuracy falls to $0.64$.
The recall block, which only copies the nodes of its own window, is little affected ($0.97$).

\textbf{Without both ($-\vc-\xi$):}
Nothing distinguishes a new node from a seen one anymore, and the model collapses onto a single node per window ($1.00$ node per window against $4$ in the data), which it repeats for every event, as the column independent generation of \cref{appendix:synthetic:indep}.
No new node is created ($\mathrm{Acc}_{\text{fresh}} = 0$), and the high seen and recall accuracies (in grey) come from this degenerate stream.
These results agree with the ablation of \cref{tab:ablation}, where removing $\vc$ and $\xi$ triggers \memcollapse on Wikipedia.

\begin{table}[ht]
   \centering
   \caption{\rebuttal{
      \textbf{Ablation of \modelname on the synthetic dataset} (\cref{eq:synthetic:metrics}), with the columns of \cref{tab:synthetic:indep}.
      Each row omits the components listed in its label, the random embedding $\xi$ of the new nodes and the event-class information $\vc$.
      Grey values come from a stream collapsed onto a single node per window.
      Best values in bold, taken as the closest to $4$ for nodes per window and to $0.5$ for the two balance columns.
   }}
   
{\setlength{\tabcolsep}{4pt}
\sisetup{round-mode=places,round-precision=2,table-format=1.2,detect-weight=true}
\begin{tabular}{l SSSSS S | SSS SS}
\toprule
& \multicolumn{6}{c|}{Topology} & \multicolumn{5}{c}{Features} \\
\cmidrule(lr){2-7}\cmidrule(lr){8-12}
& \multicolumn{2}{c}{seen $\uparrow$} & {new $\uparrow$} & \multicolumn{2}{c}{recall $\uparrow$} & {nodes per}
& {seen $\uparrow$} & {new $\uparrow$} & {recall $\uparrow$}
& {$+1$ first} & {$+1$ share} \\
& {order} & {pair} & {fresh} & {order} & {set} & {window}
& \multicolumn{3}{c}{$\mathrm{Acc}_f$}
& \multicolumn{2}{c}{target $0.5$} \\
\midrule
Real data & 1.0000 & 1.0000 & 1.0000 & 1.0000 & 1.0000 & 4.0000 & 1.0000 & 1.0000 & 1.0000 & 0.4533 & 0.5000 \\
\midrule
\modelname & \bfseries 1.0000 & \bfseries 1.0000 & \bfseries 1.0000 & \bfseries 1.0000 & \bfseries 1.0000 & \bfseries 4.0533 & \bfseries 1.0000 & \bfseries 1.0000 & \bfseries 1.0000 & 0.5600 & \bfseries 0.5004 \\
\modelname $-\,\xi$ & 0.9037 & \bfseries 0.9950 & \bfseries 0.9967 & 0.9380 & 0.9393 & 3.8667 & \bfseries 1.0000 & \bfseries 1.0000 & \bfseries 1.0000 & \bfseries 0.4467 & \bfseries 0.5004 \\
\modelname $-\,\vc$ & 0.6362 & 0.8060 & 0.1683 & 0.9687 & 0.9687 & 3.2800 & \bfseries 1.0000 & \bfseries 1.0000 & \bfseries 1.0000 & 0.6200 & \bfseries 0.5000 \\
\modelname $-\,\vc-\xi$ & \color{gray} 1.0000 & \color{gray} 1.0000 & 0.0000 & \color{gray} 1.0000 & \color{gray} 1.0000 & 1.0000 & \bfseries 1.0000 & \bfseries 1.0000 & \bfseries 1.0000 & 0.4200 & \bfseries 0.4998 \\
\bottomrule
\end{tabular}}

   \label{tab:synthetic:ablation}
\end{table}

\endgroup

\begingroup\rebuttaltwocolor
\subsection{Distribution Distances over the Windows}
\label{appendix:topo_dist}

The median error of \cref{tab:topo_main,tab:topo_all_ws,tab:topo_all_ws_dgb} compares each generated window with the real window at the same position.
The real stream is a single realisation, so two equally realistic streams already differ window by window, and this pairing adds the natural variability of the data to the error.
\Cref{tab:topo_dist_wasserstein,tab:topo_dist_ks,tab:topo_dist_wasserstein_dgb,tab:topo_dist_ks_dgb} remove this pairing.
For each metric, they compare the set of per-window values of the generated stream with the set of the real stream, using the Wasserstein distance and the Kolmogorov-Smirnov (KS) statistic.
Such distribution distances over the snapshots are used in the dynamic graph generation literature~\citep{zenoDYMONDDYnamicMOtifNoDes2021,zhangTGGANContinuoustimeTemporal2021}.
The generated graphs and window sizes are those of \cref{tab:topo_all_ws,tab:topo_all_ws_dgb}.

\modelname obtains the best distance in most cells, and its advantage is clearer than with the median error.
On the cells of \cref{tab:topo_main}, it outperforms \dggen on at least 44 of the 64 comparisons for each distance, against 36 for the median error.
The KS statistic of \tigger and \dggen often reaches $1$, which means that their per-window values do not overlap those of the real stream.

\begin{table}[t]
   \centering
   \caption{
      \textbf{Wasserstein distance between the generated and real per-window statistics, on the JODIE datasets.}
      Same generated graphs and window sizes as \cref{tab:topo_all_ws}. Best value in bold.
   }
   
\rebuttaltwocolor
{\setlength{\tabcolsep}{4pt}
\begin{tabular}{c l l TTTTTTTT}
\toprule
{} & {\colmodel} & {\colwindow} & {\nodedegree $\downarrow$} & {\clocen $\downarrow$} & {\btwc $\downarrow$} & {WC $\downarrow$} & {PLE $\downarrow$} & {NCC $\downarrow$} & {LCC $\downarrow$} & {REL $\downarrow$} \\
\midrule
\multirow{12}{*}{\rotatebox{90}{Reddit}}
& \tigger & 512 & 1.1483010053634644 & 0.2040309756994247 & 0.0041550267487764 & 10626.853515625 & 0.8603728413581848 & 29.68103408813477 & 366.9827575683594 & 0.1075826585292816 \\
& \dggen & 512 & 0.6042024493217468 & 0.2694260776042938 & 0.0105590187013149 & 3988.677734375 & 0.2612590789794922 & 48.584716796875 & 103.12458801269533 & 0.1810338348150253 \\
& \modelname & 512 & \bfseries 0.1493870615959167 & \bfseries 0.0258333329111337 & \bfseries 0.0016557634808123 & \bfseries 111.10442352294922 & \bfseries 0.213795006275177 & \bfseries 16.81941032409668 & \bfseries 5.426290035247803 & \bfseries 0.0297730416059494 \\
\cmidrule{2-11}
& \tigger & 1024 & 1.0354751348495483 & 0.2053380310535431 & 0.0041474350728094 & 10108.79296875 & 0.690065860748291 & 57.33620834350586 & 352.3189697265625 & 0.0916815623641014 \\
& \dggen & 1024 & 0.6037313938140869 & 0.3058738112449646 & 0.0069277645088732 & 15226.2822265625 & 0.1664832085371017 & 77.05315399169922 & 209.3986663818359 & 0.182693749666214 \\
& \modelname & 1024 & \bfseries 0.1480547040700912 & \bfseries 0.0328323692083358 & \bfseries 0.0023944226559251 & \bfseries 396.5724792480469 & \bfseries 0.0647056177258491 & \bfseries 24.788698196411133 & \bfseries 13.53808307647705 & \bfseries 0.0379015132784843 \\
\cmidrule{2-11}
& \tigger & 2048 & 0.91689532995224 & 0.2039386928081512 & 0.0039715720340609 & 8485.837890625 & 0.5364819169044495 & 88.82051086425781 & 298.4187927246094 & 0.0780421495437622 \\
& \dggen & 2048 & 0.5612298846244812 & 0.3260127305984497 & 0.0037984566297382 & 57731.03125 & 0.1370429247617721 & 108.39735412597656 & 387.2119140625 & 0.181197315454483 \\
& \modelname & 2048 & \bfseries 0.1432246267795562 & \bfseries 0.0240972898900508 & \bfseries 0.0021666146349161 & \bfseries 1646.88232421875 & \bfseries 0.0851200371980667 & \bfseries 39.01470565795898 & \bfseries 29.54901885986328 & \bfseries 0.0298550557345151 \\
\cmidrule{2-11}
& \tigger & 4096 & 0.819267988204956 & 0.2001744359731674 & 0.0031531122513115 & 6103.2451171875 & 0.4014391601085663 & 116.85294342041016 & 120.4313735961914 & 0.0663013979792594 \\
& \dggen & 4096 & 0.479858934879303 & 0.3290866613388061 & \bfseries 0.0013264808803796 & 215833.046875 & 0.2180170565843582 & 136.6447296142578 & 600.4342041015625 & 0.1768423914909362 \\
& \modelname & 4096 & \bfseries 0.1236882880330085 & \bfseries 0.0445952229201793 & 0.0026620314456522 & \bfseries 5079.36279296875 & \bfseries 0.1674173772335052 & \bfseries 57.911766052246094 & \bfseries 65.70587921142578 & \bfseries 0.0327721461653709 \\
\midrule
\midrule
\multirow{12}{*}{\rotatebox{90}{Wikipedia}}
& \tigger & 512 & 0.3085565865039825 & 0.0034311560448259 & 0.0015994750428944 & 65.8515625 & 31.337867736816406 & 8.25 & 14.1015625 & 0.0350485071539878 \\
& \dggen & 512 & 0.0431609973311424 & 0.0065298927947878 & 0.0001986129936994 & 2.0088987350463867 & 45.77573776245117 & 1.780867576599121 & 0.9321468472480774 & 0.0053073116578161 \\
& \modelname & 512 & \bfseries 0.0084427632391452 & \bfseries 0.001744917477481 & \bfseries 9.728568693390116e-05 & \bfseries 0.6718575954437256 & \bfseries 2.934448003768921 & \bfseries 0.5406006574630737 & \bfseries 0.3303670883178711 & \bfseries 0.001852297107689 \\
\cmidrule{2-11}
& \tigger & 1024 & 0.2803383469581604 & 0.0128142787143588 & 0.0015993006527423 & 61.6953125 & 5.155694007873535 & \bfseries 1.7890625 & 12.9609375 & 0.0321338102221488 \\
& \dggen & 1024 & 0.0658824816346168 & 0.0078180134296417 & 0.0001313401589868 & 5.595555782318115 & 35.60015106201172 & 8.571110725402832 & 1.5733333826065063 & 0.0078037641942501 \\
& \modelname & 1024 & \bfseries 0.0118732629343867 & \bfseries 0.004030835814774 & \bfseries 7.594703492941335e-05 & \bfseries 1.6733332872390747 & \bfseries 3.643708229064941 & 4.204444408416748 & \bfseries 0.54666668176651 & \bfseries 0.0028543204534798 \\
\cmidrule{2-11}
& \tigger & 2048 & 0.2361642122268676 & 0.0212652161717414 & 0.0016471394337713 & 49.1484375 & 1.0547757148742676 & 16.8125 & 10.984375 & 0.0280199963599443 \\
& \dggen & 2048 & 0.0965767279267311 & 0.007018809672445 & 0.0001002370045171 & 15.457777976989746 & 7.74625825881958 & 24.68000030517578 & 2.866666555404663 & 0.010945001617074 \\
& \modelname & 2048 & \bfseries 0.0484831295907497 & \bfseries 0.0034100669436156 & \bfseries 5.452732148114592e-05 & \bfseries 8.977777481079102 & \bfseries 0.5167052149772644 & \bfseries 9.088889122009276 & \bfseries 0.9155555367469788 & \bfseries 0.0033116966951638 \\
\cmidrule{2-11}
& \tigger & 4096 & 0.2090259343385696 & 0.0282286349684 & 0.0020712765399366 & 41.54867172241211 & \bfseries 0.6605512499809265 & 39.92920303344727 & 12.300885200500488 & 0.0261699277907609 \\
& \dggen & 4096 & \bfseries 0.128061443567276 & 0.0057908361777663 & 7.761571032460779e-05 & \bfseries 37.66371536254883 & 1.3488792181015017 & 57.0177001953125 & 5.513274192810059 & 0.0137360179796814 \\
& \modelname & 4096 & 0.1628570556640625 & \bfseries 0.0012984115164726 & \bfseries 3.3933356462512165e-05 & 49.371681213378906 & 0.7293851375579834 & \bfseries 5.2123894691467285 & \bfseries 2.238938093185425 & \bfseries 0.0048055057413876 \\
\midrule
\midrule
\multirow{12}{*}{\rotatebox{90}{MOOC}}
& \tigger & 512 & 8.615635871887207 & 0.2638344764709472 & 0.0351805761456489 & 654036.5 & \bfseries 0.475879818201065 & \bfseries 5.0 & 1008.0 & 0.115395650267601 \\
& \dggen & 512 & \bfseries 0.3899450898170471 & \bfseries 0.109187476336956 & 0.0293380990624427 & 169.69192504882812 & 4.650803089141846 & 6.814815044403076 & 22.87878799438477 & 0.0582379773259162 \\
& \modelname & 512 & 0.4422444701194763 & 0.1485691964626312 & \bfseries 0.0292048845440149 & \bfseries 45.695556640625 & 3.292903423309326 & 42.51777648925781 & \bfseries 10.855555534362791 & \bfseries 0.0148989558219909 \\
\cmidrule{2-11}
& \tigger & 1024 & 8.10215950012207 & 0.2467716783285141 & 0.0280229244381189 & 653742.75 & \bfseries 0.2899585664272308 & \bfseries 4.0 & 972.25 & 0.1234108060598373 \\
& \dggen & 1024 & 0.7855606079101562 & \bfseries 0.0799453556537628 & \bfseries 0.0184120703488588 & 864.5882568359375 & 3.005622386932373 & 6.470588207244873 & \bfseries 30.05228805541992 & 0.0808219090104103 \\
& \modelname & 1024 & \bfseries 0.6665012240409851 & 0.1291600465774536 & 0.0236821491271257 & \bfseries 234.5603485107422 & 2.363208770751953 & 38.58620834350586 & 32.36206817626953 & \bfseries 0.0071491911076009 \\
\cmidrule{2-11}
& \tigger & 2048 & 7.039968013763428 & 0.183543711900711 & 0.0210883524268865 & 652723.75 & \bfseries 0.1188230812549591 & \bfseries 1.5 & 920.0 & 0.1232969760894775 \\
& \dggen & 2048 & 1.494163990020752 & \bfseries 0.0366998612880706 & \bfseries 0.0118167838081717 & 3753.8076171875 & 1.3606332540512085 & 3.6282050609588623 & \bfseries 128.82691955566406 & 0.1027706190943718 \\
& \modelname & 2048 & \bfseries 1.1898794174194336 & 0.0808510184288024 & 0.0131147224456071 & \bfseries 1060.5 & 1.149352788925171 & 11.474575996398926 & 277.6525573730469 & \bfseries 0.0089196329936385 \\
\cmidrule{2-11}
& \tigger & 4096 & 5.028779029846191 & 0.1210537552833557 & 0.0115660242736339 & 648518.25 & \bfseries 0.0174873173236846 & \bfseries 0.0 & 861.5 & 0.1197467744350433 \\
& \dggen & 4096 & 2.538737535476685 & 0.0303882211446762 & \bfseries 0.010530212894082 & 15132.8857421875 & 0.800061047077179 & 1.4556962251663208 & \bfseries 298.73419189453125 & 0.1227228865027427 \\
& \modelname & 4096 & \bfseries 1.4834321737289429 & \bfseries 0.0258620716631412 & 0.0119215929880738 & \bfseries 5921.5400390625 & 0.2027632445096969 & 0.5400000214576721 & 529.1599731445312 & \bfseries 0.0205136183649301 \\
\midrule
\midrule
\multirow{12}{*}{\rotatebox{90}{LastFM}}
& \tigger & 512 & 1.3700072765350342 & 0.1810177564620971 & 0.0281812250614166 & 502.5921936035156 & 39.41596984863281 & 1.167597770690918 & 60.357540130615234 & 0.0858300924301147 \\
& \dggen & 512 & \bfseries 0.080675221979618 & \bfseries 0.0724979043006897 & \bfseries 0.0077964588999748 & 4.410299301147461 & \bfseries 38.07106018066406 & \bfseries 0.7175697684288025 & 1.0279146432876587 & 0.0150425741448998 \\
& \modelname & 512 & 0.0811406597495079 & 0.1880508065223693 & 0.0098752491176128 & \bfseries 3.3679652214050293 & 43.92637252807617 & 1.933097243309021 & \bfseries 0.92606920003891 & \bfseries 0.0137876262888312 \\
\cmidrule{2-11}
& \tigger & 1024 & 1.2969797849655151 & 0.1874247044324875 & 0.0293711517006158 & 505.0806579589844 & \bfseries 15.695145606994627 & 4.569892406463623 & 59.60752868652344 & 0.0743017420172691 \\
& \dggen & 1024 & \bfseries 0.130961462855339 & \bfseries 0.1090119704604148 & 0.008957453072071 & 7.387740612030029 & 39.65644836425781 & \bfseries 1.6466906070709229 & \bfseries 1.5934789180755615 & 0.0211046524345874 \\
& \modelname & 1024 & \bfseries 0.1310970932245254 & 0.1552326381206512 & \bfseries 0.0086775003001093 & \bfseries 6.695917129516602 & 54.06063461303711 & 3.5854649543762207 & 1.6638150215148926 & \bfseries 0.0201642345637083 \\
\cmidrule{2-11}
& \tigger & 2048 & 1.2306745052337646 & 0.1869687288999557 & 0.0246156323701143 & 501.9010314941406 & \bfseries 8.30788803100586 & 9.505208015441896 & 58.16666793823242 & 0.0657169967889785 \\
& \dggen & 2048 & \bfseries 0.1840354800224304 & 0.0985077396035194 & 0.0081682465970516 & \bfseries 17.212888717651367 & 12.02513027191162 & 4.850551605224609 & \bfseries 2.3748745918273926 & \bfseries 0.028411716222763 \\
& \modelname & 2048 & 0.2041643112897873 & \bfseries 0.0725927352905273 & \bfseries 0.0080287028104066 & 17.58827781677246 & 46.76123809814453 & \bfseries 4.144310474395752 & 3.020063161849976 & 0.0296491682529449 \\
\cmidrule{2-11}
& \tigger & 4096 & 1.1395180225372314 & 0.1869139820337295 & 0.0267389882355928 & 461.1958618164063 & 4.707253932952881 & 16.36598014831543 & 54.55154800415039 & 0.0578175112605094 \\
& \dggen & 4096 & 0.2304709404706955 & 0.0680913031101226 & 0.0056739798747003 & \bfseries 45.188961029052734 & \bfseries 4.030118942260742 & 11.867947578430176 & \bfseries 3.7080562114715576 & 0.0351373739540576 \\
& \modelname & 4096 & \bfseries 0.2036898583173752 & \bfseries 0.0428421534597873 & \bfseries 0.0031013460829854 & 49.10028076171875 & 20.943723678588867 & \bfseries 9.433359146118164 & 4.707540035247803 & \bfseries 0.0321219451725482 \\
\bottomrule
\end{tabular}}

   \label{tab:topo_dist_wasserstein}
\end{table}

\begin{table}[t]
   \centering
   \caption{
      \textbf{KS statistic between the generated and real per-window statistics, on the JODIE datasets.}
      Same generated graphs and window sizes as \cref{tab:topo_all_ws}. Best value in bold.
   }
   
\rebuttaltwocolor
{\setlength{\tabcolsep}{4pt}
\begin{tabular}{c l l TTTTTTTT}
\toprule
{} & {\colmodel} & {\colwindow} & {\nodedegree $\downarrow$} & {\clocen $\downarrow$} & {\btwc $\downarrow$} & {WC $\downarrow$} & {PLE $\downarrow$} & {NCC $\downarrow$} & {LCC $\downarrow$} & {REL $\downarrow$} \\
\midrule
\multirow{12}{*}{\rotatebox{90}{Reddit}}
& \tigger & 512 & 1.0 & 1.0 & 1.0 & 1.0 & 1.0 & 0.9568965435028076 & 1.0 & 1.0 \\
& \dggen & 512 & 1.0 & 1.0 & 1.0 & 0.9485049843788148 & 0.5132890343666077 & 1.0 & 1.0 & 0.9717608094215392 \\
& \modelname & 512 & \bfseries 0.5196560025215149 & \bfseries 0.2346437275409698 & \bfseries 0.3685503602027893 & \bfseries 0.601965606212616 & \bfseries 0.3452088534832001 & \bfseries 0.3169533312320709 & \bfseries 0.4914004802703857 & \bfseries 0.5540540814399719 \\
\cmidrule{2-11}
& \tigger & 1024 & 1.0 & 1.0 & 1.0 & 1.0 & 1.0 & 1.0 & 1.0 & 1.0 \\
& \dggen & 1024 & 1.0 & 1.0 & 1.0 & 0.9401993155479432 & 0.3488371968269348 & 1.0 & 1.0 & 0.980066418647766 \\
& \modelname & 1024 & \bfseries 0.4545454680919647 & \bfseries 0.4643734693527221 & \bfseries 0.3636363744735718 & \bfseries 0.5724815726280212 & \bfseries 0.1498771458864212 & \bfseries 0.3095823228359222 & \bfseries 0.4422604441642761 & \bfseries 0.5896806120872498 \\
\cmidrule{2-11}
& \tigger & 2048 & 1.0 & 1.0 & 1.0 & 0.9059829115867616 & 1.0 & 1.0 & 0.9829059839248656 & 1.0 \\
& \dggen & 2048 & 1.0 & 1.0 & 1.0 & 0.940397322177887 & \bfseries 0.2847682237625122 & 1.0 & 0.9933775067329408 & 0.9536423683166504 \\
& \modelname & 2048 & \bfseries 0.4901960790157318 & \bfseries 0.4950980246067047 & \bfseries 0.343137264251709 & \bfseries 0.7401960492134094 & 0.3823529481887817 & \bfseries 0.4607843160629273 & \bfseries 0.4166666567325592 & \bfseries 0.5980392098426819 \\
\cmidrule{2-11}
& \tigger & 4096 & 1.0 & 1.0 & 0.9803921580314636 & \bfseries 0.343137264251709 & 1.0 & 1.0 & 0.3529411852359772 & 1.0 \\
& \dggen & 4096 & 1.0 & 1.0 & 0.8815789222717285 & 0.9473684430122375 & \bfseries 0.5789473652839661 & 1.0 & 0.9868420958518982 & 0.9473684430122375 \\
& \modelname & 4096 & \bfseries 0.3333333432674408 & \bfseries 0.3921568691730499 & \bfseries 0.4607843160629273 & 0.7156862616539001 & 0.8235294222831726 & \bfseries 0.5490196347236633 & \bfseries 0.294117659330368 & \bfseries 0.5686274766921997 \\
\midrule
\midrule
\multirow{12}{*}{\rotatebox{90}{Wikipedia}}
& \tigger & 512 & 0.9609375 & 0.1171875 & 0.8125 & 0.9921875 & 0.953125 & 0.609375 & 0.984375 & 0.9296875 \\
& \dggen & 512 & 0.517241358757019 & 0.2658509314060211 & 0.5127919912338257 & 0.5050055384635925 & 0.517241358757019 & 0.1957730799913406 & 0.5050055384635925 & 0.5517241358757019 \\
& \modelname & 512 & \bfseries 0.0767519474029541 & \bfseries 0.095661848783493 & \bfseries 0.0912124589085578 & \bfseries 0.0923248082399368 & \bfseries 0.0511679649353027 & \bfseries 0.07119020819664 & \bfseries 0.1379310339689254 & \bfseries 0.1190211325883865 \\
\cmidrule{2-11}
& \tigger & 1024 & 0.953125 & 0.6015625 & 0.875 & 0.921875 & 0.921875 & \bfseries 0.140625 & 0.8984375 & 0.890625 \\
& \dggen & 1024 & 0.7133333086967468 & 0.6311110854148865 & 0.7977777719497681 & 0.6044444441795349 & 0.7111111283302307 & 0.54666668176651 & 0.5066666603088379 & 0.7288888692855835 \\
& \modelname & 1024 & \bfseries 0.1577777713537216 & \bfseries 0.3066666722297668 & \bfseries 0.2000000029802322 & \bfseries 0.1733333319425583 & \bfseries 0.1822222173213958 & 0.273333340883255 & \bfseries 0.1466666609048843 & \bfseries 0.1555555611848831 \\
\cmidrule{2-11}
& \tigger & 2048 & 0.9375 & 0.90625 & 0.921875 & 0.703125 & 0.8984375 & 0.78125 & 0.7734375 & 0.859375 \\
& \dggen & 2048 & 0.8933333158493042 & 0.9333333373069764 & 0.959999978542328 & 0.6977777481079102 & 0.8844444155693054 & 0.8711110949516296 & 0.644444465637207 & 0.8799999952316284 \\
& \modelname & 2048 & \bfseries 0.3866666555404663 & \bfseries 0.4399999976158142 & \bfseries 0.1733333319425583 & \bfseries 0.3155555427074432 & \bfseries 0.4755555689334869 & \bfseries 0.4088888764381408 & \bfseries 0.1155555546283722 & \bfseries 0.2044444382190704 \\
\cmidrule{2-11}
& \tigger & 4096 & 0.8495575189590454 & 0.9734513163566588 & 0.9203540086746216 & \bfseries 0.2654867172241211 & \bfseries 0.8141592741012573 & 0.982300877571106 & 0.5575221180915833 & 0.778761088848114 \\
& \dggen & 4096 & 0.9734513163566588 & 1.0 & 1.0 & 0.769911527633667 & 0.9734513163566588 & 0.982300877571106 & 0.7876105904579163 & 0.9734513163566588 \\
& \modelname & 4096 & \bfseries 0.778761088848114 & \bfseries 0.2654867172241211 & \bfseries 0.2212389409542083 & 0.5663716793060303 & 0.8849557638168335 & \bfseries 0.2477876096963882 & \bfseries 0.2212389409542083 & \bfseries 0.3716814219951629 \\
\midrule
\midrule
\multirow{12}{*}{\rotatebox{90}{MOOC}}
& \tigger & 512 & 1.0 & 1.0 & 1.0 & 1.0 & 1.0 & 1.0 & 1.0 & 1.0 \\
& \dggen & 512 & \bfseries 0.4949494898319245 & \bfseries 0.5521885752677917 & \bfseries 0.6700336933135986 & 0.4478114545345306 & 0.7121211886405945 & \bfseries 0.6632996797561646 & 0.2727272808551788 & 0.804713785648346 \\
& \modelname & 512 & 0.6600000262260437 & 0.8288888931274414 & 0.8266666531562805 & \bfseries 0.246666669845581 & \bfseries 0.6511111259460449 & 0.8777777552604675 & \bfseries 0.1511111110448837 & \bfseries 0.3777777850627899 \\
\cmidrule{2-11}
& \tigger & 1024 & 1.0 & 1.0 & 1.0 & 1.0 & 1.0 & 1.0 & 1.0 & 1.0 \\
& \dggen & 1024 & 0.656862735748291 & \bfseries 0.5784313678741455 & \bfseries 0.6633986830711365 & 0.5816993713378906 & 0.7483659982681274 & \bfseries 0.6830065250396729 & 0.330065369606018 & 0.9183006286621094 \\
& \modelname & 1024 & \bfseries 0.5991379022598267 & 0.7974137663841248 & 0.818965494632721 & \bfseries 0.3189655244350433 & \bfseries 0.6206896305084229 & 0.8793103694915771 & \bfseries 0.2198275923728943 & \bfseries 0.1724137961864471 \\
\cmidrule{2-11}
& \tigger & 2048 & 1.0 & 1.0 & 1.0 & 1.0 & 1.0 & 0.75 & 1.0 & 1.0 \\
& \dggen & 2048 & 0.6730769276618958 & \bfseries 0.4358974397182464 & \bfseries 0.7307692170143127 & 0.6858974099159241 & 0.807692289352417 & \bfseries 0.5128205418586731 & \bfseries 0.7692307829856873 & 0.9679487347602844 \\
& \modelname & 2048 & \bfseries 0.6101694703102112 & 0.5593220591545105 & 0.7711864113807678 & \bfseries 0.4491525292396545 & \bfseries 0.6186440587043762 & 0.7457627058029175 & 0.8389830589294434 & \bfseries 0.3644067943096161 \\
\cmidrule{2-11}
& \tigger & 4096 & 1.0 & 1.0 & 1.0 & 1.0 & 0.5 & \bfseries 0.0 & 1.0 & 1.0 \\
& \dggen & 4096 & 0.7215189933776855 & 0.3670886158943176 & 0.8734177350997925 & 0.7721518874168396 & 0.8354430198669434 & 0.3417721390724182 & 0.9746835231781006 & 0.9746835231781006 \\
& \modelname & 4096 & \bfseries 0.4199999868869781 & \bfseries 0.2800000011920929 & \bfseries 0.7599999904632568 & \bfseries 0.5 & \bfseries 0.4600000083446502 & 0.3000000119209289 & \bfseries 0.8600000143051147 & \bfseries 0.4000000059604645 \\
\midrule
\midrule
\multirow{12}{*}{\rotatebox{90}{LastFM}}
& \tigger & 512 & 0.9888268113136292 & 0.6536312699317932 & 0.9385474920272828 & 0.9776536226272584 & 1.0 & 0.2122904956340789 & 1.0 & 0.9385474920272828 \\
& \dggen & 512 & \bfseries 0.4167278110980987 & \bfseries 0.1392935812473297 & \bfseries 0.4167278110980987 & \bfseries 0.4167278110980987 & \bfseries 0.4167278110980987 & \bfseries 0.1030639708042144 & \bfseries 0.4167278110980987 & 0.4834573566913605 \\
& \modelname & 512 & 0.4811223149299621 & 0.3284118771553039 & 0.4811223149299621 & 0.4811223149299621 & 0.4811223149299621 & 0.3399073779582977 & 0.4811223149299621 & \bfseries 0.4811767935752868 \\
\cmidrule{2-11}
& \tigger & 1024 & 0.9946236610412598 & 0.7473118305206299 & 0.9193548560142516 & 0.9677419066429138 & 1.0 & 0.5806451439857483 & 0.9946236610412598 & 0.8870967626571655 \\
& \dggen & 1024 & \bfseries 0.5831757187843323 & \bfseries 0.2951418459415436 & \bfseries 0.4952722489833832 & \bfseries 0.5201174020767212 & \bfseries 0.5791326761245728 & \bfseries 0.2225627601146698 & \bfseries 0.4301271736621856 & \bfseries 0.5788066387176514 \\
& \modelname & 1024 & 0.5902083516120911 & 0.3818397521972656 & 0.5902083516120911 & 0.5902083516120911 & 0.5902083516120911 & 0.3997119963169098 & 0.5902083516120911 & 0.5902083516120911 \\
\cmidrule{2-11}
& \tigger & 2048 & 0.9947916865348816 & 0.78125 & 0.921875 & 0.9479166865348816 & 1.0 & 0.71875 & 0.9947916865348816 & 0.8958333134651184 \\
& \dggen & 2048 & \bfseries 0.7383400201797485 & 0.5072718262672424 & 0.6124624013900757 & \bfseries 0.6031845808029175 & \bfseries 0.7186559438705444 & 0.3872868716716766 & \bfseries 0.4768054187297821 & 0.7538866400718689 \\
& \modelname & 2048 & 0.7586700916290283 & \bfseries 0.2325881272554397 & \bfseries 0.6008884906768799 & 0.6887360215187073 & 0.7469189167022705 & \bfseries 0.2956434488296509 & 0.6898825168609619 & \bfseries 0.7484952807426453 \\
\cmidrule{2-11}
& \tigger & 4096 & 0.9896907210350036 & 0.8298969268798828 & 0.938144326210022 & 0.8814433217048645 & 0.9896907210350036 & 0.7628865838050842 & 0.9639175534248352 & 0.9020618796348572 \\
& \dggen & 4096 & 0.8122690320014954 & 0.7339246273040771 & 0.6898250579833984 & \bfseries 0.6457255482673645 & 0.774821400642395 & 0.5723084211349487 & \bfseries 0.468095600605011 & 0.864745020866394 \\
& \modelname & 4096 & \bfseries 0.6278243064880371 & \bfseries 0.4259964525699615 & \bfseries 0.2267072796821594 & 0.669205367565155 & \bfseries 0.5425235033035278 & \bfseries 0.5476009249687195 & 0.5704493522644043 & \bfseries 0.7377507090568542 \\
\bottomrule
\end{tabular}}

   \label{tab:topo_dist_ks}
\end{table}

\begin{table}[t]
   \centering
   \caption{
      \textbf{Wasserstein distance between the generated and real per-window statistics, on the DyGLib datasets.}
      Same generated graphs and window sizes as \cref{tab:topo_all_ws_dgb}. Best value in bold.
   }
   
\rebuttaltwocolor
{\setlength{\tabcolsep}{4pt}
\begin{tabular}{c l l TTTTTTTT}
\toprule
{} & {\colmodel} & {\colwindow} & {\nodedegree $\downarrow$} & {\clocen $\downarrow$} & {\btwc $\downarrow$} & {WC $\downarrow$} & {PLE $\downarrow$} & {NCC $\downarrow$} & {LCC $\downarrow$} & {REL $\downarrow$} \\
\midrule
\multirow{12}{*}{\rotatebox{90}{Enron}}
& \tigger & 6\,h & 0.31905069947242737 & 0.2209404557943344 & 0.09200167655944824 & \bfseries 14.600000381469728 & 18.70701789855957 & 3.799999952316284 & 3.0 & 0.025748109444975853 \\
& \dggen & 6\,h & \bfseries 0.13508717715740204 & 0.2576931118965149 & 0.02742023579776287 & 26.770666122436523 & 13.251096725463867 & 3.8533332347869873 & 2.7706665992736816 & 0.03239860758185387 \\
& \modelname & 6\,h & 0.2146653383970261 & \bfseries 0.0895309075713157 & \bfseries 0.0153061691671609 & 22.4946231842041 & \bfseries 3.335827112197876 & \bfseries 0.8172042965888977 & \bfseries 2.6236557960510254 & \bfseries 0.0247194841504097 \\
\cmidrule{2-11}
& \tigger & 12\,h & 0.3147028684616089 & 0.2209404557943344 & 0.09200167655944824 & \bfseries 15.0 & 18.704116821289062 & 3.799999952316284 & \bfseries 3.0 & 0.02577143907546997 \\
& \dggen & 12\,h & \bfseries 0.16879567503929138 & 0.2009061574935913 & \bfseries 0.023174257948994637 & 39.380393981933594 & 11.080733299255373 & 3.698039293289185 & 11.84705924987793 & 0.028567206114530563 \\
& \modelname & 12\,h & 0.2784185707569122 & \bfseries 0.0461140573024749 & 0.0288607608526945 & 33.852630615234375 & \bfseries 1.4200679063796997 & \bfseries 0.6631578803062439 & 3.768420934677124 & \bfseries 0.0240743868052959 \\
\cmidrule{2-11}
& \tigger & 1\,d & \bfseries 0.12159581482410432 & 0.2944469153881073 & 0.11643316596746443 & 70.0 & 18.498470306396484 & 8.399999618530273 & \bfseries 5.599999904632568 & 0.029218722134828568 \\
& \dggen & 1\,d & 0.39180177450180054 & 0.1379530280828476 & 0.028413156047463417 & 89.69677734375 & 6.905555248260498 & 1.1548386812210083 & 34.051612854003906 & 0.02610679157078266 \\
& \modelname & 1\,d & 0.3811531662940979 & \bfseries 0.0688232704997062 & \bfseries 0.0167903546243906 & \bfseries 59.87058639526367 & \bfseries 1.5964771509170532 & \bfseries 0.9176470637321472 & 6.929411888122559 & \bfseries 0.0222667455673217 \\
\cmidrule{2-11}
& \tigger & 2\,d & \bfseries 0.27718567848205566 & 0.3172500729560852 & 0.11130902171134947 & 129.14285278320312 & 13.403624534606934 & 5.142857074737549 & 18.0 & 0.020828494802117348 \\
& \dggen & 2\,d & 0.9226210713386536 & 0.1599043607711792 & 0.02863780222833157 & 342.9204406738281 & \bfseries 4.62317419052124 & 2.0 & 62.477272033691406 & 0.02492920495569706 \\
& \modelname & 2\,d & 0.5319749712944031 & \bfseries 0.1246771141886711 & \bfseries 0.0188910216093063 & \bfseries 99.09677124023438 & 4.946832656860352 & \bfseries 1.4838709831237793 & \bfseries 10.612903594970703 & \bfseries 0.0143690668046474 \\
\midrule
\midrule
\multirow{12}{*}{\rotatebox{90}{UCI}}
& \tigger & 3\,h & 5.391523838043213 & 0.25903239846229553 & \bfseries 0.012651821598410606 & 30400.0 & 33.0074348449707 & \bfseries 5.666666507720947 & 240.0 & 0.11383279412984848 \\
& \dggen & 3\,h & 0.1401572972536087 & 0.18891213834285736 & 0.016725772991776466 & \bfseries 1.75 & 26.193706512451172 & 16.52777862548828 & \bfseries 1.0277777910232544 & \bfseries 0.017973365262150764 \\
& \modelname & 3\,h & \bfseries 0.1148916855454444 & \bfseries 0.1859261393547058 & 0.0137990023940801 & 12.472527503967283 & \bfseries 9.865991592407228 & 7.692307472229004 & 1.4249083995819092 & 0.0188857782632112 \\
\cmidrule{2-11}
& \tigger & 6\,h & 5.3992533683776855 & 0.27389997243881226 & 0.012833282351493835 & 30400.0 & 33.179840087890625 & \bfseries 7.333333492279053 & 240.0 & 0.11493966728448868 \\
& \dggen & 6\,h & 0.1883877068758011 & \bfseries 0.0889492779970169 & \bfseries 0.006764823105186224 & \bfseries 5.357142925262451 & 33.67959976196289 & 20.60714340209961 & 2.392857074737549 & 0.020480308681726456 \\
& \modelname & 6\,h & \bfseries 0.1052994206547737 & 0.098544679582119 & 0.0088629499077796 & 24.82638931274414 & \bfseries 8.751677513122559 & 13.770833015441896 & \bfseries 2.0625 & \bfseries 0.0178253185003995 \\
\cmidrule{2-11}
& \tigger & 12\,h & 5.257042407989502 & 0.2961259186267853 & 0.012374824844300749 & 30391.666015625 & \bfseries 1.7480051517486572 & \bfseries 14.0 & 237.3333282470703 & 0.10191156715154648 \\
& \dggen & 12\,h & 0.2340860366821289 & \bfseries 0.04634133353829384 & \bfseries 0.004001239780336618 & \bfseries 8.380952835083008 & 44.6854248046875 & 27.4761905670166 & 4.4285712242126465 & 0.023115213960409164 \\
& \modelname & 12\,h & \bfseries 0.0783504545688629 & 0.0513443499803543 & 0.0045356662012636 & 56.75172424316406 & 2.0289595127105717 & 21.29655265808105 & \bfseries 2.951724052429199 & \bfseries 0.0150564387440681 \\
\cmidrule{2-11}
& \tigger & 1\,d & 4.940710067749023 & 0.2832775413990021 & 0.010294093750417233 & 30205.333984375 & 1.1562495231628418 & \bfseries 20.33333396911621 & 212.6666717529297 & 0.07075474411249161 \\
& \dggen & 1\,d & 0.40190601348876953 & \bfseries 0.027169324457645416 & 0.0044360291212797165 & \bfseries 58.400001525878906 & 32.24980545043945 & 36.20000076293945 & \bfseries 14.266666412353516 & 0.03488100692629814 \\
& \modelname & 1\,d & \bfseries 0.0553069561719894 & 0.0327473804354667 & \bfseries 0.0032444125972688 & 145.9589080810547 & \bfseries 0.0619167722761631 & 26.50684928894043 & 21.917808532714844 & \bfseries 0.0113401673734188 \\
\midrule
\midrule
\multirow{12}{*}{\rotatebox{90}{SocialEvo}}
& \tigger & 1\,h & 6.934163570404053 & 0.2760625183582306 & 0.02108740247786045 & 6041.53857421875 & \bfseries 0.2320655733346939 & 1.7692307233810425 & 26.769229888916016 & 0.04164693504571915 \\
& \dggen & 1\,h & 5.865054607391357 & 0.1572120189666748 & 0.023045916110277176 & 7017.5712890625 & 9.017427444458008 & \bfseries 0.9523809552192688 & \bfseries 21.66666603088379 & 0.018751611933112144 \\
& \modelname & 1\,h & \bfseries 0.9344729781150818 & \bfseries 0.061512753367424 & \bfseries 0.0091481758281588 & \bfseries 79.64077758789062 & 0.7949137687683105 & 1.6084142923355105 & 27.179611206054688 & \bfseries 0.0182260312139987 \\
\cmidrule{2-11}
& \tigger & 2\,h & 6.438403129577637 & 0.24851347506046295 & 0.01956559345126152 & 5897.0771484375 & 0.20252768695354464 & 1.692307710647583 & \bfseries 25.461538314819336 & 0.03835076838731766 \\
& \dggen & 2\,h & 9.836217880249023 & 0.15668541193008423 & 0.016987040638923645 & 15161.9091796875 & 0.203845277428627 & \bfseries 1.0 & 30.18181800842285 & \bfseries 0.013312323950231075 \\
& \modelname & 2\,h & \bfseries 0.6624510884284973 & \bfseries 0.0222914330661296 & \bfseries 0.0131613314151763 & \bfseries 862.3009643554688 & \bfseries 0.0311470758169889 & 1.1197410821914673 & 38.08090591430664 & 0.0212289281189441 \\
\cmidrule{2-11}
& \tigger & 4\,h & 5.301340579986572 & 0.18812039494514465 & 0.023482387885451317 & 5507.46142578125 & 0.13322900235652924 & 1.384615421295166 & \bfseries 22.846153259277344 & 0.03905906155705452 \\
& \dggen & 4\,h & 16.511953353881836 & 0.14668679237365723 & \bfseries 0.015618501231074331 & 31496.666015625 & 0.4125277101993561 & \bfseries 0.3333333432674408 & 30.0 & \bfseries 0.0172288715839386 \\
& \modelname & 4\,h & \bfseries 3.890486717224121 & \bfseries 0.0686692297458648 & 0.0196748860180377 & \bfseries 3901.342041015625 & \bfseries 0.0868294090032577 & 1.0387096405029297 & 38.98709487915039 & 0.0276136063039302 \\
\cmidrule{2-11}
& \tigger & 8\,h & \bfseries 3.2491533756256104 & 0.12412618100643158 & 0.015260647982358932 & \bfseries 4597.23095703125 & \bfseries 0.04983934015035629 & 1.076923131942749 & \bfseries 18.30769157409668 & 0.042090218514204025 \\
& \dggen & 8\,h & 23.620100021362305 & 0.16998079419136047 & \bfseries 0.01078004203736782 & 51351.25 & 0.8929119110107422 & \bfseries 0.25 & 27.0 & \bfseries 0.023599758744239807 \\
& \modelname & 8\,h & 9.785775184631348 & \bfseries 0.0982617363333702 & 0.018586553633213 & 13072.8720703125 & 0.3685839772224426 & 0.7820512652397156 & 36.25640869140625 & 0.0311739891767501 \\
\midrule
\midrule
\multirow{12}{*}{\rotatebox{90}{Contacts}}
& \tigger & 10\,min & \bfseries 0.6694278120994568 & 0.14064909517765045 & 0.01770157925784588 & \bfseries 942.2342529296876 & \bfseries 0.08280034363269806 & 34.01615524291992 & 71.19709014892578 & \bfseries 0.01184767670929432 \\
& \dggen & 10\,min & 1.9045941829681396 & 0.35747072100639343 & 0.012008300051093102 & 3120.351806640625 & 46.772560119628906 & 43.185184478759766 & \bfseries 19.11111068725586 & 0.048745207488536835 \\
& \modelname & 10\,min & 2.118053674697876 & \bfseries 0.0185376778244972 & \bfseries 0.0002505111624486 & 3464.720458984375 & 5.661632537841797 & \bfseries 21.98757743835449 & 20.542442321777344 & 0.0373804606497287 \\
\cmidrule{2-11}
& \tigger & 20\,min & \bfseries 0.5533249378204346 & 0.20915557444095612 & 0.016532370820641518 & 6288.416015625 & \bfseries 0.15279683470726013 & 36.25806427001953 & 103.348388671875 & \bfseries 0.02315780334174633 \\
& \dggen & 20\,min & 2.358163356781006 & 0.3525475859642029 & 0.006386695429682732 & \bfseries 4682.02490234375 & 44.467567443847656 & 41.95000076293945 & \bfseries 27.625 & 0.04644105210900307 \\
& \modelname & 20\,min & 2.4578628540039062 & \bfseries 0.0088201342150568 & \bfseries 0.0012051934609189 & 5192.81201171875 & 3.0183968544006348 & \bfseries 26.16312026977539 & 79.86524963378906 & 0.0359894931316375 \\
\cmidrule{2-11}
& \tigger & 40\,min & \bfseries 1.664589524269104 & 0.25642362236976624 & 0.01028965786099434 & 19589.4765625 & \bfseries 0.21787361800670624 & 36.458065032958984 & 124.00645446777344 & \bfseries 0.0344645194709301 \\
& \dggen & 40\,min & 3.004631996154785 & 0.3467561602592468 & 0.005192379001528025 & \bfseries 7429.7744140625 & 45.338958740234375 & 37.64516067504883 & \bfseries 48.93548202514648 & 0.04880845174193382 \\
& \modelname & 40\,min & 2.655019760131836 & \bfseries 0.0261708684265613 & \bfseries 0.0035773464478552 & 7945.97314453125 & 0.9329614639282228 & \bfseries 16.553333282470703 & 247.8800048828125 & 0.0346349030733108 \\
\cmidrule{2-11}
& \tigger & 80\,min & 3.3098599910736084 & 0.2813764810562134 & 0.005590076558291912 & 47387.71875 & 0.2604716718196869 & 34.487178802490234 & 132.65383911132812 & 0.04236592724919319 \\
& \dggen & 80\,min & 3.659244298934937 & 0.3083900511264801 & 0.004856551997363567 & \bfseries 11443.9599609375 & 48.73542022705078 & 32.119998931884766 & \bfseries 88.76000213623047 & 0.04682326689362526 \\
& \modelname & 80\,min & \bfseries 2.845965385437012 & \bfseries 0.0863969922065734 & \bfseries 0.0031719135586172 & 13910.974609375 & \bfseries 0.1682176291942596 & \bfseries 23.0256404876709 & 433.2564086914063 & \bfseries 0.0381097495555877 \\
\bottomrule
\end{tabular}}

   \label{tab:topo_dist_wasserstein_dgb}
\end{table}

\begin{table}[t]
   \centering
   \caption{
      \textbf{KS statistic between the generated and real per-window statistics, on the DyGLib datasets.}
      Same generated graphs and window sizes as \cref{tab:topo_all_ws_dgb}. Best value in bold.
   }
   
\rebuttaltwocolor
{\setlength{\tabcolsep}{4pt}
\begin{tabular}{c l l TTTTTTTT}
\toprule
{} & {\colmodel} & {\colwindow} & {\nodedegree $\downarrow$} & {\clocen $\downarrow$} & {\btwc $\downarrow$} & {WC $\downarrow$} & {PLE $\downarrow$} & {NCC $\downarrow$} & {LCC $\downarrow$} & {REL $\downarrow$} \\
\midrule
\multirow{12}{*}{\rotatebox{90}{Enron}}
& \tigger & 6\,h & \bfseries 0.20000000298023224 & 0.6000000238418579 & 0.6000000238418579 & \bfseries 0.20000000298023224 & 0.6000000238418579 & 0.6000000238418579 & 0.4000000059604645 & 0.4000000059604645 \\
& \dggen & 6\,h & 0.2080000042915344 & 0.5306666493415833 & \bfseries 0.1653333306312561 & 0.2826666533946991 & \bfseries 0.14399999380111694 & 0.47999998927116394 & 0.2906666696071625 & 0.4586666524410248 \\
& \modelname & 6\,h & 0.4086021482944488 & \bfseries 0.1720430105924606 & 0.1827957034111023 & 0.2580645084381103 & 0.3548386991024017 & \bfseries 0.107526883482933 & \bfseries 0.2150537669658661 & \bfseries 0.2688172161579132 \\
\cmidrule{2-11}
& \tigger & 12\,h & \bfseries 0.20000000298023224 & 0.6000000238418579 & 0.6000000238418579 & \bfseries 0.20000000298023224 & 0.6000000238418579 & 0.6000000238418579 & 0.4000000059604645 & 0.4000000059604645 \\
& \dggen & 12\,h & 0.2549019753932953 & 0.4470588266849518 & 0.23137255012989044 & 0.3960784375667572 & \bfseries 0.3333333432674408 & 0.4588235318660736 & 0.5098039507865906 & 0.3843137323856354 \\
& \modelname & 12\,h & 0.4631578922271729 & \bfseries 0.1368421018123626 & \bfseries 0.168421059846878 & 0.3473684191703796 & 0.410526305437088 & \bfseries 0.0631578937172889 & \bfseries 0.2526315748691559 & \bfseries 0.2736842036247253 \\
\cmidrule{2-11}
& \tigger & 1\,d & \bfseries 0.4000000059604645 & 1.0 & 0.800000011920929 & 0.6000000238418579 & \bfseries 0.20000000298023224 & 1.0 & 0.4000000059604645 & 0.4000000059604645 \\
& \dggen & 1\,d & 0.4516128897666931 & 0.3096774220466614 & 0.5612903237342834 & 0.47096773982048035 & 0.5354838967323303 & 0.25806450843811035 & 0.625806450843811 & 0.4322580695152282 \\
& \modelname & 1\,d & 0.5058823823928833 & \bfseries 0.1176470592617988 & \bfseries 0.1529411822557449 & \bfseries 0.3411764800548553 & 0.5058823823928833 & \bfseries 0.0941176488995552 & \bfseries 0.3058823645114898 & \bfseries 0.294117659330368 \\
\cmidrule{2-11}
& \tigger & 2\,d & \bfseries 0.4285714328289032 & 0.7142857313156128 & 0.8571428656578064 & 0.4285714328289032 & \bfseries 0.2857142984867096 & 0.7142857313156128 & 0.4285714328289032 & 0.4285714328289032 \\
& \dggen & 2\,d & 0.6931818127632141 & 0.4545454680919647 & 0.5795454382896423 & 0.6022727489471436 & 0.75 & 0.3181818127632141 & 0.7045454382896423 & 0.4545454680919647 \\
& \modelname & 2\,d & 0.6451612710952759 & \bfseries 0.2419354766607284 & \bfseries 0.2258064448833465 & \bfseries 0.338709682226181 & 0.6612903475761414 & \bfseries 0.1774193495512008 & \bfseries 0.2580645084381103 & \bfseries 0.2258064448833465 \\
\midrule
\midrule
\multirow{12}{*}{\rotatebox{90}{UCI}}
& \tigger & 3\,h & 1.0 & 1.0 & 1.0 & 1.0 & 1.0 & 1.0 & 1.0 & 1.0 \\
& \dggen & 3\,h & 0.6111111044883728 & \bfseries 0.5555555820465088 & 0.6666666865348816 & 0.2777777910232544 & 0.6111111044883728 & \bfseries 0.5277777910232544 & 0.2777777910232544 & 0.6666666865348816 \\
& \modelname & 3\,h & \bfseries 0.3919413983821869 & 0.7582417726516724 & \bfseries 0.5622710585594177 & \bfseries 0.1410256475210189 & \bfseries 0.380952388048172 & 0.7069597244262695 & \bfseries 0.1556776612997055 & \bfseries 0.4578754603862762 \\
\cmidrule{2-11}
& \tigger & 6\,h & 1.0 & 1.0 & 1.0 & 1.0 & 1.0 & 1.0 & 1.0 & 1.0 \\
& \dggen & 6\,h & 0.75 & \bfseries 0.6071428656578064 & 0.8571428656578064 & 0.3928571343421936 & 0.75 & \bfseries 0.5714285969734192 & 0.4642857015132904 & 0.8214285969734192 \\
& \modelname & 6\,h & \bfseries 0.3645833432674408 & 0.875 & \bfseries 0.6770833134651184 & \bfseries 0.2847222089767456 & \bfseries 0.3333333432674408 & 0.8611111044883728 & \bfseries 0.1493055522441864 & \bfseries 0.4201388955116272 \\
\cmidrule{2-11}
& \tigger & 12\,h & 1.0 & 1.0 & 1.0 & 1.0 & 1.0 & 1.0 & 1.0 & 1.0 \\
& \dggen & 12\,h & 0.9523809552192688 & \bfseries 0.4761904776096344 & 1.0 & 0.523809552192688 & 0.9523809552192688 & \bfseries 0.4285714328289032 & 0.6190476417541504 & 1.0 \\
& \modelname & 12\,h & \bfseries 0.2689655125141144 & 0.8896551728248596 & \bfseries 0.6482758522033691 & \bfseries 0.475862056016922 & \bfseries 0.2137930989265442 & 0.917241394519806 & \bfseries 0.2620689570903778 & \bfseries 0.2965517342090606 \\
\cmidrule{2-11}
& \tigger & 1\,d & 1.0 & 1.0 & 0.6666666865348816 & 1.0 & 1.0 & 1.0 & 1.0 & 0.6666666865348816 \\
& \dggen & 1\,d & 1.0 & \bfseries 0.6000000238418579 & 1.0 & 0.800000011920929 & 1.0 & \bfseries 0.6000000238418579 & 0.800000011920929 & 1.0 \\
& \modelname & 1\,d & \bfseries 0.1780821979045868 & 0.6027397513389587 & \bfseries 0.3013698756694793 & \bfseries 0.6438356041908264 & \bfseries 0.1232876703143119 & 0.931506872177124 & \bfseries 0.465753436088562 & \bfseries 0.2054794579744339 \\
\midrule
\midrule
\multirow{12}{*}{\rotatebox{90}{SocialEvo}}
& \tigger & 1\,h & 0.9230769276618958 & 1.0 & 0.692307710647583 & 1.0 & 0.7692307829856873 & 0.9230769276618958 & 0.9230769276618958 & 1.0 \\
& \dggen & 1\,h & 0.6190476417541504 & 0.8095238208770752 & 0.9047619104385376 & 0.8095238208770752 & 0.4761904776096344 & 0.4761904776096344 & \bfseries 0.7142857313156128 & \bfseries 0.523809552192688 \\
& \modelname & 1\,h & \bfseries 0.2783171534538269 & \bfseries 0.268608421087265 & \bfseries 0.1618122905492782 & \bfseries 0.0954692587256431 & \bfseries 0.2508090734481811 & \bfseries 0.2864077687263489 & 0.7200646996498108 & 0.647249162197113 \\
\cmidrule{2-11}
& \tigger & 2\,h & 0.8461538553237915 & 1.0 & 0.7692307829856873 & 1.0 & 0.692307710647583 & 0.9230769276618958 & \bfseries 0.9230769276618958 & 1.0 \\
& \dggen & 2\,h & 0.7272727489471436 & 0.8181818127632141 & 1.0 & 1.0 & 0.5454545617103577 & 0.6363636255264282 & 1.0 & \bfseries 0.3636363744735718 \\
& \modelname & 2\,h & \bfseries 0.1359223276376724 & \bfseries 0.1423948258161544 & \bfseries 0.3333333432674408 & \bfseries 0.252427190542221 & \bfseries 0.1003236249089241 & \bfseries 0.6181229948997498 & 0.9514563083648682 & 0.7443365454673767 \\
\cmidrule{2-11}
& \tigger & 4\,h & 0.7692307829856873 & 0.9230769276618958 & 0.7692307829856873 & 0.8461538553237915 & 0.4615384638309479 & 0.9230769276618958 & \bfseries 0.9230769276618958 & 1.0 \\
& \dggen & 4\,h & 0.8333333134651184 & 0.8333333134651184 & 0.8333333134651184 & 0.8333333134651184 & 0.6666666865348816 & \bfseries 0.3333333432674408 & 1.0 & \bfseries 0.5 \\
& \modelname & 4\,h & \bfseries 0.3935483992099762 & \bfseries 0.3483870923519134 & \bfseries 0.6387096643447876 & \bfseries 0.4838709533214569 & \bfseries 0.2193548381328582 & 0.7096773982048035 & 0.9935483932495116 & 0.896774172782898 \\
\cmidrule{2-11}
& \tigger & 8\,h & \bfseries 0.4615384638309479 & 0.8461538553237915 & \bfseries 0.692307710647583 & 0.8461538553237915 & \bfseries 0.3076923191547394 & 0.8461538553237915 & \bfseries 0.8461538553237915 & 1.0 \\
& \dggen & 8\,h & 0.75 & 0.75 & 0.75 & \bfseries 0.75 & 0.75 & \bfseries 0.25 & 1.0 & \bfseries 0.5 \\
& \modelname & 8\,h & 0.6153846383094788 & \bfseries 0.5769230723381042 & 0.8974359035491943 & 0.7564102411270142 & 0.5769230723381042 & 0.7051281929016113 & 1.0 & 0.9743589758872986 \\
\midrule
\midrule
\multirow{12}{*}{\rotatebox{90}{Contacts}}
& \tigger & 10\,min & \bfseries 0.16639742255210876 & 0.8449111580848694 & 0.9402261972427368 & \bfseries 0.25363489985466003 & \bfseries 0.15024232864379883 & 0.9983844757080078 & 0.591276228427887 & \bfseries 0.2681744694709778 \\
& \dggen & 10\,min & 0.7962962985038757 & 0.9074074029922484 & 0.5740740895271301 & 0.8703703880310059 & 0.7222222089767456 & 0.9629629850387572 & 0.7037037014961243 & 0.7037037014961243 \\
& \modelname & 10\,min & 0.8074533939361572 & \bfseries 0.3105590045452118 & \bfseries 0.1283643841743469 & 0.4265010356903076 & 0.8012422323226929 & \bfseries 0.5693581700325012 & \bfseries 0.2008281499147415 & 0.6314699649810791 \\
\cmidrule{2-11}
& \tigger & 20\,min & \bfseries 0.42580646276474 & 0.9548386931419371 & 0.8870967626571655 & \bfseries 0.3129032254219055 & \bfseries 0.4354838728904724 & 0.996774196624756 & 0.7483870983123779 & \bfseries 0.43870967626571655 \\
& \dggen & 20\,min & 0.75 & 0.8500000238418579 & 0.574999988079071 & 0.8500000238418579 & 0.699999988079071 & 0.975000023841858 & 0.7250000238418579 & 0.625 \\
& \modelname & 20\,min & 0.6276595592498779 & \bfseries 0.2269503474235534 & \bfseries 0.2553191483020782 & 0.3439716398715973 & 0.652482271194458 & \bfseries 0.6773049831390381 & \bfseries 0.3333333432674408 & 0.6595744490623474 \\
\cmidrule{2-11}
& \tigger & 40\,min & 0.5870967507362366 & 0.9741935729980468 & 0.7548387050628662 & 0.5612903237342834 & 0.6000000238418579 & 0.9935483932495116 & 0.7225806713104248 & \bfseries 0.6451612710952759 \\
& \dggen & 40\,min & 0.8064516186714172 & 0.8064516186714172 & 0.6129032373428345 & 0.8709677457809448 & 0.774193525314331 & 0.9677419066429138 & 0.8064516186714172 & \bfseries 0.6451612710952759 \\
& \modelname & 40\,min & \bfseries 0.3733333349227905 & \bfseries 0.3400000035762787 & \bfseries 0.5199999809265137 & \bfseries 0.3400000035762787 & \bfseries 0.3400000035762787 & \bfseries 0.4266666769981384 & \bfseries 0.5733333230018616 & 0.7066666483879089 \\
\cmidrule{2-11}
& \tigger & 80\,min & 0.6025640964508057 & 0.9743589758872986 & 0.6282051205635071 & 0.6025640964508057 & 0.6282051205635071 & 1.0 & \bfseries 0.692307710647583 & 0.7820512652397156 \\
& \dggen & 80\,min & 0.8399999737739563 & 0.6800000071525574 & 0.6000000238418579 & 0.9200000166893004 & 0.800000011920929 & 0.959999978542328 & 0.800000011920929 & \bfseries 0.6399999856948853 \\
& \modelname & 80\,min & \bfseries 0.3461538553237915 & \bfseries 0.5641025900840759 & \bfseries 0.5641025900840759 & \bfseries 0.4487179517745972 & \bfseries 0.2820512950420379 & \bfseries 0.6282051205635071 & 0.8717948794364929 & 0.7948718070983887 \\
\bottomrule
\end{tabular}}

   \label{tab:topo_dist_ks_dgb}
\end{table}

\endgroup

\subsection{Extension to Link Prediction}
\label{sec:link_prediction}
A trained \modelname can also serve as a zero-shot, parameter-free link predictor. We outline this extension here and leave its evaluation to future work.
To derive a link predictor from a trained CTDG generative diffusion model, we leverage the density-estimation~\citep{liu2025survey} properties of the diffusion process and estimate the conditional edge probability between a node pair $(\src, \dst)$ from the reconstruction error after a forward-diffusion step.
Intuitively, positive edges drawn from the real data distribution should yield a lower reconstruction error than negative edges.
We adopt a simple formulation for the reconstruction score $\reconscore$:
\begin{equation}
\label{eq:reconstruction_score}
   \reconscore =
   \E_{\epsilon \sim \mathcal{N}(0, \mI), \tdiff \sim \tstrat}
      \left[
         - \sum_{f = 1}^{F} \frac{1}{\|z^f\|}d^f\left( z_\tdiff^f + (1 - \tdiff) v_\theta(z_\tdiff, \tdiff)^f , z^f \right)
      \right]
\end{equation}
where $F$ is the number of components of $z$ (\cref{eq:tokenization:with_class}), $z^f$ denotes the $f$-th component of $z$, and $d^f(\cdot)$ is a distance function (\eg the mean absolute error). The distances are averaged over multiple random noise samples and time steps (\eg with uniform time-step sampling, although a fixed $\tdiff^*$ could be further fine-tuned).
This formulation is similar to casting link prediction as anomaly detection \citep{wolleb2022diffusion}, where negative edges lie off the learned data manifold and so produce a high reconstruction error.

\end{document}